\documentclass{UVKABoA4}
\pdfoutput=1
\usepackage[english]{babel}
\usepackage[utf8]{inputenc}
\usepackage{graphicx} 
\usepackage[table]{xcolor}
\usepackage{amssymb}
\usepackage{amsmath}
\usepackage{mathtools}
\usepackage{IEEEtrantools}
\usepackage{relsize} 
\usepackage[absolute,overlay]{textpos}
\usepackage{vmargin}          
\usepackage[inline]{enumitem} 
\usepackage{parskip}
\usepackage{shortcuts}
\usepackage{pgfplots}
\pgfplotsset{compat=1.13}
\usepackage{wrapfig}
\usepackage{makecell}
\usepackage{tikz}
\usetikzlibrary{arrows.meta}
\usetikzlibrary{decorations.pathreplacing}
\usetikzlibrary{positioning}
\usetikzlibrary{decorations.text}
\usetikzlibrary{decorations.pathmorphing}
\usetikzlibrary{shapes.multipart, calc}

\usepackage{mathtools}
\usepackage{mathdots}
\usepackage{mathrsfs}
\def\wasyfamily{\fontencoding{U}\fontfamily{wasy}\selectfont}
\usepackage{upgreek}
\usepackage{marvosym}
\usepackage{textcomp}
\usepackage{dsfont}
\usepackage{esint}
\usepackage{cmll}
\usepackage{stmaryrd}
\usepackage{skull}
\usepackage{gensymb}
\usepackage{url}

\DeclareMathOperator*{\sech}{sech}
\DeclareMathOperator*{\conv}{conv}
\DeclareMathOperator*{\ReLU}{ReLU}

\usepackage[english]{babel}

\usepackage[raiselinks=true,
            bookmarks=true,
            bookmarksopenlevel=1,
            bookmarksopen=true,
            bookmarksnumbered=true,
            hyperindex=true,
            plainpages=false,
            pdfpagelabels=true,
            pdfborder={0 0 0.5}]{hyperref}

\makeatletter
\def\tagform@#1{\maketag@@@{\ignorespaces [#1]\unskip\@@italiccorr}}	
\makeatother

\makeindex

\newcommand{\myname}{Martin~Thoma}
\newcommand{\headtitle}{Analysis and Optimization of Convolutional Neural Network Architectures}
\newcommand{\floatingtitle}{\headtitle}
\newcommand{\thesistype}{Master Thesis}

\newcommand{\reviewerone}{Prof.~Dr.--Ing.~R.~Dillmann}
\newcommand{\reviewertwo}{Prof.~Dr.--Ing.~J.~M.~Zöllner}
\newcommand{\advisor}{Dipl.--Inform.~Michael~Weber}

\newcommand{\timestart}{03. May 2017} 
\newcommand{\releaseyear}{2017}
\newcommand{\timeend}{03. August \releaseyear} 
\newcommand{\releasemonth}{August \releaseyear}

\newcommand{\department}{\iflanguage{english}{Department of Computer Science}
                                             {Fakultät für Informatik}}
\newcommand{\institute}{\iflanguage{english}{Institute for Anthropomatics}
                                            {Institut für Anthropomatik}}
\newcommand{\fzidepartment}{Abteilung Technisch Kognitive Assistenzsysteme}
\newcommand{\fziname}{\iflanguage{english}{FZI Research Center for Information Technology}
                                          {FZI Forschungszentrum Informatik}}

\newcommand{\titlefig}{\includegraphics[width=1.0\textwidth]{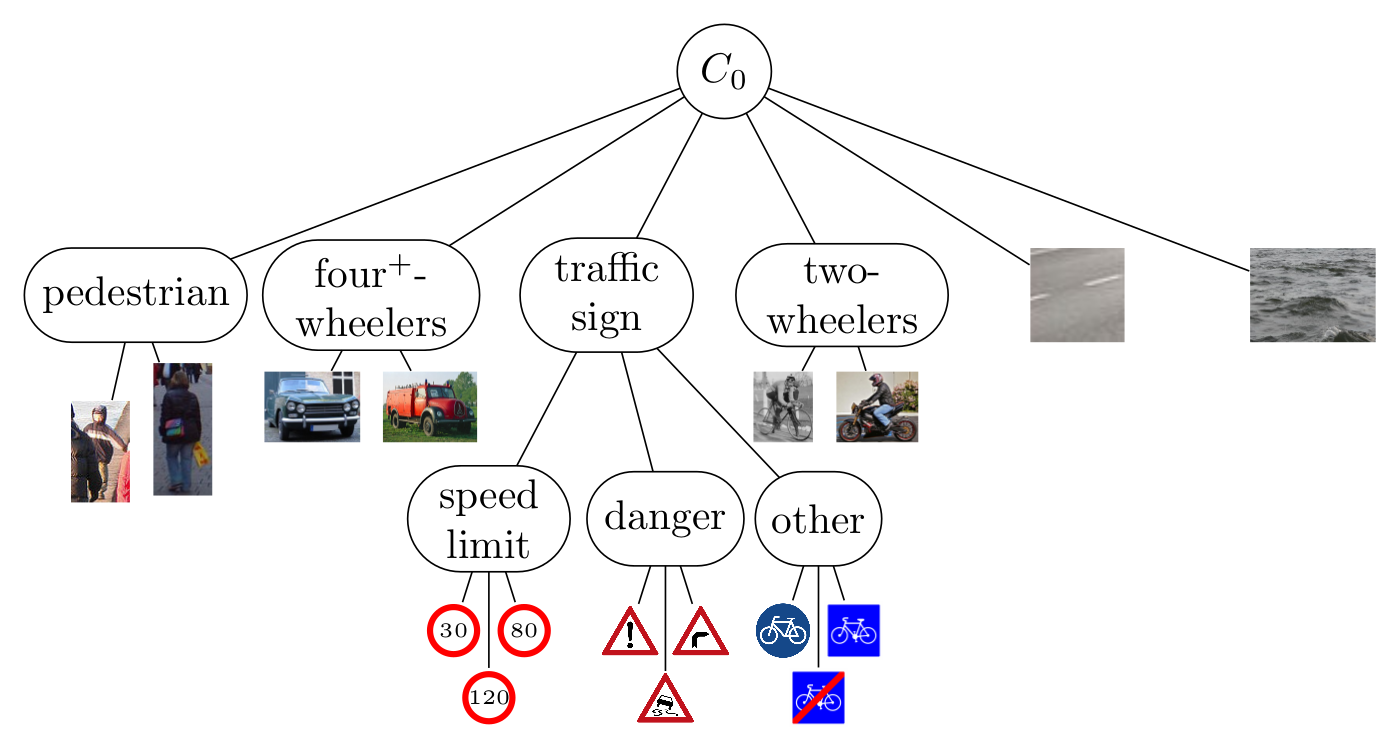}}

\hypersetup{
  pdfauthor   = {\myname},
  pdfkeywords = {recognition, machine learning, neural networks, cnn},
  pdfsubject  = {CNN Construction},
  pdftitle    = {\headtitle},
}
\usepackage{csquotes}
\usepackage{booktabs}
\usepackage{multirow}
\usepackage{braket}
\usepackage{algorithm,algpseudocode}
\algtext*{EndIf}        
\algtext*{EndWhile}     
\algtext*{EndFunction}  
\algtext*{EndFor}       
\algtext*{EndProcedure}       
\algnewcommand{\LineComment}[1]{\State  \(\triangleright\) #1 \hfill~} 
\usepackage{subfig}
\usepackage{float}
\usepackage[nameinlink, noabbrev,capitalise]{cleveref} 
\usepackage[xindy,toc,section=chapter,numberedsection=autolabel]{glossaries}
\usepackage[binary-units,group-separator={,}]{siunitx}
\DeclareSIUnit\pixel{px}
\DeclareSIUnit\epoch{epoch}
\DeclareSIUnit\float{float}
\DeclareSIUnit\floats{floats}
\DeclareSIUnit{\nothing}{\relax}
\usepackage{etoolbox}
\newcommand{\ubold}{\fontseries{b}\selectfont}
\robustify\ubold

\usepackage{microtype}
\newacronym{ANN}{ANN}{artificial neural network}
\newacronym{ANOVA}{ANOVA}{analysis of variance}
\newacronym{ASO}{ASO}{Automatic Structure Optimization}
\newacronym{CMO}{CMO}{Confusion Matrix Ordering}
\newacronym{CE}{CE}{cross entropy}
\newacronym{CUDA}{CUDA}{Compute Unified Device Architecture}
\newacronym{CNN}{CNN}{Convolutional Neural Network}
\newacronym{CSR}{CSR}{cursive script recognition}
\newacronym{CFM}{CFM}{classification figure of merit}
\newacronym{DTW}{DTW}{dynamic time warping}
\newacronym{ELU}{ELU}{Exponential Linear Unit}
\newacronym{ES}{ES}{early stopping}
\newacronym{FLOP}{FLOP}{floating point operation}
\newacronym{FC}{FC}{Fully Connected}
\newacronym{GA}{GA}{genetic algorithm}
\newacronym{GPU}{GPU}{graphics processing unit}
\newacronym{GAN}{GAN}{Generative Adverserial Network}
\newacronym{GMM}{GMM}{Gaussian mixture model}
\newacronym{GTW}{GTW}{greedy time warping}
\newacronym{HMM}{HMM}{hidden Markov model}
\newacronym{HWR}{HWR}{handwriting recognition}
\newacronym{HWRT}{HWRT}{handwriting recognition toolkit}
\newacronym{HSV}{HSV}{hue, saturation, value}
\newacronym{LReLU}{LReLU}{leaky rectified linear unit}
\newacronym{LDA}{LDA}{linear discriminant analysis}
\newacronym{LCN}{LCN}{Local Contrast Normalization}
\newacronym{MLP}{MLP}{multilayer perceptron}
\newacronym{MSE}{MSE}{mean squared error}
\newacronym{NAG}{NAG}{Nesterov Accellerated Momentum}
\newacronym{NEAT}{NEAT}{NeuroEvolution of Augmenting Topologies}
\newacronym{OBD}{OBD}{Optimal Brain Damage}
\newacronym{OOV}{OOV}{out of vocabulary}
\newacronym{PCA}{PCA}{principal component analysis}
\newacronym{PyPI}{PyPI}{Python Package Index}
\newacronym{PReLU}{PReLU}{parametrized rectified linear unit}
\newacronym{SGD}{SGD}{stochastic gradient descent}
\newacronym{TDNN}{TDNN}{time delay neural network}
\newacronym{SVM}{SVM}{support vector machine}
\newacronym{SLP}{SLP}{supervised layer-wise pretraining}
\newacronym{ReLU}{ReLU}{rectified linear unit}
\newacronym{ZCA}{ZCA}{Zero Components Analysis}

\makeglossaries
\DeclareMathVersion{normal2}
\begin{document}
\bstctlcite{IEEEexample:BSTcontrol}
\pagenumbering{roman}

\newcommand{\diameter}{20}
\newcommand{\xone}{-20}
\newcommand{\xtwo}{150}
\newcommand{\yone}{25}
\newcommand{\ytwo}{-243}

\begin{titlepage}
\begin{tikzpicture}[overlay]
\draw[color=gray]
 		 (\xone mm, \yone mm)
  -- (\xtwo mm, \yone mm)
 arc (90:0:\diameter pt)
  -- (\xtwo mm + \diameter pt , \ytwo mm)
	-- (\xone mm + \diameter pt , \ytwo mm)
 arc (270:180:\diameter pt)
	-- (\xone mm, \yone mm);
\end{tikzpicture}
  \begin{textblock}{10}[0,0](3.35,2.55)
	  \iflanguage{english}  {\includegraphics[width=.3\textwidth]{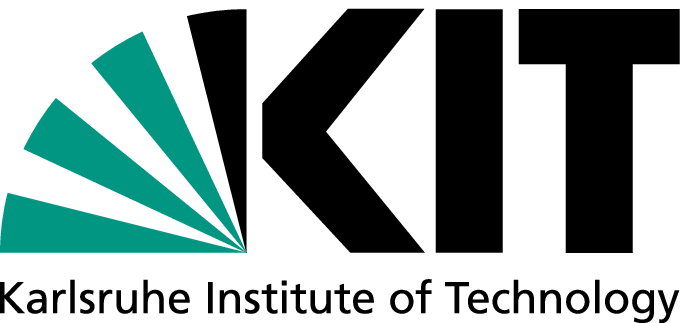}}
                          {\includegraphics[width=.3\textwidth]{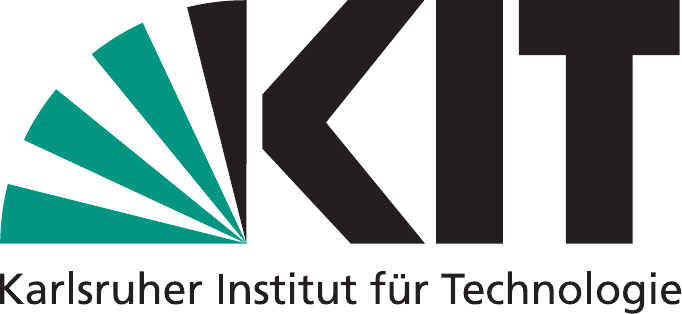}}
  \end{textblock}
	\changefont{phv}{m}{n}	
	\vspace*{2.5cm}
	\begin{center}
		\Huge{\headtitle}
		\vspace*{2cm}\\
		\Large{
			\iflanguage{english}{Master Thesis of}
												  {Masterarbeit\\von}
		}\\
		\vspace*{1cm}
		\huge{\myname}\\
		\vspace*{1cm}
		\Large{
			\department\\ \institute\\ \iflanguage{english}{and}{und}\\ \fziname
		}
	\end{center}
	\vspace*{1.5cm}
\Large{
\begin{center}
\begin{tabular}[ht]{l c l}
  \iflanguage{english}{Reviewer}{Erstgutachter}: & \hfill  & \reviewerone\\
  \iflanguage{english}{Second reviewer}{Zweitgutachter}: & \hfill  & \reviewertwo\\
  \iflanguage{english}{Advisor}{Betreuender Mitarbeiter}: & \hfill  & \advisor\\
\end{tabular}
\end{center}
}

\vspace{2cm}
\begin{center}
\large{\iflanguage{english}{Research Period}{Bearbeitungszeit}: \timestart \hspace*{0.25cm} -- \hspace*{0.25cm} \timeend}
\end{center}

\begin{textblock}{10}[0,0](4,16.8)
\tiny{
	\iflanguage{english}
		{KIT -- University of the State of Baden-Wuerttemberg and National Research Center of the Helmholtz Association}
		{KIT -- Universität des Landes Baden-Württemberg und nationales Forschungszentrum in der Helmholtz-Gemeinschaft}
}
\end{textblock}

\begin{textblock}{10}[0,0](14,16.75)
\large{
	\textbf{www.kit.edu}
}
\end{textblock}

\end{titlepage}

\pagestyle{empty}
\begingroup
\changefont{phv}{m}{n}
\renewcommand{\baselinestretch}{1}

\begin{titlepage}
  \parindent 0em 
  \LARGE\textbf{\floatingtitle}
  \vspace{24pt}\\
  \iflanguage{english}{by}{von}\\
  \myname \\[1.5cm]
  \begin{center}
  \titlefig\\[1cm]
  \end{center}
  \vfill
  \begin{minipage}{0.4\textwidth}
    \begin{flushleft} \large
      \LARGE\textbf{\thesistype}\\
      \LARGE\releasemonth
    \end{flushleft}
  \end{minipage}
  \begin{minipage}{0.6\textwidth}
    \begin{flushright}
      \includegraphics[width=1.5cm]{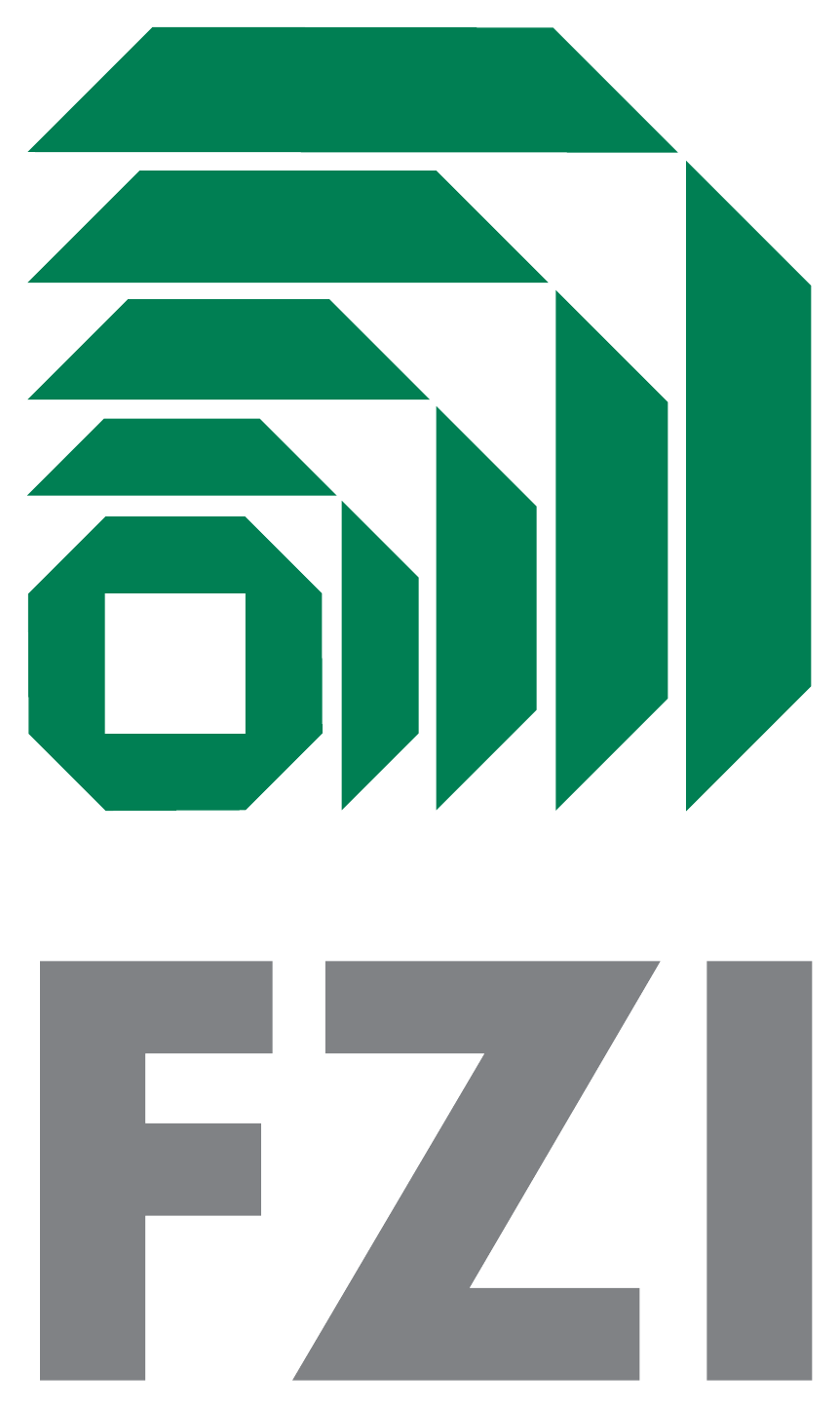}
    \end{flushright}
  \end{minipage}
  \newpage
  \small \thesistype, FZI\\
  \department, \releaseyear\\
  Gutachter: \reviewerone, \reviewertwo
  \vfill
  \fzidepartment\\
  \fziname
  \newpage
\end{titlepage}

\endgroup

\affirmation

Ich versichere wahrheitsgemäß, die Arbeit selbstständig angefertigt, alle benutzten Hilfsmittel vollständig und genau angegeben und alles kenntlich gemacht zu haben, was aus Arbeiten anderer unverändert oder mit Abänderungen entnommen wurde.

\vspace{2cm}
\begin{flushright}\noindent
    Karlsruhe,\hfill {\textit{\myname}}\\
    \releasemonth \hfill { }
\end{flushright}

\abstract

Convolutional Neural Networks (CNNs) dominate various computer vision tasks
since Alex Krizhevsky showed that they can be trained effectively and reduced
the top-5 error from \SI{26.2}{\percent} to \SI{15.3}{\percent} on the ImageNet
large scale visual recognition challenge. Many aspects of CNNs are examined in
various publications, but literature about the analysis and construction of
neural network architectures is rare. This work is one step to close this gap.
A comprehensive overview over existing techniques for CNN analysis and topology
construction is provided. A novel way to visualize classification errors with
confusion matrices was developed. Based on this method, hierarchical
classifiers are described and evaluated. Additionally, some results are
confirmed and quantified for CIFAR-100. For example, the positive impact of
smaller batch sizes, averaging ensembles, data augmentation and test-time
transformations on the accuracy. Other results, such as the positive impact of
learned color transformation on the test accuracy could not be confirmed. A
model which has only one million learned parameters for an input size of $32
\times 32 \times 3$ and 100~classes and which beats the state of the art on the
benchmark dataset Asirra, GTSRB, HASYv2 and STL-10 was developed.

\clearpage

\section*{Zusammenfassung}
Modelle welche auf Convolutional Neural Networks (CNNs) basieren sind in
verschiedenen Aufgaben der Computer Vision dominant seit Alex Krizhevsky
gezeigt hat dass diese effektiv trainiert werden können und er den Top-5 Fehler
in dem ImageNet large scale visual recognition challenge Benchmark von
\SI{26.2}{\percent} auf \SI{15.3}{\percent} drücken konnte. Viele Aspekte von
CNNs wurden in verschiedenen Publikationen untersucht, aber es wurden
vergleichsweise wenige Arbeiten über die Analyse und die Konstruktion von
Neuronalen Netzen geschrieben. Diese Masterarbeit stellt einen Schritt dar um
diese Lücke zu schließen. Eine umfassende Überblick über Analyseverfahren und
Topologielernverfahren wird gegeben. Ein neues Verfahren zur Visualisierung der
Klassifikationsfehler mit Konfusionsmatrizen wurde entwickelt. Basierend auf
diesem Verfahren wurden hierarchische Klassifizierer eingeführt und evaluiert.
Zusätzlich wurden einige bereits in der Literatur beschriebene Beobachtungen
wie z.B. der positive Einfluss von kleinen Batch-Größen, Ensembles, Erhöhung
der Trainingsdatenmenge durch künstliche Transformationen (Data Augmentation)
und die Invarianzbildung durch künstliche Transformationen zur Test-Zeit
(Test-time transformations) experimentell bestätigt. Andere Beobachtungen, wie
beispielsweise der positive Einfluss gelernter Farbraumtransformationen konnten
nicht bestätigt werden. Ein Modell welches weniger als eine Millionen Parameter
nutzt und auf den Benchmark-Datensätzen Asirra, GTSRB, HASYv2 und STL-10 den
Stand der Technik neu definiert wurde entwickelt.

\ack

I would like to thank Stephan Gocht and Marvin Teichmann for the many inspiring
conversations we had about various topics, including machine learning.

I also want to thank my father for the support he gave me. He made it possible
for me to study without having to worry about anything besides my studies.
Thank you!

Finally, I want to thank Timothy Gebhard, Daniel Schütz and Yang Zhang for
proof-reading my masters thesis and Stephan Gocht for giving me access to a
GTX~1070.

\clearpage

This work can be cited the following way:

\begin{verbatim}
@MastersThesis{Thoma:2017,
    Title     = {Analysis and Optimization of Convolutional Neural Network
                 Architectures},
    Author    = {Martin Thoma},
    School    = {Karlsruhe Institute of Technology},
    Year      = {2017},

    Address   = {Karlsruhe, Germany},
    Month     = jun,
    Type      = {Masters's Thesis},

    Keywords  = {machine learning; artificial neural networks;
                 classification; supervised learning; CNNs},
    Url       = {https://martin-thoma.com/msthesis/}
}
\end{verbatim}

A DVD with a digital version of this master thesis and the source code as
well as the used data is part of this work.

\begingroup
\changefont{phv}{m}{n}
\tableofcontents
\endgroup

\mainmatter
\renewcommand{\chapterpagestyle}{plain}
\pagestyle{scrheadings}
\pagenumbering{arabic}

\chapter{Introduction}
Computer vision is the academic field which aims to gain a high-level
understanding of the low-level information given by raw pixels from digital
images.

Robots, search engines, self-driving cars, surveillance agencies and many
others have applications which include one of the following six problems
in computer vision as sub-problems:
\begin{itemize}
    \item \textbf{Classification}:\footnote{Classification is also called
    \textit{identification} if the classes are humans. Another name is
    \textit{object recognition}, although the classes can be humans and animals
    as well.} The algorithm is given an image and $k$ possible classes. The task
    is to decide which of the $k$ classes the image belongs to. For example,
    an image from a self-driving cars on-board camera contains either
    \verb+paved road+, \verb+unpaved road+ or \verb+no road+: Which of those
    given three classes is in the image?
    \item \textbf{Localization}: The algorithm is given an image and one class~$k$.
    The task is to find bounding boxes for all instances of $k$.
    \item \textbf{Detection}: Given an image and $k$ classes, find bounding
    boxes for all instances of those classes.
    \item \textbf{Semantic Segmentation}: Given an image and $k$ classes,
    classify each pixel.
    \item \textbf{Instance segmentation}: Given an image and $k$ classes,
    classify each pixel as one of the $k$ classes, but distinguish different
    instances of the classes.
    \item \textbf{Content-based Image Retrieval}: Given an image $x$ and $n$
    images in a database, find the top $u$ images which are most similar to
    $x$.
\end{itemize}

There are many techniques to approach those problems, but since
AlexNet~\cite{AlexNet-2012} was published, all of those problems have
high-quality solutions which make use of
\glspl{CNN}~\cite{deep-residual-networks-2015,liu2016ssd,ronneberger2015u,dai2016instance,schroff2015facenet}.

Today, most neural networks are constructed by rules of thumb and gut feeling.
The architectures evolved and got deeper, more hyperparameters were added.
Although there are methods for analyzing \glspl{CNN}, those
methods are not enough to determine all steps in the development of
network architectures without gut feeling. A detailed introduction to
\glspl{CNN} as well as nine methods for analysis of
\glspl{CNN} is given in \cref{ch:CNN}.

Despite the fact that most researchers and developers do not use topology
learning, a couple of algorithms have been proposed for this task. Five classes
of topology learning algorithms are introduced in~\cref{ch:topology-learning}.

When datasets and the number of classes are large, evaluating a single idea
how to improve the network can take several weeks just for the training. Hence
the idea of building a hierarchy of classifiers which allows to split the
classification task into various sub-tasks that can easily be combined is
evaluated in~\cref{ch:hierarchical-classification}.

\Gls{CMO}, the hierarchical classifier, 9~types of hyperparameters and label
smoothing are evaluated in \cref{ch:experimental-evaluation}.

This work focuses on classification problems to keep the presented ideas as
pure and simple as possible. The described techniques are relevant to all six
described computer vision problems due to the fact that Encoder-Decoder
architectures are one component of state-of-the-art algorithms for all six of
them.

\chapter{Convolutional Neural Networks}\label{ch:CNN}

In the following, it is assumed that the reader knows what a \gls{MLP} is and
how they are designed for classification problems, what activation functions
are and how gradient descent works. In case the reader needs a refresher on any
of those topics, I recommend chapter~4.3~and~4.4 of~\cite{Thoma:2014} as well
as~\cite{lecun2015deep}.

This chapter introduces linear image filters in~\cref{sec:image-filters},
then standard layer types of \glspl{CNN} are explained
in~\cref{sec:CNN-layer-types}. The layer block pattern is
described in~\cref{sec:cnn-blocks}, transition layers
in~\cref{sec:transition-layers} and nine~ways to analyze
\glspl{CNN} are described in~\cref{sec:analyzation-techniques}.

\section{Linear Image Filters}\label{sec:image-filters}
\begin{figure}[H]
    \centering
    \resizebox {0.85\columnwidth} {!} {
    \input{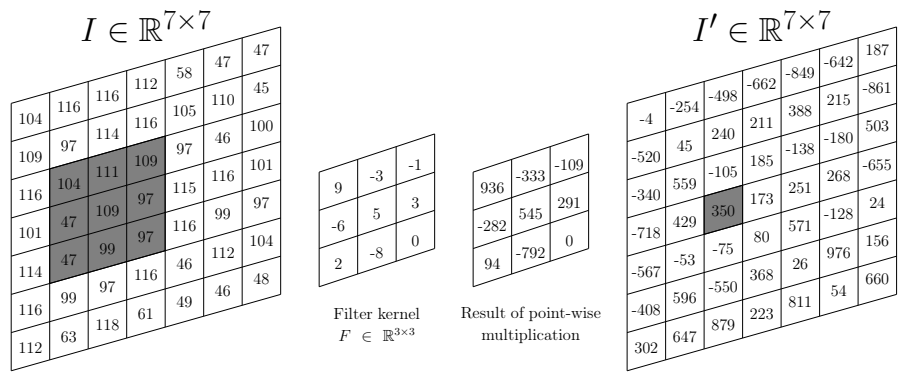}
    }
    \caption[Application of a single image filter (Convolution)]{Visualization of the
               application of a linear $k \times k \times 1$ image filter. For
               each pixel of the output image, $k^2$ multiplications and $k^2$
               additions of the products have to be calculated.}
    \label{fig:image-filter-visualization}
\end{figure}

A \textit{linear image filter} (also called a \textit{filter bank} or a
\textit{kernel}) is an element $F \in \mathbb{R}^{k_w \times k_h \times d}$,
where $k_w$ represents the filter's width, $k_h$ the filter's height and $d$
the number of input channels.
The filter $F$ is convolved with the image $I \in \mathbb{R}^{w \times h \times
d}$ to produce a new image $I'$. The output image $I'$ has only one channel.
Each pixel $I'(x, y)$ of the output image gets calculated by point-wise
multiplication of one filter element with one element of the original image
$I$:
\[I'(x, y) = \sum_{i_x = 1 - \lceil \frac{k_w}{2} \rceil}^{\lfloor \frac{k_w}{2} \rfloor} \sum_{i_y = 1 - \lceil \frac{k_h}{2} \rceil}^{\lfloor \frac{k_h}{2} \rfloor} \sum_{i_c = 1}^{d} I(x + i_x, y+i_y, i_c) \cdot F(i_x, i_y, i_c)\]
This procedure is explained by~\cref{fig:image-filter-visualization}. It
is essentially a discrete convolution.

One important detail is how boundaries are treated. There are four common ways
of boundary treatment:
\begin{itemize}
    \item \textbf{don't compute}: The image $I'$ will be smaller than the
          original image. $I' \in \mathbb{R}^{(w - k_w + 1) \times (h - k_h + 1) \times d_3}$, to be exact.
    \item \textbf{zero padding}: The image $I$ is padded by zeros where the filter
          would access elements which do not exist. This will result in edges
          being detected at the border if the border pixels are not black, but
          doesn't need any computation.
    \item \textbf{nearest}: Repeat the pixel which is closest to the boundary.
    \item \textbf{reflect}: Reflect the image at the boundaries.
\end{itemize}

Common tasks that can be done with linear filters include edge detection,
corner detection, smoothing, sharpening, median filtering, box filtering.
See~\cref{fig:image-filters} for five examples.

Please note that the result of a filtering operation is again an image. This
means filters can be applied successively. While each pixel after one filtering
operation with a $3\times3$ filter got influenced by $3 \cdot 3 = 9$~pixels of
the original image, two successively applied $3 \times 3$ filters increase the
area of the original image which influenced the output. The output is then
influenced by 25~pixel. This is called the \textit{receptive field}. The kind of
pattern which is detected by a filter is called a \textit{feature}. The bigger
the receptive field is, the more complex can features get as they are able to
consider more of the original image. Instead of taking one $5 \times 5$ filter
with 25~parameters, one might consider to take two successive $3 \times 3$
filters with $2 \cdot (3 \cdot 3) = 18$~parameters. The $5 \times 5$ filter is
a strict superset of possible filtering operations compared to the two $3
\times 3$ filters, but the relevance of this technique will become clear
in~\cref{sec:CNN-layer-types}.

\section{CNN Layer Types}\label{sec:CNN-layer-types}
While the idea behind deep \glspl{MLP} is that \textit{feature hierarchies}
capture the important parts of the input more easily, \glspl{CNN} are inspired
by the idea of \textit{translational invariance}: Many features in an image are
translationally invariant. For example, if a car is developed, one could try to
detect it by its parts~\cite{felzenszwalb2010object}. But then there are many
positions at which the wheels could be. Combining those, it is desirable to
capture low-level, translationally invariant features at lower layers of an
\gls{ANN} and in higher layers high-level features which are combinations of
the low-level features.

Also, models should utilize the fact that the pixels of images are ordered. One
way to use this is by learning image filters in so called \textit{convolutional
layers}.

While \glspl{MLP} vectorize the input, the input of a layer in a \gls{CNN} are
\textit{feature maps}. A feature map is a matrix $m \in \mathbb{R}^{w \times h}$, but typically the width equals the height ($w = h$). For an RGB
input image, the number of feature maps is $d=3$.
Each color channel is a feature map.

Since AlexNet~\cite{AlexNet-2012} almost halved the error in the ImageNet
challenge, \glspl{CNN} are state-of-the-art in various computer vision tasks.

Traditional \glspl{CNN} have three important building tools:
\begin{itemize}
    \item Convolutional layers with a non-linear activation function as
          described in~\cref{subsec:convolutional-layers},
    \item pooling layers as described in~\cref{subsec:pooling-layers} and
    \item normalization layers as described in~\cref{sec:batch-normalization}.
\end{itemize}

\subsection{Convolutional Layers}\label{subsec:convolutional-layers}
Convolutional layers take several feature maps as input and produce $n$ feature
maps\footnote{also called \textit{activation maps} or \textit{channels}} as
output, where $n$ is the number of filters in the convolution layer. The filter
weights of the linear convolutions are the parameters which are adapted to the
training data. The number~$n$ of filters as well as the filter's size $k_w \times k_h$
are hyperparameters of convolutional layers. Sometimes, it is denoted as $n@k_w
\times k_h$. Although the filter depth is usually omitted in the notation,
the filters are of dimension $k_w \times k_h \times d^{(i-1)}$, where
$d^{(i-1)}$ is the number of feature maps of the input layer $(i-1)$.

Another hyperparameter of convolution layers is the
stride $s \in \mathbb{N}_{\geq 1}$ and the padding. Padding (usually
zero-padding~\cite{sermanet2012convolutional,sermanet2013overfeat,deep-residual-networks-2015})
is used to make sure that the size of the feature maps doesn't change.

The hyperparameters of convolutional layers are
\begin{itemize}
    \item the number of filters $n \in \mathbb{N}_{\geq 1}$,
    \item $k_w, k_h \in \mathbb{N}_{\geq 1}$ of the filter size $k_w \times k_h \times d^{(i-1)}$,
    \item the activation function of the layer (see~\cref{table:activation-functions-overview}) and
    \item the stride $s \in \mathbb{N}_{\geq 1}$
\end{itemize}

Typical choices are $n \in \Set{32, 64, 128}$, $k_w = k_h = k \in \Set{1, 3, 5, 11}$ such
as in~\cite{AlexNet-2012,VGG-16,GoogleNet-Inception}, \gls{ReLU} activation and $s=1$.

The concept of weight sharing is crucial for \glspl{CNN}. This concept was
introduced in~\cite{waibel1989phoneme}. With weight sharing, the filters can be
learned with \gls{SGD} just like \glspl{MLP}. In fact, every \gls{CNN} has an
equivalent \gls{MLP} which computes the same function if only the flattened
output is compared.

This is easier to see when the filtering operation is denoted formally:
\begin{align}
    o^{(i)}(\mathbf{x}) &= b + \sum_{j=1}^k w_{ij} \cdot \mathbf{x}_{j} \qquad\text{with } i \in \Set{1, \dots, w} \times \Set{1, \dots, h} \times \Set{1, \dots, d}\\
    o^{(x, y, z)}(I) &= b + \sum_{i_x = 1 - \lceil \frac{k_w}{2} \rceil}^{\lfloor \frac{k_w}{2} \rfloor} \sum_{i_y = 1 - \lceil \frac{k_h}{2} \rceil}^{\lfloor \frac{k_h}{2} \rfloor} \sum_{i_c = 1}^{d} F_z(i_x, i_y, i_c) \cdot I(x + i_x, y+i_y, i_c)\\
    &\hphantom{=}\text{with a bias } b \in \mathbb{R}\text{, } x \in \Set{1, \dots, w} \text{, } y \in \Set{1, \dots, h}\text{ and } z \in \Set{1, \dots, d}\nonumber
\end{align}

One can see that most weights of the equivalent \gls{MLP} are zero and many
weights are equivalent. Hence the advantage of \glspl{CNN} compared to
\glspl{MLP} is the reduction of parameters. The effect of fewer parameters is
that less training data is necessary to get suitable estimations for those.
This means a \gls{MLP} which is able to compute the same functions as a
\gls{CNN} will likely have worse results on the same
dataset, if a \gls{CNN} architecture is suitable for the dataset.

\begin{figure}[ht]
    \centering
    \newcommand{\distance}{6}
\newcommand{\xup}{3.5}
\newcommand{\yup}{6}
\newcommand{\upsizex}{1}
\newcommand{\upsizey}{2}
\newcommand{\upshift}{3/4*\upsizey}
\newcommand{\distancedots}{1}

\begin{tikzpicture}
    \foreach \i in {0, 0.2, ..., 0.6} {
        \draw[fill=white] (0+\i, 0) -- (2+\i, 3) -- (2+\i, 7) -- (\i, 4) -- (\i, 0);
    }

    \foreach \i in {0, 0.2, ..., 0.6, 1.3} {
        \draw[fill=white] (\xup+\i, \yup) -- (\xup+\upsizex+\i, \yup+\upshift) -- (\xup+\upsizex+\i, \yup+\upsizey+\upshift) -- (\xup+\i, \yup+\upsizey) -- (\xup+\i, \yup);
        \draw[fill=white] (\xup+\i, \yup) -- (\xup+\i+0.1, \yup) -- (\xup+\i+0.1, \yup+\upsizey) -- (\xup+\i, \yup+\upsizey) -- (\xup+\i, \yup);
        \draw[fill=white] (\xup+\i+0.1, \yup) -- (\xup+\upsizex+\i+0.1, \yup+\upshift) -- (\xup+\upsizex+\i+0.1, \yup+\upsizey+\upshift) -- (\xup+\i+0.1, \yup+\upsizey) -- (\xup+\i+0.1, \yup);
        \draw[fill=white] (\xup+\i, \yup+\upsizey) -- (\xup+\i+0.1, \yup+\upsizey) -- (\xup+\upsizex+\i+0.1, \yup+\upsizey+\upshift) -- (\xup+\upsizex+\i, \yup+\upsizey+\upshift) -- (\xup+\i, \yup+\upsizey);
    }

    \foreach \i in {0, 0.2, ..., 0.6, 1.2} {
        \draw[fill=white] (\distance+\i, 0) -- (\distance+2+\i, 3) -- (\distance+2+\i, 7) -- (\distance+\i, 4) -- (\distance+\i, 0);
    }

    \draw [decorate,decoration={brace,amplitude=+4pt,mirror},xshift=0pt,yshift=-2pt]
(-0.1,0) -- (0.7,0) node [black,midway,yshift=-0.6cm, align=center] {\footnotesize$3$ feature maps\\\footnotesize(e.g. RGB)};
    \draw [decorate,decoration={brace,amplitude=+4pt,mirror},xshift=0pt,yshift=-2pt]
(\distance-0.1,0) -- (\distance+1.3,0) node [black,midway,yshift=-0.6cm, align=center] {\footnotesize$n$ feature maps};
    \draw [decorate,decoration={brace,amplitude=+4pt},xshift=0pt,yshift=+2pt]
(\xup-0.1+\upsizex,\yup+\upsizey+\upshift) -- (\xup+1.5+\upsizex,\yup+\upsizey+\upshift) node [black,midway,yshift=+0.6cm, align=center] {\footnotesize $n$ filters of\\\footnotesize size $k \times k \times 3$};
    \draw[very thick, ->,>=latex] (3, 4.5) [out=70, in=110] to  (\distance-0.5, 4.5);
    \draw [color=white,decorate,decoration={brace,amplitude=+4pt, mirror},xshift=0pt,yshift=+2pt]
(1.1, 0) -- (3.1, 3) node [sloped,black,midway,yshift=+0.6cm, align=center] {width $w$};
    \draw [color=white,decorate,decoration={brace,amplitude=+4pt, mirror},xshift=0pt,yshift=+2pt]
(\distance+1.7, 0) -- (\distance+3.7, 3) node [sloped,black,midway,yshift=+0.6cm, align=center] {width $w$};
    \draw [decorate,decoration={brace,amplitude=+4pt},xshift=-2pt,yshift=0pt]
(0, 0) -- (0, 4) node [sloped,black,midway,yshift=+0.6cm, align=center] {height $h$};
    \draw [decorate,decoration={brace,amplitude=+4pt},xshift=-2pt,yshift=0pt]
(\distance, 0) -- (\distance, 4) node [sloped,black,midway,yshift=+0.6cm, align=center] {height $h$};
    \node at (-2.5,7.5) {\Large neural};
    \node at (-2.5,7) {\Large network};
    \node at (-2.5,2) {\Large data};

    \node at (+4.3,5.5) {apply};
    \node at (\distance+0.95,3.9) {\dots};
    \node at (\distance+0.95,2) {\dots};
    \node at (\distance+0.95,0.1) {\dots};

    \node at (\xup+1.05,\yup+1.9) {\dots};
    \node at (\xup+1.05,\yup+1.0) {\dots};
    \node at (\xup+1.05,\yup+0.1) {\dots};
\end{tikzpicture}
    \caption[Application of a convolutional layer]{Application of a single convolutional layer with $n$ filters of size $k \times k \times 3$ with stride $s=1$ to input data of size $\text{width} \times \text{height}$ with three channels.}
    \label{fig:convolution-layer}
\end{figure}
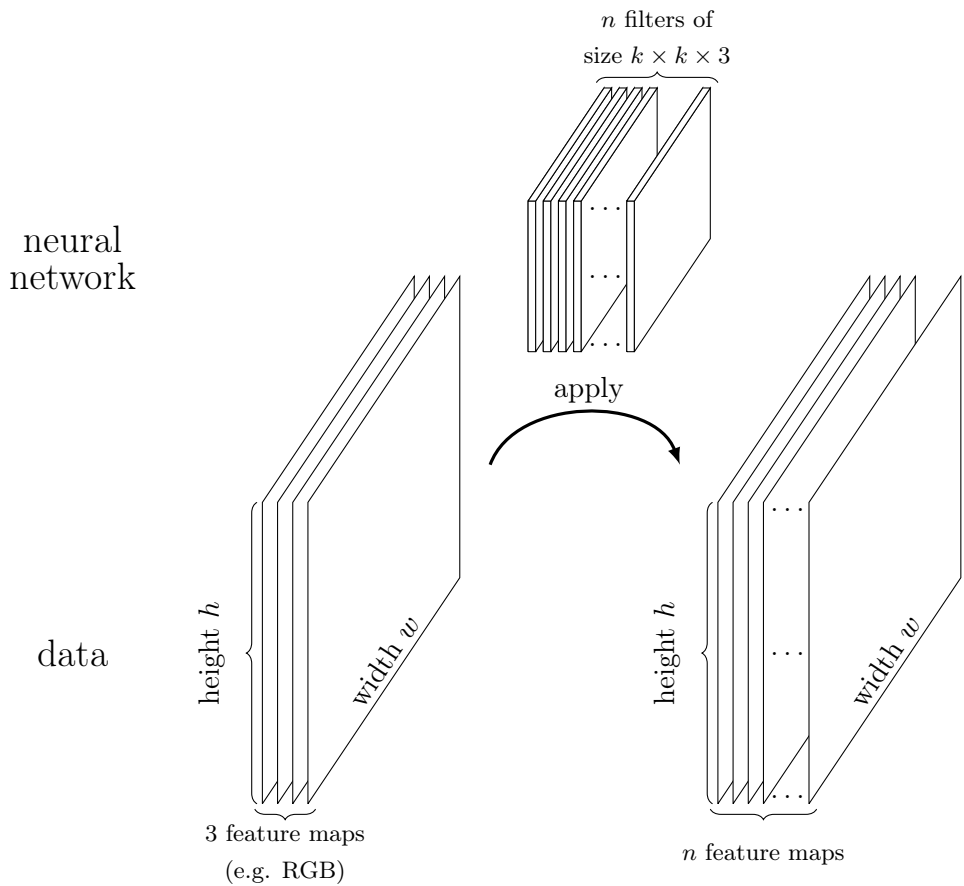

See~\cref{fig:convolution-layer} for a visualization of the application of a
convolutional layer.

A convolutional layer with $n$~filters of size $k_w \times k_h$ and
\texttt{SAME} padding after $d^{(i-1)}$ feature maps of size $s_x \times s_y$
has $n \cdot d^{(i-1)} \cdot (k_w \cdot k_h)$ parameters if no bias is used. In
contrast, a fully connected layer which produces the same output size and does
not use a bias would have $n \cdot d^{(i-1)} \cdot (s_x
\times s_y)^2$ parameters. This means a convolutional layer has drastically
fewer parameters. One the one hand, this means it can learn less complex
decision boundaries. On the other hand, it means fewer parameters have to be
learned and hence the optimization procedure needs fewer examples and the
optimization objective is simpler.

It is particularly interesting to notice that even a convolutional layer of $1
\times 1$ filters does learn a linear combination of the $d$ input feature
maps. This can be used for dimensionality reduction, if there are fewer $1
\times 1$ filters in a convolutional layer than input feature maps.
Another insight recently got important: Every fully connected layer has an
equivalent convolutional layer which has the same weights.\footnote{But
convolutional layers only have equivalent fully connected layers if the output
feature map is $1 \times 1$} This way, one can use the complete classification
network as a very complex non-linear image filter which can be used for
semantic segmentation.

A fully connected layer with $d \in \mathbb{N}_{\geq 1}$ inputs and $n \in
\mathbb{N}_{\geq 1}$ nodes can be interpreted as a convolutional layer with an
input of shape $1 \times 1 \times d$ and $n$ filters of size $1 \times 1$.
This will produce an output shape $1 \times 1 \times n$. Every single output
is connected to all of the inputs.

When a convolutional layer is followed by a fully connected layer, it is
necessary to vectorize to feature maps. If the $1 \times 1$ convolutional
filter layer is applied to the vectorized output, it is completely equivalent
to a fully connected layer. However, the vectorization can be omitted if a
convolution layer without padding and a filter size equal to the feature maps
size is applied. This was used by~\cite{long2015fully}.

\subsection{Pooling Layers}\label{subsec:pooling-layers}
Pooling summarizes a $p \times p$ area of the input feature
map. Just like convolutional layers, pooling can be used with a stride of $s
\in \mathbb{N}_{> 1}$. As $s \geq 2$ is the usual choice, pooling layers are sometimes
also called \textit{subsampling layers}. Typically, $p \in \Set{2, 3, 4, 5}$
and $s = 2$ such as for AlexNet~\cite{AlexNet-2012} and VGG-16~\cite{VGG-16}.

The type of summary for the set of activations $A$ varies between the functions
listed in~\cref{table:pooling-types-overview}, spatial pyramid pooling as
introduced in~\cite{he2014spatial} and generalizing pooling functions
as introduced in~\cite{lee2016generalizing}.

\begin{table}[H]
    \centering
    \begin{tabular}{lll}
    \toprule
    \textbf{Name}          & \textbf{Definition}              & \textbf{Used by} \\\midrule
    Max pooling            & $\max \Set{a \in A}$             & \cite{boureau2010theoretical,AlexNet-2012}\\
    Average / mean pooling & $\frac{1}{|A|} \sum_{a \in A} a$ & LeNet-5~\cite{LeNet-5} and \cite{kavukcuoglu2010learning} \\
    $\ell_2$ pooling       & $\sqrt{\sum_{a \in A} a^2}$      & \cite{le2013building} \\
    Stochastic pooling     & *                                & \cite{zeiler2013stochastic}\\
    \bottomrule
    \end{tabular}
    \caption[Pooling types]{Pooling types for a set $A$ of activations $a \in \mathbb{R}$.\\
             (*) For stochastic pooling, each of the $p \times p$ activation
             values $a_i$ in the pooling region gets picked with probability
             $p_i = \frac{a_i}{\sum_{a_j \in A} a_j}$. This assumes the activations
             $a_i$ are non-negative.}
    \label{table:pooling-types-overview}
\end{table}

Pooling is applied for three~reasons: To get local translational invariance, to
get invariance against minor local changes and, most important, for data reduction to $\frac{1}{s^2}$th of the data by using strides of $s > 1$.

\begin{figure}[ht]
    \centering
    \definecolor{c1}{HTML}{9ACFC6}
\definecolor{c2}{HTML}{DABBD6}
\definecolor{c3}{HTML}{CBDCB9}
\definecolor{c4}{HTML}{9AC3E1}
\definecolor{c5}{HTML}{DEBBA5}
\definecolor{c6}{HTML}{C4DDE5}
\newcommand*{\xMin}{0}%
\newcommand*{\xMax}{6}%
\newcommand*{\yMin}{0}%
\newcommand*{\yMax}{4}%

\newcommand*{\xMinR}{9.5}%
\newcommand*{\xMaxR}{12.5}%
\newcommand*{\yMinR}{1}%
\newcommand*{\yMaxR}{3}%

\begin{tikzpicture}
    \fill [c1] (0, 0) rectangle (2, 2);
    \fill [c2]   (2, 0) rectangle (4, 2);
    \fill [c3]  (4, 0) rectangle (6, 2);
    \fill [c4] (0, 2) rectangle (2, 4);
    \fill [c5]   (2, 2) rectangle (4, 4);
    \fill [c6]   (4, 2) rectangle (6, 4);

    \fill [c1]  (9.5, 1) rectangle (10.5, 2);
    \fill [c2]   (10.5, 1) rectangle (11.5, 2);
    \fill [c3]  (11.5, 1) rectangle (12.5, 2);
    \fill [c4]  (9.5, 2) rectangle (10.5, 3);
    \fill [c5]   (10.5, 2) rectangle (11.5, 3);
    \fill [c6]   (11.5, 2) rectangle (12.5, 3);

    \foreach \i in {\xMin,...,\xMax} {
        \draw [very thin,gray] (\i,\yMin) -- (\i,\yMax)  node [below] at (\i,\yMin) {};
    }
    \foreach \i in {\yMin,...,\yMax} {
        \draw [very thin,gray] (\xMin,\i) -- (\xMax,\i) node [left] at (\xMin,\i) {};
    }

    \foreach \i in {\xMin,2,...,\xMax} {
        \draw [thick,gray] (\i,\yMin) -- (\i,\yMax)  node [below] at (\i,\yMin) {};
    }
    \foreach \i in {\yMin,2,...,\yMax} {
        \draw [thick,gray] (\xMin,\i) -- (\xMax,\i) node [left] at (\xMin,\i) {};
    }
    \node at (0.5, 0.5) {7};
    \node at (1.5, 0.5) {9};
    \node at (2.5, 0.5) {3};
    \node at (3.5, 0.5) {5};
    \node at (4.5, 0.5) {9};
    \node at (5.5, 0.5) {4};
    \node at (0.5, 1.5) {0};
    \node at (1.5, 1.5) {7};
    \node at (2.5, 1.5) {0};
    \node at (3.5, 1.5) {0};
    \node at (4.5, 1.5) {9};
    \node at (5.5, 1.5) {0};
    \node at (0.5, 2.5) {5};
    \node at (1.5, 2.5) {0};
    \node at (2.5, 2.5) {9};
    \node at (3.5, 2.5) {3};
    \node at (4.5, 2.5) {7};
    \node at (5.5, 2.5) {5};
    \node at (0.5, 3.5) {9};
    \node at (1.5, 3.5) {2};
    \node at (2.5, 3.5) {9};
    \node at (3.5, 3.5) {6};
    \node at (4.5, 3.5) {4};
    \node at (5.5, 3.5) {3};

    \draw[draw=black,line width=12pt,-{Latex[length=9mm]}] (6.5, 2)  -- (9,2);
    \node[font=\footnotesize\bfseries] at (7.7, 2.5) {$\mathbf{2\times 2}$ max pooling};

    \foreach \i in {\xMinR,...,\xMaxR} {
        \draw [thick,gray] (\i,\yMinR) -- (\i,\yMaxR)  node [below] at (\i,\yMinR) {};
    }
    \foreach \i in {\yMinR,...,\yMaxR} {
        \draw [thick,gray] (\xMinR,\i) -- (\xMaxR,\i) node [left] at (\xMinR,\i) {};
    }

    \node at (10, 1.5) {9};
    \node at (11, 1.5) {5};
    \node at (12, 1.5) {9};
    \node at (10, 2.5) {9};
    \node at (11, 2.5) {9};
    \node at (12, 2.5) {7};

    \draw [decorate,decoration={brace,amplitude=4pt},xshift=-2pt,yshift=0pt]
(0,2) -- (0,4) node [black,midway,xshift=-0.3cm] {\footnotesize $2$};

    \draw [decorate,decoration={brace,amplitude=4pt},xshift=0pt,yshift=2pt]
(0,4) -- (2,4) node [black,midway,yshift=+0.3cm] {\footnotesize $2$};
\end{tikzpicture}
    \caption[Max pooling]{$2 \times 2$ max pooling applied to a feature map of size $6 \times 4$ with stride $s=2$ and padding.}
    \label{fig:max-pooling}
\end{figure}
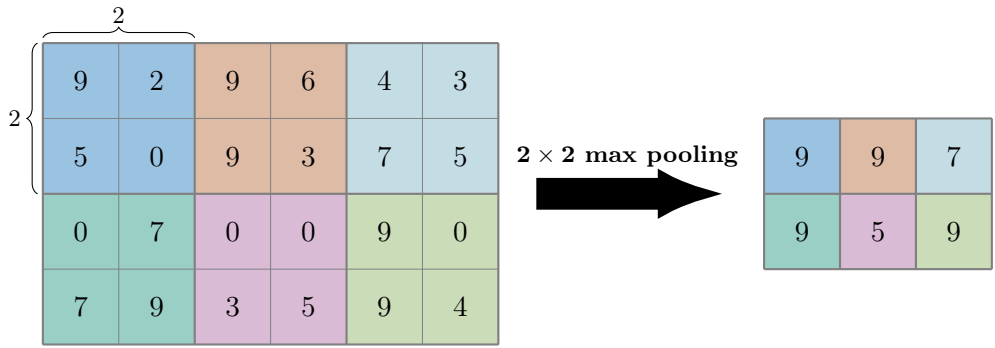

See~\cref{fig:max-pooling} for a visualization of max pooling.

Average pooling of $p \times p$~areas with stride~$s$ can be replaced by a
convolutional layer. If the input of the pooling layer are $d^{(i-1)}$~feature
maps, the convolutional layer has to have $d^{(i-1)}$~filters of size
$p \times p$ and stride~$s$. The $i$th~filter has the values
\[\begin{pmatrix}
    \frac{1}{p^2} & \dots  & \frac{1}{p^2}\\
    \vdots        & \ddots & \vdots\\
    \frac{1}{p^2} & \dots  & \frac{1}{p^2}\\
  \end{pmatrix}\]
for the dimension $i$ and the zero matrix
\[\begin{pmatrix}
    0 & \dots  & 0\\
    \vdots        & \ddots & \vdots\\
    0 & \dots  & 0\\
  \end{pmatrix}\]
for all other dimensions $i = 1, \dots, d^{(i-1)}$.

\subsection{Dropout}\label{sec:dropout}
Dropout is a technique used to prevent overfitting and co-adaptations of
neurons by setting the output of any neuron to zero with probability~$p$. It
was introduced in~\cite{hinton2012-dropout} and is well-described
in~\cite{srivastava2014dropout}.

A Dropout layer can be implemented as follows: For an input \verb+in+ of any
shape~$s$, a tensor of the same shape $D \in \Set{0, 1}^{s}$ is sampled, where
each element $d_i$ is sampled independently from a Bernoulli distribution. The
results are element-wise multiplied to calculate the output \verb+out+ of the
Dropout layer:
\[\text{out} = D \odot \text{in} \qquad\text{with }d_i \sim B(1, p)\]
where $\odot$ is the Hadamard product
\[(A \odot B)_{i,j} := (A)_{i,j} (B)_{i,j}\]

Hence every value of the input gets set to zero with a dropout~probability
of~$p$. Typically, Dropout is used with $p=0.5$. Layers closer to the input
usually have a lower dropout~probability than later layers. In order to keep
the expected output at the same value, the output of a dropout layer is
multiplied with $\frac{1}{1-p}$ when dropout is
enabled~\cite{Lasagne-Dropout,tf-dropout}. At inference time, dropout is
disabled.

Dropout is usually only applied after fully connected layers, but not after
convolutional layers as it usually increases the test error as pointed out
in~\cite{gal2015bayesian}.

Models which use Dropout can be interpreted as an ensemble of models with
different numbers of neurons in each layer, but also with weight sharing.

Conceptually similar are DropConnect and networks with stochastic depth.
DropConnect~\cite{wan2013regularization} is a generalization of Dropout, which
sets weights to zero in contrast to setting the output of a neuron to zero.
Networks with stochastic depth as introduced in~\cite{huang2016deep} dropout
only complete layers. This can be done by having Residual networks which have
one identity connection and one residual feature connection. Hence the residual
features can be dropped out and the identity connection remains.

\subsection{Normalization Layers}\label{sec:batch-normalization}
One problem when training deep neural networks is \textit{internal covariate shift}:
While the parameters of layers close to the output are adapted to some input
produced by lower layers, those lower layers parameters are also adapted. This
leads to the parameters in the upper layers being worse. A very low
learning rate has to be chosen to adjust for the fact that the input features
might drastically change over time.

One way to approach this problem is by normalizing mini-batches as described
in~\cite{BatchNormalization-2015}. A Batch Normalization layer with
$d$-dimensional input $x = (x^{(1)}, \dots, x^{(d)})$ is first normalized
point-wise to

\[\hat{x}^{(k)} = \frac{x^{(k)} - \bar{x}^{(k)}}{\sqrt{s'[x^{(k)}]^2 + \varepsilon}}\]

with $\bar{x}^{(k)} = \frac{1}{m} \sum_{i=1}^m x_i^{(k)}$ being the sample mean
and
$s'[x^{(k)}]^2 = \frac{1}{m} \sum_{i=1}^m (x_i^{(k)} - \bar{x}^{(k)})$ the sample variance where
$m \in \mathbb{N}_{\geq 1}$ is the number of training samples per mini-batch,
$\varepsilon > 0$ being a small constant to prevent division by zero
and $x_i^{(k)}$ is the activation of neuron $k$ for training sample $i$.

Additionally, for each activation $x^{(k)}$ two parameters $\gamma^{(k)},
\beta^{(k)}$ are introduced which scale and shift the feature:
\[y^{(k)} = \gamma^{(k)} \cdot \hat{x}^{(k)} + \beta^{(k)}\]

In the case of fully connected layers, this is applied to the activation,
before the non-linearity is applied. If it is applied after the activation, it
harms the training in early stages. For convolution, only one $\gamma$ and one
$\beta$ is learned per feature map.

One important special case is $\gamma^{(k)} = \sqrt{s'[x^{(k)}]^2 + \varepsilon}$
and $\beta^{(k)} = \bar{x}^{(k)}$, which would make the Batch Normalization layer
an identity layer.

During evaluation time,\footnote{also called \textit{inference time}} the
expected value and the variance are calculated once for the complete dataset.
An unbiased estimate of the empirical variance is used.

The question where Batch Normalization layers (BN) should be applied and for
which reasons is still open. For Dropout, it doesn't matter if it is applied
before or after the activation function. Considering this, the possible options
for the order are:
\begin{enumerate}[noitemsep,nolistsep]
    \item CONV / FC $\rightarrow$ BN $\rightarrow$ activation function  $\rightarrow$ Dropout $\rightarrow$ \dots \label{item:bn-act-dropout}
    \item CONV / FC $\rightarrow$ activation function $\rightarrow$ BN $\rightarrow$ Dropout $\rightarrow$ \dots \label{item:act-bn-dropout}
    \item CONV / FC $\rightarrow$ activation function $\rightarrow$ Dropout $\rightarrow$ BN $\rightarrow$ \dots \label{item:act-dropout-bn}
    \item CONV / FC $\rightarrow$ Dropout $\rightarrow$ BN $\rightarrow$ activation function  $\rightarrow$ \dots \label{item:dropout-bn-act}
\end{enumerate}
The authors of~\cite{BatchNormalization-2015} suggest to use Batch
Normalization before the activation function as
in~\cref{item:bn-act-dropout,item:dropout-bn-act}. Batch Normalization after
the activation lead to better results in
\url{https://github.com/ducha-aiki/caffenet-benchmark/blob/master/batchnorm.md}

Another normalization layer is Local Response Normalization as described
in~\cite{AlexNet-2012}, which includes $\ell_2$ normalization as described
in~\cite{Wang2013}. Those two normalization layers, however, are superseded by
Batch Normalization.

\section{CNN Blocks}\label{sec:cnn-blocks}
This section describes more complex building blocks than simple layers.
CNN~blocks act similar to a layer, but they are themselves composed of layers.

\subsection{Residual Blocks}
Residual blocks as introduced in~\cite{deep-residual-networks-2015} are a
milestone in computer vision. They enabled the computer vision community to go
from about 16~layers as in VGG 16-D (see \cref{subsec:vgg-16-d}) to several
hundred layers. The key idea of deep residual networks (ResNets) as introduced
in~\cite{deep-residual-networks-2015} is to add an identity connection which
skips two layers. This identity connection adds the feature maps onto the other
feature maps and thus requires the output of the input layer of the residual
block to be of the same dimension as last layer of the residual block.

\begin{figure}[ht]
    \centering
    \includegraphics*[width=0.4\linewidth, keepaspectratio]{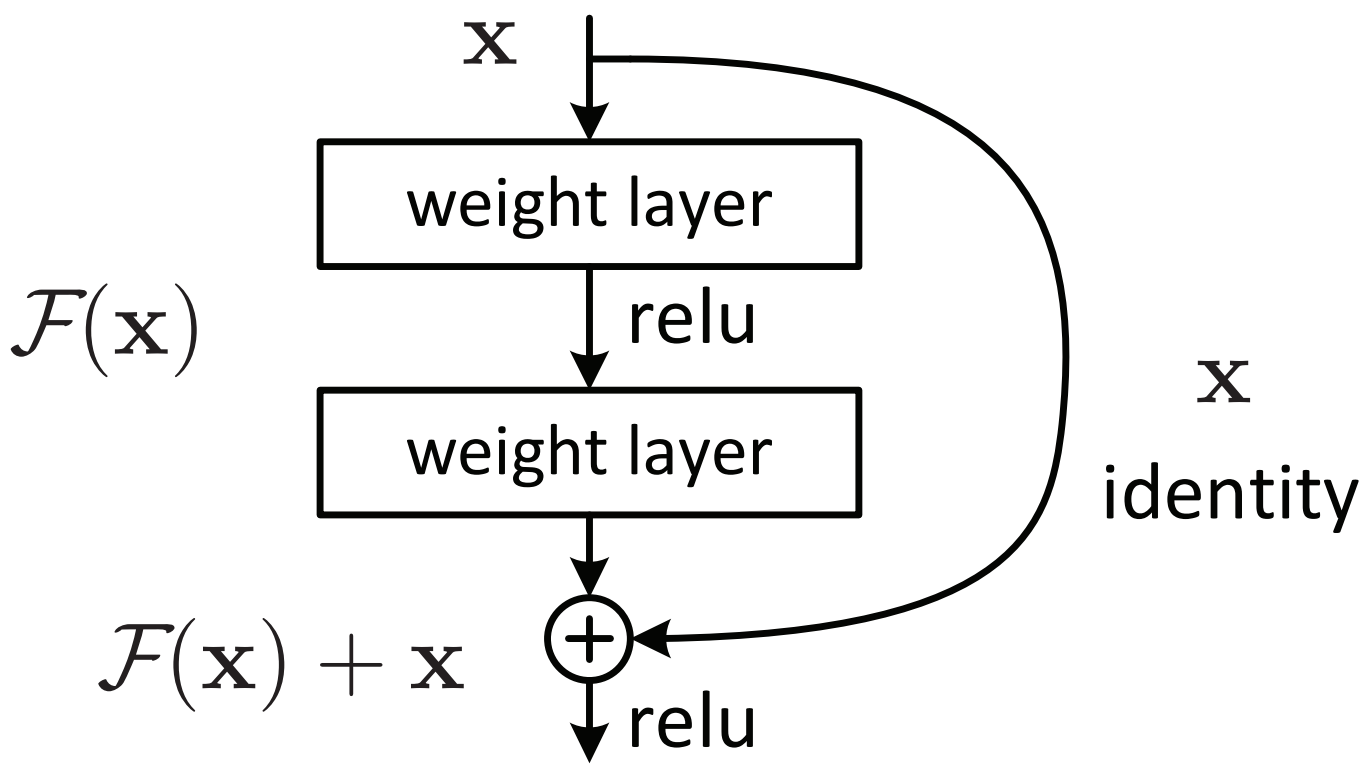}
    \caption[ResNet module]{ResNet module\\Image source:~\cite{deep-residual-networks-2015}}
    \label{fig:ResNet-module}
\end{figure}

Formally, it can be described as follows. If $x_i$ are the feature maps after
layer $i$ and $x_0$ is the input image, $H$ is a non-linear transformation
of feature maps, then
\[y = H(x)\]
describes a traditional \gls{CNN}. Note that this could be multiple layers. A
residual block as visualized in~\cref{fig:ResNet-module} is described by
\[y = H(x) + x\]

In~\cite{deep-residual-networks-2015}, they only used residual skip connections
to skip two layers. Hence, if $\conv_i(x_i)$ describes the application of the
convolutional layer $i$ to the input $x_i$ without the nonlinearity, then such
a residual block is
\[x_{i+2} = \conv{}_{i+1}(\ReLU(\conv{}_i(x_i)))+x_i\]

\cite{hardt2016identity} provides some insights why deep residual networks are
successful.

\subsection{Aggregation Blocks}
\begin{figure}[ht]
    \centering
    \begin{tikzpicture}
    \tikzstyle{conv-layer}=[draw,minimum width=3cm,minimum height=1cm]
    \tikzstyle{arrow}=[->, -Latex, thick]

    \draw[fill=black!5,dashed] (-4.2, -5.4) rectangle (7.5, 1.2);

    \node (input) at (1.5, 1) {};
    \path[arrow] (1.5, 2) edge node[right, midway] {$256$-d in} (input.south);
    \node[draw, dashed] (concatenate) at (1.5, -5) {concatenate};
    \node (between) at (3.4, -2.0) [align=center,text width=1.5cm] {total $32$ groups\\ \dots};
    \path[arrow] (concatenate.south) edge node[right, midway]  {$128$-d out} (1.5, -6);

    \node (rect11) at (-2.5,-1.5) [conv-layer] {4 @ $1 \times 1 \times 256$};
    \node (rect12) at (-2.5,-3.0) [conv-layer] {4 @ $3 \times 3 \times 4$};
    \draw[arrow] (input.south) -- (rect11.north);
    \draw[arrow] (rect11.south) -- (rect12.north);
    \draw[arrow] (rect12.south) -- (concatenate);

    \node (rect21) at ( 1,-1.5) [conv-layer] {4 @ $1 \times 1 \times 256$};
    \node (rect22) at ( 1,-3.0) [conv-layer] {4 @ $3 \times 3 \times 4$};
    \draw[arrow] (input.south) -- (rect21.north);
    \draw[arrow] (rect21.south) -- (rect22.north);
    \draw[arrow] (rect22.south) -- (concatenate);

    \node (rect31) at ( 5.8,-1.5) [conv-layer] {4 @ $1 \times 1 \times 256$};
    \node (rect32) at ( 5.8,-3.0) [conv-layer] {4 @ $3 \times 3 \times 4$};
    \draw[arrow] (input.south) -- (rect31.north);
    \draw[arrow] (rect31.south) -- (rect32.north);
    \draw[arrow] (rect32.south) -- (concatenate);

\end{tikzpicture}
    \caption[Aggregation block]{Aggregation block with a cardinality of
             $C = 32$. Each of the 32
             groups is a 2-layer convolutional network. The first layer
             receives 256 feature maps and applies four $1 \times 1$ filters
             to it. The second layer applies four $3 \times 3$ filters.
             Although every group has the same topology, the learned weights
             are different. The outputs of the groups are concatenated.}
    \label{fig:aggregation-block}
\end{figure}
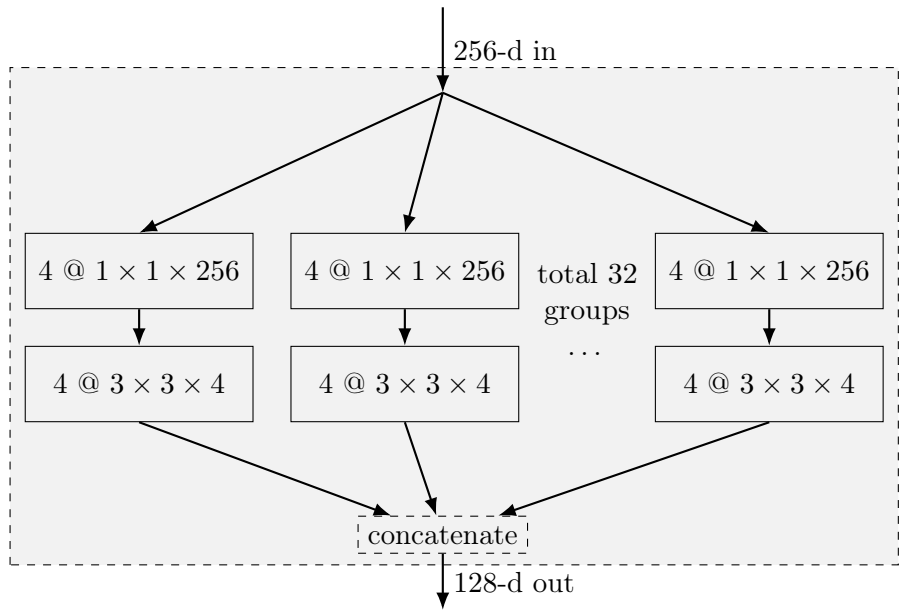

Two common ways to add more parameters to neural networks are increasing their
depth by adding more layers or increasing their width by adding more neurons /
filters. Inception blocks~\cite{AlexanderMordvintsev2015} implicitly started a
new idea which was explicitly described in~\cite{xie2016aggregated} as
\enquote{ResNeXt block}: Increasing the cardinality~$C \in \mathbb{N}_{\geq
1}$. By cardinality, the authors describe the concept of having $C$ small
convolutional networks with the same topology but different weights. This
concept is visualized in~\cref{fig:aggregation-block}. Please note that
\cref{fig:aggregation-block} does not combine aggregation blocks with residual
blocks as the authors did.

The hyperparameters of an aggregation block are:
\begin{itemize}
    \item The topology of the group members.
    \item The cardinality $C \in \mathbb{N}_{\geq 1}$. Note that a cardinality
          of $C=1$ is equivalent in every aspect to using the group network
          without an aggregation block.
\end{itemize}
\clearpage

\subsection{Dense Blocks}\label{sec:dense-blocks}
Dense blocks are collections of convolutional layers which are introduced
in~\cite{huang2016densely}. The idea is to connect each convolutional layer
directly to subsequent convolutional layers. Traditional \glspl{CNN} with
$L$~layers and one input layer have $L$ connections between layers, but dense
blocks have $\frac{L(L+1)}{2}$ connections between layers. The input feature
maps are concatenated in depth. According to the authors, this prevents
features from being re-learned and allows much fewer filters per convolutional
layer. Where AlexNet and VGG-16 have several hundred filters per convolutional
layer (see~\cref{table:AlexNet-architecture,table:VGG-16-D-architecture}), the
authors used only on the order of 12~feature maps per layer.

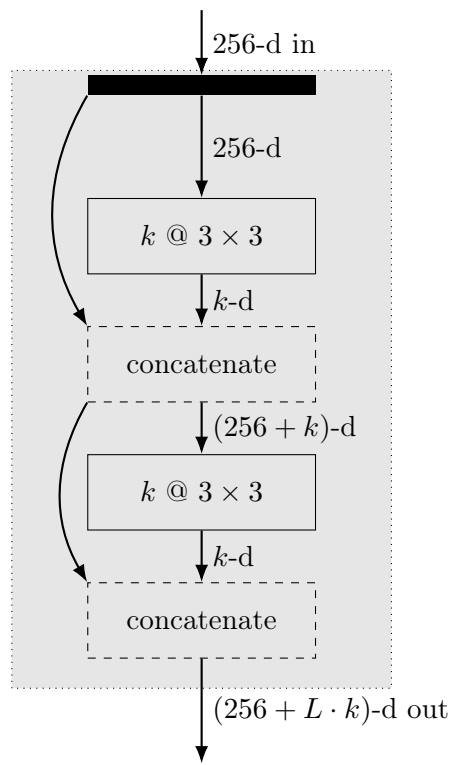
\begin{figure}[ht]
    \centering
    \begin{tikzpicture}
    \tikzstyle{conv-layer}=[draw,minimum width=3cm,minimum height=1cm]
    \tikzstyle{arrow}=[->, -Latex, thick]

    \draw[fill=black!10,dotted] (-2.5, -7.0) rectangle (2.5, 1.2);

    \node[rectangle, fill=black,minimum width=3cm] (input) at (0, 1) {};
    \path[arrow] (0, 2) edge node[right, midway] {$256$-d in} (input.north);

    \node (conv1) at (0,-1.0) [conv-layer] {$k$ @ $3 \times 3$};
    \node (conc1) at (0,-2.7) [conv-layer, dashed] {concatenate};
    \node (conv2) at (0,-4.4) [conv-layer] {$k$ @ $3 \times 3$};
    \node (conc2) at (0,-6.1) [conv-layer, dashed] {concatenate};

    \path[arrow] (input.south) edge node[right, midway] {$256$-d} (conv1.north);
    \path[arrow] (conv1.south) edge node[right, midway] {$k$-d} (conc1.north);
     \path[arrow] (conc1.south) edge node[right, midway] {$(256 + k)$-d} (conv2.north);
    \path[arrow] (conv2.south) edge node[right, midway] {$k$-d} (conc2.north);

    \path[arrow] (input.south west) edge[bend right] node[right, midway] {} (conc1.north west);
    \path[arrow] (conc1.south west) edge[bend right] node[right, midway] {} (conc2.north west);

    \path[arrow] (conc2.south) edge node[right] {$(256+L \cdot k)$-d out} (0, -8);

\end{tikzpicture}
    \caption[Dense block]{Dense block with $L=2$ layers and a growth factor of $k$.}
    \label{fig:dense-block}
\end{figure}

A dense block is visualized in~\cref{fig:dense-block}.

Dense block have five~hyperparameters:
\begin{itemize}
    \item The activation function being used. The authors use \gls{ReLU}.
    \item The size $k_w \times k_h$ of filters. The authors use $k_w = k_h = 3$.
    \item The number of layers $L$, where $L=2$ is a simple convolutional layer.
    \item The number $k$ of filters added per layer (called \textit{growth rate} in the paper)
\end{itemize}

It might be necessary use $1 \times 1$ convolutions to reduce the number of $L
\cdot k$ feature maps.

\section{Transition Layers}\label{sec:transition-layers}
Transition layers are used to overcome constraints imposed by resource
limitations or architectural design choices. One constraint is the number of
feature maps (see~\cref{sec:memory-footprint} for details). In order to reduce
the number of feature maps while still keeping as much relevant information as
possible in the network, a convolutional layer $i$ with $k_i$~filters of the
shape $1 \times 1 \times k_{i-1}$ is added. The number of filters~$k_i$
directly controls the number of generated feature maps.

In order to reduce the dimensionality (width and height) of the feature maps,
one typically applies pooling.

Global pooling is another type of transition layer. It applies pooling over
the complete feature map size to shrink the input to a constant $1 \times 1$
feature map and hence allows one network to have different input sizes.

\clearpage
\section{Analysis Techniques}\label{sec:analyzation-techniques}
\glspl{CNN} have dozens of hyperparameters and ways to tune them. Although
there are automatic methods like random search~\cite{bergstra2012random}, grid
search~\cite{EfficientBackprop}, gradient-based hyperparameter
optimization~\cite{maclaurin2015gradient} and Hyperband~\cite{li2016hyperband}
some actions need a manual investigation to improve the model's quality. For
this reason, analysis techniques which guide developers and researchers to the
important hyperparameters are necessary. In the following, nine~diagnostic
techniques are explained.

A machine learning developer has the following choices to improve the model's
quality:
\begin{enumerate}[label=(I\arabic*)]
    \item Change the problem definition (e.g., the classes which are to be distinguished)
    \item Get more training data
    \item Clean the training data
    \item Change the preprocessing (see~\cref{sec:preprocessing})
    \item Augment the training data set (see~\cref{sec:data-augmentation})
    \item Change the training setup (see~\cref{sec:initialization,sec:objective-function,sec:optimization-techniques})
    \item Change the model (see~\cref{sec:network-design,sec:regularization})
\end{enumerate}

The preprocessing is usually not changed in modern architectures. However, this
still leaves six very different ways to improve the classifier. Changing the
training setup and the model each have too many possible choices to explore
them completely. Thus, techniques are necessary to guide the developer to
changes which are most promising to improve the model.

For all of the following methods, it is important to use only the training set
and the validation set.

\subsection{Qualitative Analysis by Example}
The most basic analysis technique which should always be used is looking at
examples which the network correctly predicted with a high certainty and what
the classifier got wrong with a high certainty. Those examples can be arranged
by applying t-SNE~\cite{maaten2008visualizing}.

One the one hand, this might reveal errors in the training data. Most of the
time, training data is manually labeled by humans who make mistakes. If a
model is fit to those errors, its quality decreases.

On the other hand, this can show differences in the distribution of validation
data which are not covered by the training set and thus indicate the need to
collect more data.

\subsection{Confusion Matrices}\label{subsec:confusion-matrices}
A \textit{confusion matrix} is a matrix $(c)_{ij} \in \mathbb{N}_{\geq 0}^{K \times K}$,
where $K \in \mathbb{N}_{\geq 2}$ is the number of classes, which contains all
correct and wrong classifications. The item $c_{ij}$ is the number of times
items of class $i$ were classified as class $j$. This means the correct
classification is on the diagonal $c_{ii}$ and all wrong classifications are of
the diagonal. The sum $\sum_{i=1}^K \sum_{j=1}^K c_{ij}$ is the total number of
samples which were evaluated and $\frac{\sum_{i=1} c_{ii}}{\sum_{i=1}^K
\sum_{j=1}^K c_{ij}}$ is the accuracy.

The sums $r(i) = \sum_{j=1}^K c_{ij}$ of each class~$i$ are worth being
investigated as they show if the classes are skewed. If the number of samples
of one class dominates the data set, then the classifier can get a high
accuracy by simply always prediction the most common class. If the accuracy of
the classifier is close to the a priory probability of the most common class,
techniques to deal with skewed classes might help.

An automatic criterion to check for this problem is
\[\text{accuracy} \leq \frac{\max(\Set{r(i) | i = 1, \dots, k})}{\sum_{i=1}^k r(i)} + \varepsilon\]
where $\varepsilon$ is a small value to compensate the fact that some examples
might be correct just by chance.

Other values which should be checked are the class-wise sensitivities:
\[s(k) = \frac{\text{\# correctly identified instances of class } k}{\text{\# instances of class }k} = \frac{c_{kk}}{r(k)} \in [0, 1]\]
If $s(i)$ is much lower than $s(j)$, it is an indicator that
more or cleaner training data is necessary for $s(i)$.

The class-wise confusion
\[f_\text{confusability}(k_1, k_2) = \frac{c_{k_1 k_2}}{\sum_{j = 1}^K c_{k_1 j}}\]
indicates if class $k_1$ gets often classified as class $k_2$. The highest
values here can indicate if two classes should be merged or a specialized model
for separating those classes could improve the overall system.

\subsection{Validation Curves: Accuracy, loss and other metrics}
Validation curves display a hyperparameter (e.g., the training epoch) on the
horizontal axis and a quality metric on the vertical axis. Accuracy,
$\text{error} = (1 - \text{accuracy})$ or loss are typical quality metrics.
Other quality metrics can be found in~\cite{ortigosa2016towards}.

\begin{figure}[h]
    \centering
    \definecolor{c1}{HTML}{0072B2}
\definecolor{c2}{HTML}{009E73}
\begin{tikzpicture}
    \tikzstyle{training}=[c1, thick,samples=200,dashed]
    \tikzstyle{testing}=[c2, thick,samples=200]
    \begin{axis}[
        legend pos=north east,
        legend cell align=left,
        axis x line=middle,
        axis y line=middle,
        grid = major,
        width=14cm,
        height=8cm,
        grid style={dashed, gray!30},
        xmin=0,       
        xmax= 104,    
        ymin=0,       
        ymax= 0.98,   
        axis background/.style={fill=white},
        xlabel=Epochs,
        ylabel=Error,
        y label style={at={(-0.1,1.0)}},
        tick align=outside,
        minor tick num=-3,
        tension=0.08]
      \addplot[domain=2:50, training] {(x-50)^2/2700 + 0.1};
      \addplot[domain=4:50, testing] {(x-50)^2/2700 + 0.2};

      \addplot[domain=50:100, training] {0.1};
      \addplot[domain=50:100, testing] {(x-50)^2/10000 + 0.2};

      \draw[dashed, very thick] (axis cs:50,0.0) -- (axis cs:50,0.3);
      \draw[decoration={text along path, text={overfitting}, text align={center}}, decorate] (axis cs:51,0.16) -- (axis cs:90,0.16);
      \draw[-{latex[scale=3.0]}, very thick] (axis cs:51,0.15) -- (axis cs:90,0.15);

      \addlegendentry{Training set}
      \addlegendentry{Validation set}
    \end{axis}
\end{tikzpicture}
    \caption[Validation curve]{A typical validation curve: In this case, the
             hyperparameter is the number of epochs and the quality metric is
             the error $(1 - \text{accuracy})$. The longer the network is
             trained, the better it gets on the training set. At some point the
             network is fit too well to the training data and loses its
             capability to generalize. At this point the quality curve of the
             training set and the validation set diverge. While the classifier
             is still improving on the training set, it gets worse on the
             validation and the test set.}
    \label{fig:training-progress-curve}
\end{figure}
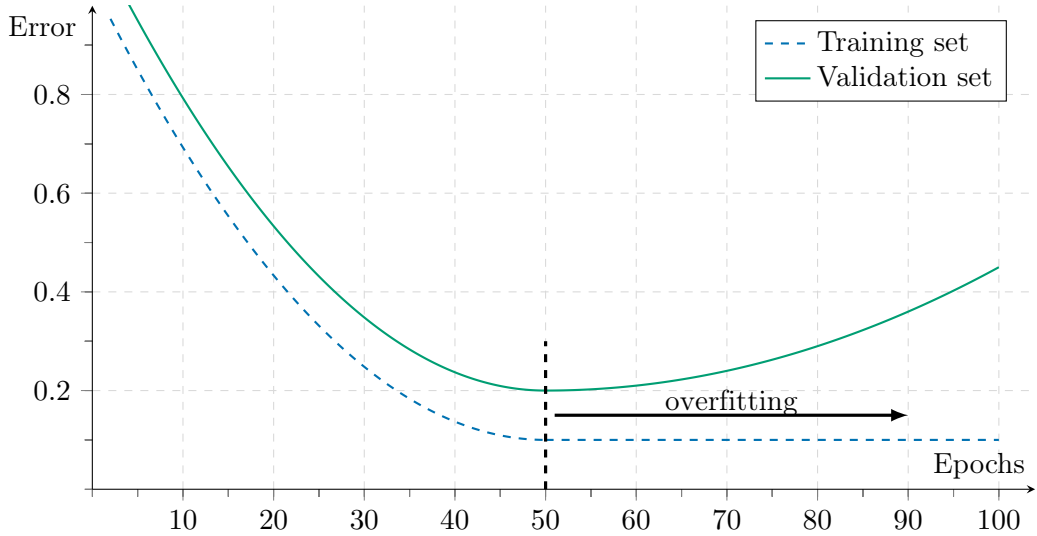

In case that the number of training epochs are used as the examined
hyperparameter, validation curves give an indicator if training longer improves
the model's performance. By plotting the error on the training set as well as
the error on a validation set, one can also estimate if overfitting might
become a problem. See \cref{fig:training-progress-curve} for an example.

\begin{figure}[h]
    \centering
    \includegraphics*[width=\linewidth, keepaspectratio]{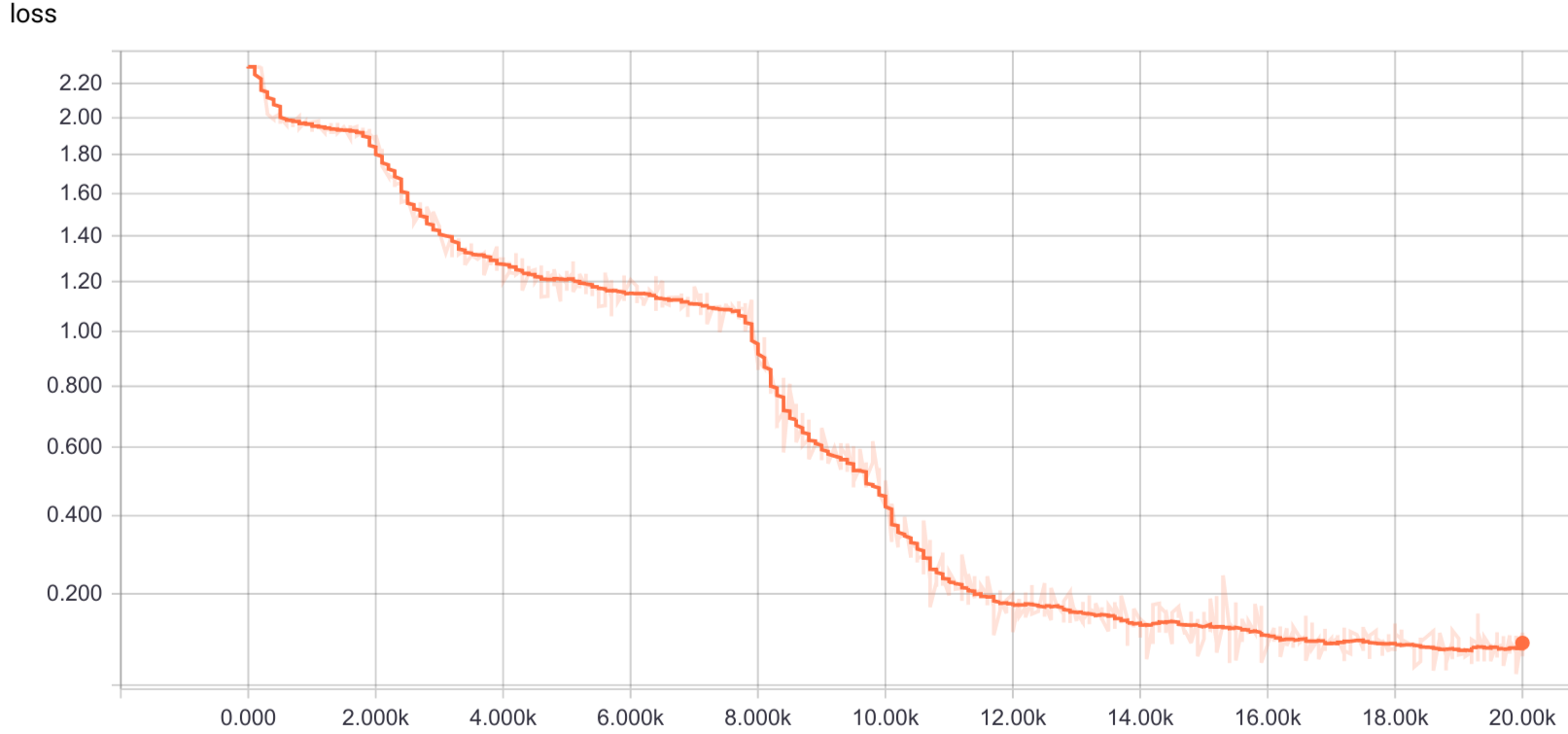}
    \caption[Validation curve with plateaus]{Example for a validation curve
             (plotted loss function) with plateaus. The dark orange curve is
             smoothed, but the non-smoothed curve is also plotted in light
             orange.}
    \label{fig:loss-plateau}
\end{figure}

When the epoch-loss validation curve has plateaus as
in~\cref{fig:loss-plateau}, this means the optimization process did not improve
for several epochs. Three possible ways to reduce the problem of plateaus are
\begin{enumerate*}[label=(\roman*)]
    \item to change weight initialization if the plateau was at the beginning,
    \item regularizing the model or
    \item changing the optimization algorithm.
\end{enumerate*}

\subsubsection{Loss functions}
The loss function (also called \textit{error function} or \textit{cost
function}) is a function which assigns a real value to a complex event like the
predicted class of a feature vector. It is used to define the \textit{objective
function}. For classification problems the loss function is typically
cross-entropy with $\ell_1$ or $\ell_2$ regularization, as it was described
in~\cite{nowlan1992simplifying}:
\[E_{CE}(W) = \underbrace{- \sum_{x \in X} \sum_{k=1}^K \left [t_k^x \log(o_k^x) + (1-t_k^x) \log(1-o_k^x) \right ]}_{\text{cross-entropy data loss}} + \underbrace{\lambda_1 \cdot \overbrace{\sum_{w \in W} |w|}^{\ell_1} + \lambda_2 \cdot \overbrace{\sum_{w \in W} w^2}^{\ell_2}}_{\text{model complexity loss}}\]
where $W$ are the weights, $X$ is the training data set, $K \in \mathbb{N}_{\geq}
0$ is the number of classes and $t_k^x$ indicates if the training example $x$ is of
class $k$. $o_k^x$ is the output of the classification algorithm which depends
on the weights. $\lambda_1, \lambda_2 \in [0, \infty)$ weights the
regularization and is typically smaller than $0.1$.

The data loss is positive whenever the classification is not correct, whereas
the model complexity loss is higher for more complex models. The model
complexity loss exists due to the intuition of \textit{Occam's razor}: If two
models explain the same data with an accuracy of \SI{100}{\percent}, the simpler
model is to be preferred.

A reason to show the loss for the validation curve technique instead of
other quality metrics is that it contains more information about the quality of
the model. A reason against the loss is that it has no upper bound like the
accuracy and can be hard to interpret. The loss only shows relative learning
progress whereas the accuracy shows absolute progress to human readers.

There are three observations in the loss validation curve which can help to
improve the network:
\begin{itemize}
    \item If the loss does not decrease for several epochs, the learning
          rate might be too low. The optimization process might also be stuck
          in a local minimum.
    \item Loss being NAN might be due to too high learning rates. Another
          reason is division by zero or taking the logarithm of zero. In both
          cases, adding a small constant like $10^{-7}$ fixes the problem.
    \item If the loss-epoch validation curve has a plateau at the beginning,
          the weight initialization might be bad.
\end{itemize}

\subsubsection{Quality criteria}\label{subsubsec:quality-criteria}
There are several quality criteria for classification models. Most quality
criteria are based the confusion matrix $c$ which denotes at $c_{ij}$ the number of times the
real class was $i$ and $j$ was predicted. This means the diagonal contains the
number of correct predictions. For the following, let $t_i = \sum_{j=1}^k c_{ij}$ be the number
of training samples for class $i$. The most common quality criterion is
accuracy:
\[\text{accuracy}(c) = \frac{\sum_{i=1}^k c_{ii}}{\sum_{i=1}^k t_i}  \in [0, 1]\]

One problem of accuracy as a quality criterion are skewed classes. If one class
is by far more common than all other classes, then the simplest way to achieve
a high score is to always classify everything as the most common class.

In order to fix this problem, one can use the mean accuracy:
\[\text{mean-accuracy}(c) = \frac{1}{k} \cdot \sum_{i=1}^k \frac{c_{ii}}{t_i} \in [0, 1]\]

For two-class problems there are many other metrics like precision, recall and
$F_\beta$-score. Quality criteria for semantic segmentation are explained
in~\cite{Thoma:2016}.

Besides the quality of the classification result, several other quality
criteria are important in practice:
\begin{itemize}
    \item Speed of evaluation for new images,
    \item latency,
    \item power consumption,
    \item robustness against (non)random perturbations in the training data (see~\cite{szegedy2013intriguing,papernot2015distillation}),
    \item robustness against (non)random perturbations in the training labels (see~\cite{natarajan2013learning,xiao2012adversarial}),
    \item model size
\end{itemize}

As reducing the floating point accuracy allows to process more data on a given
device~\cite{Harris2015}, analysis under this aspect is also highly relevant in
some scenarios.

However, the following focuses on the quality of the classification result.

\subsection{Learning Curves}
\begin{figure}[h]
    \centering
    \definecolor{c1}{HTML}{0072B2}
\definecolor{c2}{HTML}{009E73}
\begin{tikzpicture}
    \tikzstyle{training}=[c1, thick, samples=200,dashed]
    \tikzstyle{testing}=[c2, thick, samples=200]
    \tikzstyle{arrow}=[<->, Latex-Latex, thick]
    \begin{axis}[
        legend pos=north east,
        legend cell align=left,
        axis x line=middle,
        axis y line=middle,
        grid = major,
        width=14cm,
        height=8cm,
        grid style={dashed, white}, 
        xmin=0,       
        xmax= 104,    
        ymin= 0,      
        ymax= 0.79,   
        axis background/.style={fill=white},
        xlabel=Training samples,
        ylabel=Error,
        tick align=outside,
        minor tick num=-3,
        tension=0.08]
      \addplot[domain=1:100, testing] {1/(x+2)+0.4};
      \addplot[domain=1:100, training] {0.4-(1/(x+2))};

      \draw[dashed,thick] (axis cs:0,0.2) -- (axis cs:110,0.2);
      \path[arrow] (axis cs:80,0.2) edge node[right, midway] {avoidable bias} (axis cs:80,0.3878);
      \path[arrow] (axis cs:20,0.3545) edge node[left, midway] {variance} (axis cs:20,0.4454);
      \draw[decoration={text along path, text={human-level error}, text align={center}}, decorate] (axis cs:20,0.21) -- (axis cs:50,0.21);

      \addlegendentry{Validation set}
      \addlegendentry{Training set}
    \end{axis}
\end{tikzpicture}
    \caption[Learning curve]{A typical learning curve: The more data is used
             for training, the more errors a given architecture will make to
             fit the given training data. At the same time, it is expected that
             the training data gets more similar to the true distribution of
             the data which should be captured by the test data. At some point,
             the error on the training and test set should be about the same.
             The term \enquote{avoidable bias} was coined by Andrew Ng~\cite{ng2016}.
             In some cases it is not possible to classify data correctly by the
             given features. If humans can classify the data given the features
             correctly, however, then the bias is avoidable by building a
             better classifier.}
    \label{fig:learning-curve}
\end{figure}
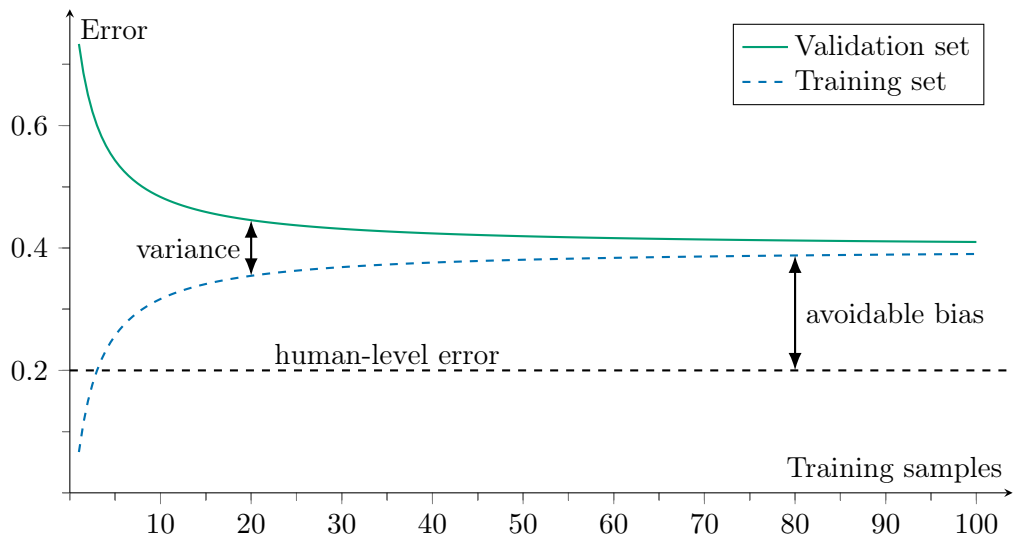

A learning curve is a plot where the horizontal axis displays the number of
training samples given to the network and the vertical axis displays the error.
Two curves are plotted: The error on the training set (of which the size is given by
the horizontal axis) and the error on the test set (which is of fixed size).
See \cref{fig:learning-curve} for an example. The learning curve for the
validation set is an indicator if more training data without any other changes
will improve the networks performance. Having the training set's learning
curve, it is possible to
estimate if the capacity of the model to fit the data is high enough for the
desired classification error. The error on the validation set should never be
expected to be significantly lower than the error on the training set. If the
error on the training set is too high, then more data will \emph{not} help.
Instead, the model or the training algorithm need to be adjusted.

If the training set's learning curve is significantly higher than the validation
set's learning curve, then removing features (e.g., by decreasing the images
resolution), more training samples or more regularization will help.

The major drawback of this analysis technique is its computational
intensity. In order to get one point on the training curve and one point on the
testing curve, a complete training has to be executed. On the full data set,
this can be several days on high-end computers.

\subsection{Input-feature based model explanations}
Understanding which clues the model took to come to its prediction is crucial
to check if the model actually learns what the developer thinks it learns. For
example, a model which has to distinguish sled dogs from Chihuahuas might
simply look at the background and check if there is snow. Depending on the
training and test data, this works exceptionally well. However, it is not the
desired solution.

For classification problems in computer vision, there are two types of
visualizations which help to diagnose such problems. Both color superpixels of
the original image to convey information how the model used those superpixels:
\begin{itemize}[noitemsep]
    \item \textbf{Correct class heatmap}: The probability of the correct class
          is encoded to give a heat map which superpixels are important for the
          correct class. This can also be done by setting the opacity
          accordingly.
    \item \textbf{Most-likely class image}: Each of the most likely classes for
          all superpixels is represented by a color. The colored image thus
          gives clues why different predictions were assigned a high
          probability.
\end{itemize}

Two methods to generate such images are explained in the following.

\subsubsection{Occlusion Sensitivity Analysis}
Occlusion sensitivity analysis is described in~\cite{zeiler2014visualizing}.
The idea is to occlude a part of the image by something. This
could be a gray square as in~\cite{zeiler2014visualizing} or a black superpixel
as in~\cite{ribeiro2016should}. Then the classifier is run on the image again.
This is done for each region (e.g., superpixel or position of the square) and
the regions are then colored to generate either a correct class heatmap of the
most-likely class image. It is important to note that the color at region $r_i$
denotes the result if $r_i$ is occluded.

\begin{figure}[h]
    \centering
    \includegraphics*[width=\linewidth, keepaspectratio]{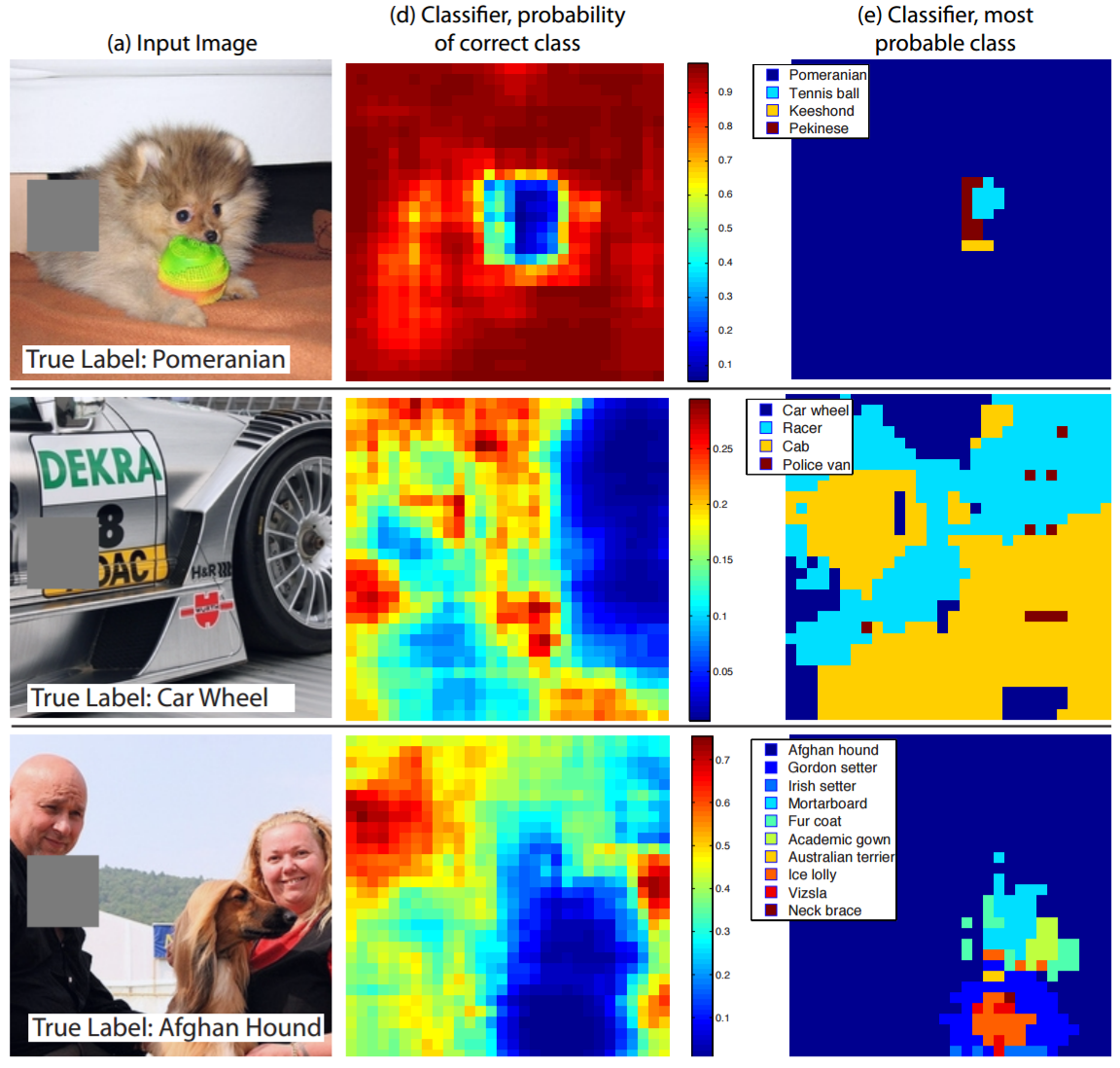}
    \caption[Occlusion analysis]{Occlusion sensitivity analysis
             by~\cite{zeiler2014visualizing}: The left column shows three
             example images, where a gray square occluded a part of the image.
             This gray squares center $(x, y)$ was moved over the complete
             image and the classifier was run on each of the occluded images.
             The probability of the correct class, depending on the gray
             squares position, is showed in the middle column. One can see that
             the predicted probability of the correct class
             \enquote{Pomeranian} drops if the face of the dog is occluded. The
             last image gives the class with the highest predicted probability.
             In the case of the Pomeranian, it always predicts the correct
             class if the head is visible. However, if the head of the dog is
             occluded, it predicts other classes.}
    \label{fig:zeiler-occlusion-sensitivity-analysis}
\end{figure}

Both visualizations are shown
in~\cref{fig:zeiler-occlusion-sensitivity-analysis}. One can see that the
network makes sensible predictions for this image of the class
\enquote{Pomeranian}. However, the image of the class \enquote{Afghan Hound}
gets confused with \enquote{Ice lolly}, which is a sign that this needs further
investigation.

\subsubsection{Gradient-based approaches}
In~\cite{simonyan2013deep}, a gradient-based approach was used to generate
\textit{image-specific class saliency maps}. The authors describe the problem
as a ranking problem, where each pixel of the image $I_0$ is assigned a score
$S_c(I_0)$ for a class~$c$ of interest. \Glspl{CNN} are non-linear
functions, but they can be approximated by the first order Taylor expansion
$S_c(I) \approx w^T I + b$ where $w$ is the derivative of $S_c$ at $I_0$.

\subsection{Argmax Method}
The \textit{argmax method} has two variants:
\begin{itemize}
    \item \textbf{Fixed class argmax}: Propagate all elements of a given class
          through the network and analyze which neurons are activated most
          often / have the highest activation.
    \item \textbf{Fixed neuron argmax}: Propagate the data through the network
          and find the $n$ data elements which cause the highest activation
          for a given neuron.
\end{itemize}

Note that a \enquote{neuron} is a filter in a \gls{CNN}. The amount of
activation of a filter $F$ by an image $I$ is calculated by applying $F$ to $I$
and calculating the element-wise sum of the result.

Fixed-neuron argmax was applied in~\cite{zeiler2014visualizing}. However, they
did not stop with that. Besides showing the 9~images which caused the highest
activation, they also trained a deconvolutional neural network to project the
activation of the filter back into pixel space.

The fixed neuron argmax can be used qualitatively to get an impression of the
kind of features which are learned. This is useful to diagnose problems, for
example in~\cite{AlexanderMordvintsev2015} it is described that the network
recognized the class \enquote{dumbbell} only if a hand was present, too.

Fixed neuron argmax can also be used quantitatively to estimate the amount of
parameters being shared between classes or how many parameters are mainly
assigned to which classes.

Going one step further from the fixed neuron argmax method is using an
optimization algorithm to change an initial image minimally in such a way that
any desired class gets predicted. This is called \textit{caricaturization}
in~\cite{mahendran2016visualizing}.

\subsection{Feature Map Reconstructions}
\begin{figure}[h]
    \centering
    \includegraphics*[width=\linewidth, keepaspectratio]{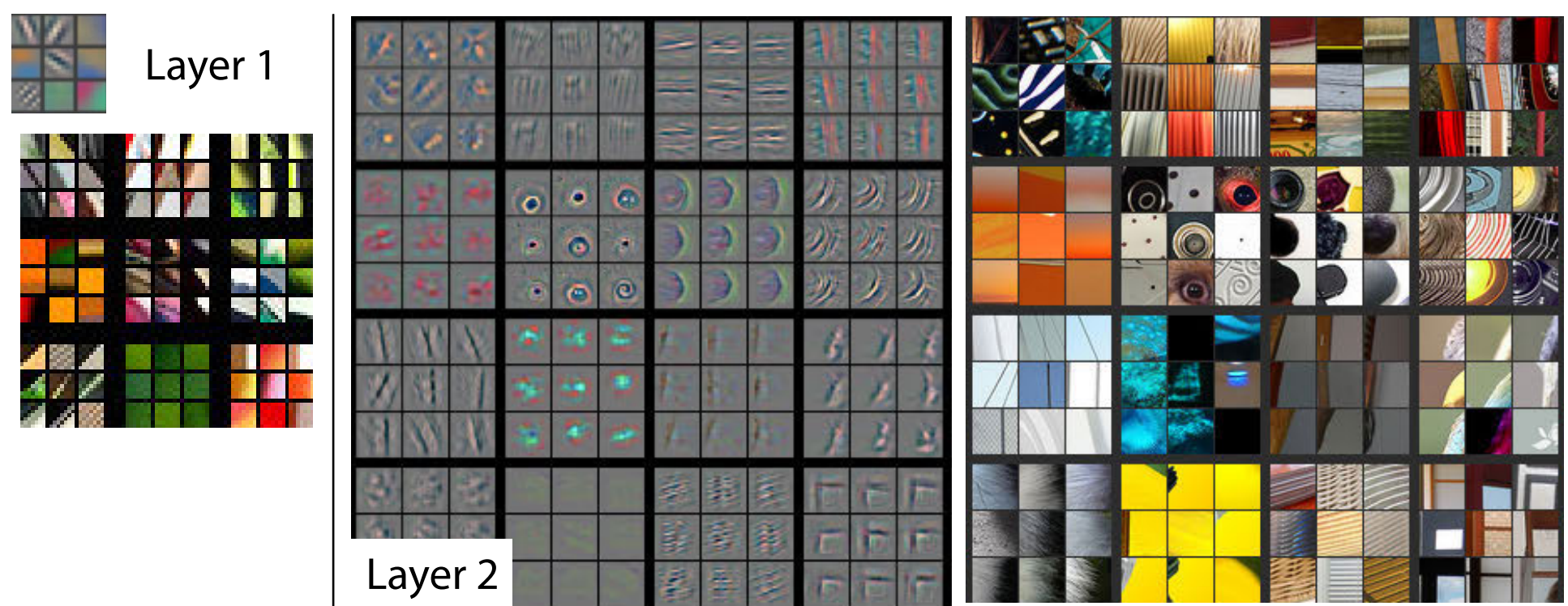}
    \caption[Filter visualization]{Filter visualization
    from~\cite{zeiler2014visualizing}: The filters themselves as well as the
    input feature maps which caused the highest activation are displayed.}
    \label{fig:zeiler-filter-vis}
\end{figure}

Feature map visualizations such as the ones made in~\cite{zeiler2014visualizing}
(see~\cref{fig:zeiler-filter-vis})
give insights into the learned features. This
shows what the network emphasizes. However, it is not necessarily the case that
the feature maps allow direct and easy conclusions about the learned features.
This technique is called \textit{inversion} in~\cite{mahendran2016visualizing}.

A key idea of feature map visualizations is to reconstruct a layers input,
given its activation. This makes it possible find which inputs would cause
neurons to activate with extremely high or low values.

More recent work like~\cite{nguyen2016multifaceted} tries to make the
reconstructions appearance look more natural.

\subsection{Filter comparison}
One question which might lead to some insight is how robust the features are
which are learned. If the same network is trained with the same data, but
different weight initializations, the learned weights should still be comparable.

If the set of learned filters changes with initialization, this might be an
indicator for too little capacity of that layer. Hence adding more filters to
that layer could improve the performance.

Filters can be compared with the $k$-translation correlation as introduced
in~\cite{zhai2016doubly}:
\[\rho_k(\mathbf{W_i,W_j})=\max_{(x,y)\in \{-k,...,k\}^2\setminus(0,0)} \frac{\langle\mathbf{W_i}, T(\mathbf{W_j}, x,y)\rangle_f}{\left \|  \mathbf{W_i}\right \|_2 \left \|  \mathbf{W_j}\right \|_2}\, \in [-1, 1],\]
where $T(\cdot, x, y)$ denotes the translation of the first operand by $(x,
y)$, with zero padding at the borders to keep the shape. $\langle \cdot, \cdot
\rangle_f$ denotes the flattened inner product, where the two operands are
flattened into column vectors before applying the standard inner product. The
closer the absolute value of the $k$-translation correlation to one, the
more similar two filters $W_i, W_j$ are. According to~\cite{zhai2016doubly},
standard \glspl{CNN} like AlexNet (see~\cref{subsec:AlexNet}) and VGG-16
(see~\cref{subsec:vgg-16-d}) have many filters which are highly correlated. They
found this by comparing the \textit{averaged maximum $k$-translational correlation}
of the networks with Gaussian-distributed initialized filters. The averaged
maximum $k$-translational correlation is defined as
\[\bar{\rho}_k(\mathbf{W}) = \frac{1}{N}\sum_{i=1}^N \max_{j=1, j \neq i}^N \rho_k(\mathbf{W}_i, \mathbf{W}_j)\]
where $N$ is the number of filters in the layer $\mathbf{W}$ and $\mathbf{W}_i$
denotes the $i$th filter.

\subsection{Weight update tracking}
Andrej Karpathy proposed in the 5th lecture of CS231n to track weight updates
to check if the learning rate is well-chosen. He suggests that the weight
update should be in the order of $10^{-3}$. If the weight update is too high,
then the learning rate has to be decreased. If the weight update is too low,
then the learning rate has to be increased.

The order of the weight updates as well as possible implications highly depend
on the model and the training algorithm. See~\cref{sec:optimization-techniques}
for a short overview of training algorithms for neural networks.

\section{Accuracy boosting techniques}
There are techniques which can almost always be applied to improve accuracy
of \gls{CNN} classifiers:
\begin{itemize}
    \item Ensembles~\cite{ciregan2012multi}
    \item Training-time augmentation (see~\cref{sec:data-augmentation})
    \item Test-time transformations~\cite{dieleman2016exploiting,howard2013some,he2015delving}
    \item Pre-training and fine-tuning~\cite{zhang2014part,girshick2014rich}
\end{itemize}

One of the most simple ensemble techniques which was introduced
in~\cite{ciregan2012multi} is averaging the prediction of $n$~classifiers. This
improves the accuracy even if the classifiers use exactly the same training
setup by reducing variance.

Data augmentation techniques give the optimizer the possibility to take
invariances like rotation into account by generating artificial training
samples from real training samples. Data augmentation hence reduces bias and
variance with no cost at inference time.

Data augmentation at inference time reduces the variance of the classifier.
Similar to using an ensemble, it increases the computational cost of inference.

Pretraining the classifier on another dataset to obtain start from a good
position or finetuning a model which was originally created for another task
is also a common technique.

\chapter{Topology Learning}\label{ch:topology-learning}
The topology of a neural network is crucial for the number of parameters, the
number of \glspl{FLOP}, the required memory, as well as the features being
learned. The choice of the topology, however, is still mainly done by
trial-and-error.

This chapter introduces three general approaches to automatic topology
learning: Growing a networks from a minimal network
in~\cref{sec:growing-approaches}, pruning in \cref{sec:pruning-approaches},
genetic approaches in~\cref{sec:genetic-approaches} and reinforcement learning
approaches in~\cref{sec:rl}.

\section{Growing approaches}\label{sec:growing-approaches}
Growing approaches for topology learning start with a minimal network, which
only has the necessary number of input nodes and the number of output nodes
which are determined by the application and the features of the input. They
then apply a criterion to insert new layers / neurons into the network.

In the following, Cascade-Correlation, Meiosis Networks and Automatic Structure
Optimization are introduced.

\subsection{Cascade-Correlation}
Cascade-Correlation was introduced in~\cite{fahlman1989cascade}. It generates a
cascading architecture which is similar to dense block described
in~\cref{sec:dense-blocks}.

Cascade-Correlation works as follows:
\begin{enumerate}
    \item \textbf{Initialization}: The number of input nodes and the number of output nodes are defined
          by the problem. Create a minimal, fully connected network for those.
    \item \textbf{Training}: Train the network until the error no longer
          decreases.
    \item \textbf{Candidate Generation}: Generate candidate nodes. Each
          candidate node is connected to all inputs. They are not connected to
          other candidate nodes and not connected to the output nodes.
    \item \textbf{Correlation Maximization}: Train the weights of the
          candidates by maximizing $S$, the correlation between candidates
          output value~$V$ with the networks residual error:
          \[S = \sum_{o \in O} \left |\sum_{p \in T} \left (V_p - \bar{V} \right) (E_{p,o} - \bar{E_o}) \right |\]
          where $O$ is the set of output nodes, $T$ is the training set, $V_p$
          is the candidate neurons activation for a training pattern $p$.
          $E_{p, o}$ is the residual output error at node $o$ for pattern $p$.
          $\bar{V}$ and $\bar{E_{o}}$ are averaged values over all elements of
          $T$. This step is finished when the correlation no longer increases.
    \item \textbf{Candidate selection}: Keep the candidate node with the
          highest correlation, freeze its incoming weights and add connections
          to the output nodes.
    \item \textbf{Continue}: If the error is higher than desired, continue
          with step~2.
\end{enumerate}

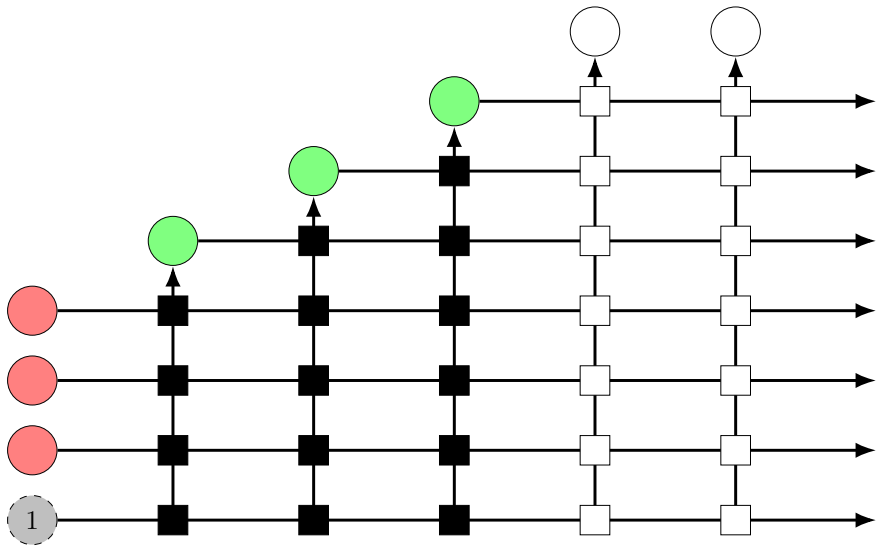
\begin{figure}[H]
    \centering
    \resizebox {0.85\columnwidth} {!} {
    \tikzset{sigmoid/.style={path picture= {
    \begin{scope}[x=1pt,y=10pt]
      \draw plot[domain=-7:7] (\x,{1/(1 + exp(-\x))-0.5});
    \end{scope}
    }
  }
}

\tikzset{relu/.style={path picture= {
    \begin{scope}[x=7pt,y=6pt]
      \draw plot[domain=-1:0] (\x,0);
      \draw plot[domain=0:1] (\x,\x);
    \end{scope}
    }
  }
}

\tikzstyle{input}=[draw,fill=red!50,circle,minimum size=20pt,inner sep=0pt]
\tikzstyle{hidden}=[draw,fill=green!50,circle,minimum size=20pt,inner sep=0pt]
\tikzstyle{output}=[draw,fill=white,circle,minimum size=20pt,inner sep=0pt]
\tikzstyle{bias}=[draw,dashed,fill=gray!50,circle,minimum size=20pt,inner sep=0pt]
\tikzstyle{layer}=[fill=gray!70]

\tikzstyle{stateTransition}=[->, very thick, -Latex]

\begin{tikzpicture}[scale=2]

    \node (i0)[bias]  at (0, 0) {1};
    \node (i1)[input] at (0, 0.5) {};
    \node (i2)[input] at (0, 1.0) {};
    \node (i3)[input] at (0, 1.5) {};

    \node (h1)[hidden] at (1, 2.0) {};
    \node (h2)[hidden] at (2, 2.5) {};
    \node (h3)[hidden] at (3, 3.0) {};

    \node (o1)[output] at (4, 3.5) {};
    \node (o2)[output] at (5, 3.5) {};

    \draw[stateTransition] (i0) -- (6, 0);
    \draw[stateTransition] (i1) -- (6, 0.5);
    \draw[stateTransition] (i2) -- (6, 1.0);
    \draw[stateTransition] (i3) -- (6, 1.5);
    \draw[stateTransition] (h1) -- (6, 2.0);
    \draw[stateTransition] (h2) -- (6, 2.5);
    \draw[stateTransition] (h3) -- (6, 3.0);

    \draw[stateTransition] (1, 0) -- (h1);
    \draw[stateTransition] (2, 0) -- (h2);
    \draw[stateTransition] (3, 0) -- (h3);

    \draw[stateTransition] (4, 0) -- (o1);
    \draw[stateTransition] (5, 0) -- (o2);

    \foreach \x in {1,...,3} {
        \pgfmathsetmacro{\limy}{(2+\x) / 2}
        \foreach \y in {0, 0.5, ..., \limy} {
            \node[rectangle,draw=black,fill=black,minimum size=12] (box) at (\x,\y) {};
        }
    }

    \foreach \x in {4,...,5} {
        \foreach \y in {0, ..., 6} {
            \pgfmathsetmacro{\yn}{0.5*\y}
            \node[rectangle,draw=black,fill=white,minimum size=12] (box) at (\x,\yn) {};
        }
    }

\end{tikzpicture}
    }
    \caption[Cascade-correlation network]{A Cascade-Correlation network with
             three input nodes (red) and one bias node (gray) to the left,
             three hidden nodes (green) in the middle and two output nodes in
             the upper right corner. The black squares represent frozen weights
             which are found by correlation maximization whereas the white
             squares are trainable weights.}
    \label{fig:cascade-correlation}
\end{figure}

One network with three hidden nodes trained by Cascade-Correlation is shown
in~\cref{fig:cascade-correlation}.

\subsection{Meiosis Networks}
Meiosis Networks are introduced in~\cite{hanson1989meiosis}. In contrast to
most \glspl{MLP} and \glspl{CNN}, where weights are deterministic and fixed at
prediction time, each weight $w_{ij}$ in Meiosis networks follows a normal
distribution:
\[w_{ij} \sim \mathcal{N}(\mu_{ij}, \sigma^2_{ij})\]

Hence every connection has two learned parameters: $\mu_{ij}$ and $\sigma^2_{ij}$.

The key idea of Meiosis networks is to allow neurons to perform Meiosis, which
is cell division. A node~$j$ is splitted, when the random part dominates the
value of the sampled weights:
\[\frac{\sum_i \sigma_{ij}}{\sum_i \mu_{ij}} > 1 \text{ and } \frac{\sum_k \sigma_{jk}}{\sum_k \mu_{jk}} > 1\]

The mean of the new nodes is sampled around the old mean, half the variance is
assigned to the new connections.

Hence Meiosis networks only change the number of neurons per layer. They do not
add layers or add skip connections.

\subsection{Automatic Structure Optimization}
\Gls{ASO} was introduced in~\cite{Bodenhausen1993} for the task of
on-line handwriting recognition. It makes use of the confusion matrix~$C = (c_{ij}) \in \mathbb{N}_{\geq 0}^{k \times k}$
(see~\cref{subsec:confusion-matrices}) to guide the topology learning. They
define a confusion-symmetry matrix $S$ with $s_ij = s_ji = c_{ij} \cdot c_{ji}$.
The maximum of $S$ defines where the \gls{ASO} algorithm adds more parameters.
The details how the resources are added are not transferable to \glspl{CNN}.

\section{Pruning approaches}\label{sec:pruning-approaches}
Pruning approaches start with a network which is bigger than necessary and
prune it. The motivation to prune a network which has the desired accuracy is
to save storage for easier model sharing, memory for easier deployment and
\glspl{FLOP} to reduce inference time and energy consumption. Especially for
embedded systems, deployment is a challenge and low energy consumption is
important.

Pruning generally works as follows:
\begin{enumerate}
    \item Train a given network until a reasonable solution is obtained,
    \item prune weights according to a pruning criterion and
    \item retrain the pruned network.
\end{enumerate}
This procedure can be repeated.

One family of pruning criterions uses the \emph{Hessian matrix}. For example,
\Gls{OBD} as introduced in~\cite{lecun1989optimal}. For every single parameter
$k$, \gls{OBD} calculates the effect on the objective function of deleting~$k$.
The authors call the effect of the deletion of parameter $k$ the saliency
$s_k$. The parameters with the lowest saliency are deleted, which means they
are set to~0 and are not updated anymore.

A follow-up method called \textit{Optimal Brain
Surgeon}~\cite{hassibi1993optimal} claims to choose the weights in a much better
way. This requires, however, to calculate the inverse Hessian matrix $H^{-1}
\in \mathbb{R}^{n \times n}$ where $n \in \mathbb{N}$ is typically $n > 10^6$.

A much simpler and computationally cheaper pruning criterion is the
\emph{weight magnitude}. \cite{han2015learning} prunes all weights $w$ which
are below a threshold $\theta$:
\[w \gets \begin{cases}w & \text{if } w \geq \theta\\0&\text{otherwise}\end{cases}\]

\section{Genetic approaches}\label{sec:genetic-approaches}
The general idea of \glspl{GA} is to encode the solution space as genes, which
can recombine themselves via crossover and inversion. An introduction to such
algorithms is given in~\cite{eiben2003introduction}.

Commonly used techniques to generate neural networks by \glspl{GA} are
NEAT~\cite{stanley2002evolving} and its successors
HyperNEAT~\cite{stanley2009hypercube} and ES-HyperNEAT~\cite{risi2010evolving}.

The results, however, are of unacceptable quality: On MNIST
(see~\cref{ch:datasets}), where random chance gives
\SI{10}{\percent} accuracy, even simple topologies trained with \gls{SGD}
achieve about \SI{92}{\percent} accuracy~\cite{TF-MNIST-2016} and state of the
art is \SI{99.79}{\percent}~\cite{wan2013regularization}, the HyperNEAT
algorithm achieves only \SI{23.9}{\percent}
accuracy~\cite{verbancsics2013generative}.

Kocm{\'a}nek shows in~\cite{kocmanek2015hyperneat} that HyperNEAT approaches
can achieve \SI{96.47}{\percent} accuracy on MNIST\@. Kocm{\'a}nek mentions that
HyperNEAT becomes slower with each hidden layer so that not more than three
hidden layers could be trained. At the same time, VGG-19~\cite{VGG-16} already
has 19~hidden layers and ResNets are successfully trained with 1202~layers
in~\cite{deep-residual-networks-2015}.

\cite{LingxiXie2017} shows that Genetic algorithms can achieve competitive
results on MNIST and SVHN, but the best results on CIFAR-10 were
\SI{7.10}{\percent} error whereas the state of the art is at
\SI{3.74}{\percent}~\cite{huang2016densely}. Similarly, the Genetic algorithm
achieves \SI{29.03}{\percent} error on CIFAR-100, but the state of the art is
\SI{17.18}{\percent}~\cite{huang2016densely}.

\section{Reinforcement Learning}\label{sec:rl}
Reinforcement learning is a sub-field of machine learning, which focuses on the
question how to choose actions that lead to high rewards.

One can think of the search for good neural network topologies as a
reinforcement learning problem. The agent is a recurrent neural network which
can generate bitstrings. Those variable-length bitstrings encode neural network
topologies.

In 2016, this approach was applied to construct neural networks for computer
vision. In~\cite{baker2016designing}, Q-learning with an $\varepsilon$-greedy
exploration was applied.

In~\cite{zoph2016neural}, the \texttt{REINFORCE} algorithm
from~\cite{williams1992simple} was used to train state of the art models for
CIFAR-10 and the Penn Treebank dataset. A drawback of this method is that
enormous amounts of computational resources were used to obtain those results.

\section{Convolutional Neural Fabrics}
Convolutional Neural Fabrics are introduced in~\cite{saxena2016convolutional}.
They side-step hard decisions about topologies by learning an ensemble of
different \gls{CNN} architectures. The idea is to define a single
architecture as a trellis through a 3D~grid of nodes. Each node represents
a convolutional layer. One dimension is the index of the layer, the other two
dimensions are the amount of filters and the feature size. Each node is
connected to nine other nodes and thus represents nine possible choices of
convolutional layers:
\begin{itemize}
    \item \textbf{Resolution}:
    \begin{enumerate*}[label=(\roman*)]
        \item convolution with \texttt{stride=1} or
        \item convolution with \texttt{stride=2} or
        \item deconvolution (doubling the resolution)
    \end{enumerate*}
    \item \textbf{Channels}:
    \begin{enumerate*}[label=(\roman*)]
        \item half the number of filters than the layer before
        \item the same number of filters as the layer before
        \item double the number of filters than the layer before
    \end{enumerate*}
\end{itemize}

They always use ReLU as an activation function and they always use filters of
size $3 \times 3$. They don't use pooling at all.

%

\chapter{Hierarchical Classification}\label{ch:hierarchical-classification}
Designing a classifier for a new dataset is hard for two main reasons: Many
design choices are not clearly superior to others and evaluating one design
choice takes much time. Especially \glspl{CNN} are known to take several
days~\cite{AlexNet-2012,GoogleNet-Inception} or even weeks~\cite{VGG-16} to
train. Additionally, some methods for analyzing a dataset become harder to use
with more classes and more training samples. Examples are t-SNE, the manual
inspection of errors and confusion matrices, and the argmax method.

\begin{figure}[ht]
    \centering
    \includegraphics*[width=\linewidth, keepaspectratio]{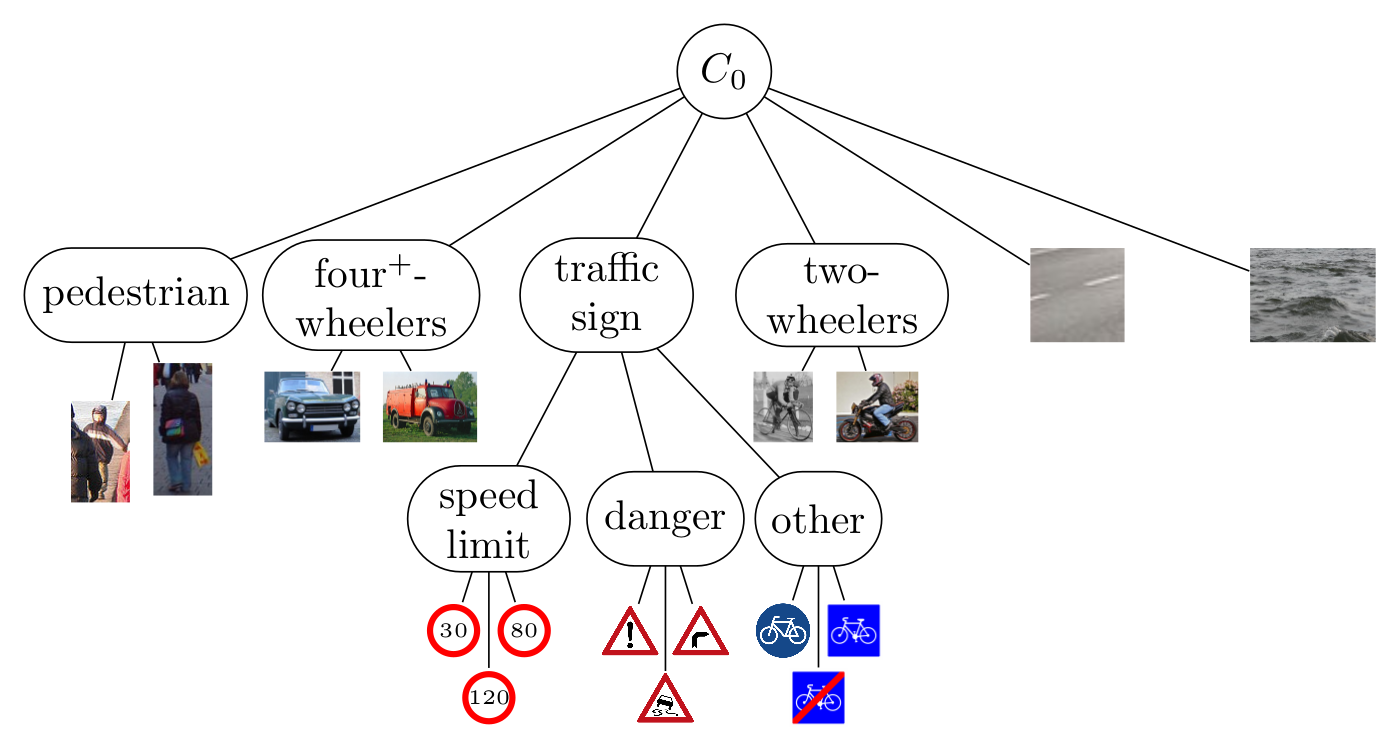}
    \caption[Class Tree]{Example for a hierarchy of classifiers. Each
             classifier is visualized by a rounded rectangle. The root
             classifier $C_0$ has to distinguish six coarse classes
             (pedestrian, four$^+$-wheelers, traffic signs, two-wheelers,
             street, other) or 17~fine-grained classes. If $C_0$ predicts a
             \texttt{pedestrian}, another classifier has to predict if it is an
             adult or a child. Similar, if $C_0$ predicts \texttt{traffic
             sign}, then another classifier has to predict if it is a speed
             limit, a sign indicating danger or something else. If $C_0$,
             however, predicts \texttt{road}, then no other classifier will
             become active.\\
             In this example, the problem has 17~classes. The hierarchical
             approach introduces 7~clusters of classes and thus uses
             8~classifiers.}
    \label{fig:hierarchical-classifier}
\end{figure}

One idea to approach this problem is by building a hierarchy of classifiers.
The root classifier distinguishes clusters of classes, whereas the leaf
classifiers distinguish single classes. \Cref{fig:hierarchical-classifier}
gives an example for an hierarchy of classifiers.

Such a hierarchy of classifiers needs clusters of classes.

\section{Advantages of classifier hierarchies}
Having a classifier hierarchy has five advantages:
\begin{itemize}
    \item \textbf{Division of labor}: Different teams can work together.
          Instead of having a monolithic task, the solutions can be combined.
    \item \textbf{Guarantees}: Changing a classifier will only change the
          prediction of itself and its children. Siblings are not affected.
          In the example from~\cref{fig:hierarchical-classifier}, the classifier
          which distinguishes traffic signs can be changed while the classification
          as \texttt{pedestrian}, \texttt{four$^+$-wheelers}, \texttt{traffic sign},
          \texttt{street}, \texttt{other} will not be affected. Also, the
          classification between speed limits, danger signs and other signs
          will not change.
    \item \textbf{Faster training}: Except for the root classifier $C_0$, each
          other classifier will have less than the total amount of training
          data. Depending on the combined classes, the models could also be
          simpler. Hence the training time is reduced.
    \item \textbf{Weighting of errors}: In practice, some errors are more
          severe than others. For example, it could be acceptable if the
          \texttt{two-wheelers} classifier has an error rate of
          \SI{40}{\percent}. But it is not acceptable if the
          \texttt{speed limit} classifier has such a high error rate.
    \item \textbf{Post-hoc explanations}: The simpler a model is, the easier it
          is to explain why a classification is made the way it is made.
\end{itemize}

\section{Clustering classes}\label{sec:clustering-classes}
There are two ways to cluster classes: By similarity or by semantics. While
semantic clustering needs either additional information or manual work, the
similarity can be automatically inferred from the data. As pointed out
in~\cite{xiao2014error}, semantically similar classes are often also visually
similar. For example, in the ImageNet dataset most dogs are semantically and
visually more similar to each other than to non-dogs. An example where this is
obviously not the case are symbols: The summation symbol \verb+\sum+ is
identical in appearance to the Greek letter \verb+\Sigma+, but semantically
much closer to the addition operator \texttt{+}.

One approach to cluster classes by similarity is to train a classifier and
examine its predictions. Each class is represented in the confusion matrix by
one row. Those rows can be directly with standard clustering algorithms such as
$k$-means, DBSCAN~\cite{ester1996density-dbscan},
OPTICS~\cite{ankerst1999optics}, CLARANS~\cite{ng2002clarans},
DIANA~\cite{kaufman2009finding-diana}, AHC (see~\cite{han2011data}) or spectral
clustering as in~\cite{xiao2014error}. Those clusterings, however, are hard to
interpret and most of them do not allow a human to improve the found clustering
manually.

The confusion matrix $(c)_{ij} \in \mathbb{N}^{k \times k}$ states how often
class~$i$ was present and class~$j$ was predicted. The more often this
confusion happens, the more similar those two classes are to the classifier.
Based on the confusion matrix, the classes can be clustered as explained in the
following.

\cite{huh2016makes} indicates that more classes make it easier to generalize,
but the accuracy gains diminish after a critical point of classes is reached.
Hence a binary tree might not be a good choice. As an alternative, an
approach which allows building arbitrary many clusters, is proposed.

The proposed algorithm has two main ideas:
\begin{itemize}
    \item The order of columns and rows in the confusion matrix is arbitrary.
          This means one can swap rows and columns. If row~$i$ and $j$ are
          swapped, then the columns~$i$ and $j$ have to be swapped to in order
          to keep the same confusion matrix.
    \item If two classes are confused often, then they are similar to the
          classifier.
\end{itemize}

Hence the order of the classes is permutated in such a way that the highest
errors are close to the diagonal. One possible objective function to be
minimized is
\begin{equation}
f(C) = \sum_{i = 1}^n \sum_{j=1}^n C_{ij} \cdot {|i-j|}\label{eq:dist}
\end{equation}
which punishes errors linearly with the distance to the diagonal. This method
is called~\gls{CMO} in the following.

As pointed out by Tobias Ribizel (personal communication), this optimization
problem is a weighted version of \textit{Optimal Linear Arrangement problem}.
That problem is NP-complete~\cite{garey2002computers,garey1976some}. Simulated
Annealing as described in~\cref{alg:simulated-annealing}, however, produces
reasonable clusterings as well as visually appealing confusion matrices. The
algorithm works as follows: First, decide with probability~$0.5$ if only two
random rows are swapped or a block is swapped. If two rows are swapped, choose
both of them randomly. If a block is swapped, then choose the start randomly
and the end of the block randomly after the start. The insert position has to
be a valid position considering the block length, but besides that it is also
chosen uniformly random.

Simple row-swapping can exploit local improvements. For example, in the context
of ImageNet, it can swap the dog-class \texttt{Silky Terrier} to the dog-class
\texttt{Yorkshire terrier} and both dog classes \texttt{Dalmatian} and
\texttt{Greyhound} next to each other. Both the two clusters of dog breeds
could be separated by \texttt{car} and \texttt{bus} due to random chance.
Moving any single class increases the score, but moving either one of the dog
breed clusters or the vehicle cluster decreases the score. Hence it is
beneficial to implement block moving.

One advantage of permutating the classes in order to minimize~\cref{eq:dist} in
comparison to spectral clustering as used in~\cite{xiao2014error} is that the
adjusted confusion matrix can be split into many much smaller matrices along
the diagonal. In the case of many classes (e.g., 1000~classes of ImageNet or
369~classes of HASYv2) this permutation makes it possible to visualize the
types of errors made. If the errors are systematic due to visual similarity,
many confusions are not made and thus many elements of the confusion matrix are
close to~0. Those will be moved to the corners of the confusion matrix by
optimizing~\cref{eq:dist}.

Once a permutation of the classes is found which has a low
score~\cref{eq:dist}, the clusters can either be made by hand by deciding why
classes should not be in one clusters. With such a permutation, only $n-1$
binary decisions have to be made and hence only the list of classes has to be
read. Alternatively, one can calculate the confusions $C_{i, i+1}' + C_{i+1,
i}'$ for each pair of classes which are neighbors in the confusion matrix. The
higher this value, the more similar are the classes according to the
classifier. Hence a threshold~$\theta$ can be applied. $\theta$ can either be
set automatically (e.g., such that
\SI{10}{\percent} of all pairs are above the threshold) or semi-automatically
by asking the user for information if two classes belong to the same cluster.
Such an approach only needs $\log(n)$ binary decisions from the user where $n$
is the number of classes.

Please note that \gls{CMO} only works if the classifier is neither too bad
nor too good. A classifier which does not solve the task at all might just
give almost uniform predictions whereas the confusion matrix of an extremely
good classifier is almost diagonal and thus contains no information about the
similarity of classes. One possible solution to this problem is to take the
prediction of the class in contrast to using only the argmax in order to find
a useful permutation.

\chapter{Experimental Evaluation}\label{ch:experimental-evaluation}
All experiments are implemented using Keras~2.0~\cite{chollet2015keras} with
Tensorflow~1.0~\cite{abadi2016tensorflow} and cuDNN~5.1~\cite{chetlur2014cudnn}
as the backend. The experiments were run on different machines with different
Nvidia \glspl{GPU}, including the Titan~Black, GeForce GTX~970 and
GeForce~940MX\@.

The GTSRB~\cite{Stallkamp2012GTSRB}, SVHN~\cite{YuvalNetzer2011}, CIFAR-10 and
CIFAR-100~\cite{CIFAR-10}, MNIST~\cite{YannLeCun1998},
HASYv2~\cite{thoma2017hasyv2}, STL-10~\cite{coates2010analysis} dataset are
used for the evaluation. Those datasets are used as their size is small enough
to be trained within a day. Other classification datasets which were considered
are listed in~\cref{ch:datasets}.

\textbf{CIFAR-10} (Canadian Institute for Advanced Research 10) is a
10-class dataset of color images of the size $\SI{32}{\pixel} \times
\SI{32}{\pixel}$. Its ten classes are airplane, automobile, bird, cat, deer,
dog, frog, horse, ship, truck. The state of the art achieves an accuracy of
\SI{96.54}{\percent}~\cite{huang2016densely}. According to~\cite{Karpathy2011},
human accuracy is at about~\SI{94}{\percent}.

\textbf{CIFAR-100} is a
100-class dataset of color images of the size $\SI{32}{\pixel} \times
\SI{32}{\pixel}$. Its 100 classes are grouped to 20~superclasses. It includes
animals, people, plants, outdoor scenes, vehicles and other items. CIFAR-100 is
not a superset of CIFAR-10, as CIFAR-100 does not contain the class
\texttt{airplane}. The state of the art achieves an accuracy of
\SI{82.82}{\percent}~\cite{huang2016densely}.

\textbf{GTSRB} (German Traffic Sign Recognition Benchmark) is a 43-class
dataset of traffic~signs. The \num{51839}~images are in color and of a minimum
size of  $\SI{25}{\pixel} \times \SI{25}{\pixel}$ up to $\SI{266}{\pixel}
\times \SI{232}{\pixel}$. The state of the art achieves \SI{99.46}{\percent}
accuracy with an ensemble of 25~\glspl{CNN}~\cite{6033589}. According
to~\cite{JohannesStallkamp}, human performance is at \SI{98.84}{\percent}.

\textbf{HASYv2} (Handwritten Symbols version 2) is a 369~class dataset of
black-and-white images of the size $\SI{32}{\pixel} \times
\SI{32}{\pixel}$. The 369~classes contain the Latin and Greek letters, arrows,
mathematical symbols. The state of the art achieves an accuracy of
\SI{82.00}{\percent}~\cite{thoma2017hasyv2}.

\textbf{STL-10} (self-taught learning 10) is a 10-class dataset of color images
of the size $\SI{96}{\pixel} \times \SI{96}{\pixel}$. Its ten classes are
airplane, bird, car, cat, deer, dog, horse, monkey, ship, truck. The state of
the art achieves an accuracy of \SI{74.80}{\percent}~\cite{zhao2015stacked}. It
contains \num{100000} unlabeled images for unsupervised training and
\num{500}~images per class for supervised training.

\textbf{SVHN} (Street View House Numbers) exists in two formats. For the
following experiments, the cropped digit format was used. It contains the
10~digits cropped from photos of Google Street View. The images are in color
and of size $\SI{32}{\pixel} \times \SI{32}{\pixel}$. The state of the art
achieves an accuracy of \SI{98.41}{\percent}~\cite{huang2016densely}. According
to~\cite{netzer2011reading}, human performance is at~\SI{98.0}{\percent}.

As a preprocessing step, the pixel-features were divided by 255 to obtain
values in $[0, 1]$. For GTSRB, the training and test data was scaled to
$\SI{32}{\pixel} \times \SI{32}{\pixel}$.

\section{Baseline Model and Training setup}
\vspace*{-0.5cm}
The baseline model is trained with Adam~\cite{kingma2014adam}, an initial
learning rate of $10^{-4}$, a batch size of~64 for at most 1000~epochs with
data augmentation. The kind of data augmentation depends on the dataset:
\begin{itemize}
    \item \textbf{CIFAR-10}, \textbf{CIFAR-100} and STL-10:
          Random width and height shift by at most $\pm3$~pixels in either
          direction; Random horizontal flip.
    \item \textbf{GTSRB}, \textbf{MNIST}: Random width and height shift by at
          most $\pm5$~pixels in either direction; random rotation by at most
          $\pm15$ degrees; random channel shift; random zoom in $[0.5, 1.5]$;
          random shear by at most 6~degrees.
    \item \textbf{HASYv2}: Random width and height shift by at most
          $\pm5$~pixels in either direction; random rotation by at most $\pm5$
          degree.
    \item \textbf{SVHN}: No data augmentation.
\end{itemize}

If the dataset does not define a training/test set, a stratified
\SI{67}{\percent} / \SI{33}{\percent} split is applied. If the dataset does not
define a validation set, the training set is split in a stratified manner into
\SI{90}{\percent} training set / \SI{10}{\percent} test set.

Early stopping~\cite{Prechelt1998} with the validation accuracy as a stopping
criterion and a patience of 10~epochs is applied. After this, the model is
trained without data augmentation for at most 1000~epochs with early stopping
and the validation accuracy as a stopping criterion and a patience of
10~epochs. Kernel weights are initialized according to the uniform
initialization scheme of He~\cite{he2015delving}
(see~\cref{sec:initialization}).

The architecture of the baseline model uses a pattern of
\[\text{Conv-Block}(n) = (\text{Convolution} - \text{Batch Normalization} - \text{Activation})^n - \text{Pooling}\]
The activation function is the \gls{ELU}
(see~\cref{table:activation-functions-overview}), except for the last layer
where softmax is used. Before the last two convolutional layer, a dropout layer
with dropout probability~$0.5$ is applied. The architecture is given in detail
in~\cref{table:baseline-architecture}. Please note that the number of input-
and output channels of the network depends on the dataset. If the input image
is larger than $\SI{32}{\pixel} \times \SI{32}{\pixel}$, for each power of
two a $\text{Conv-Block}(2)$ is added at the input. For MNIST, the images are
bilinearly upsampled to $\SI{32}{\pixel} \times \SI{32}{\pixel}$.

\begin{table}[ht]
    \renewrobustcmd{\bfseries}{\fontseries{b}\selectfont}
    \sisetup{detect-weight,mode=text,group-minimum-digits = 4}
    \centering
\small
\addtolength{\tabcolsep}{-1.5pt}
\begin{tabular}{
  @{}cl
  S[table-format=3.0]@{\,}c@{\,}l@{\,}c@{\,}l
  S[table-format=5.0]
  S[table-format=8.0]
  S[table-format=4.0]@{\,}c@{\,}S[table-format=2.0]@{}>{${}}c<{{}$}@{}S[table-format=2.0]
  @{}
    }
\toprule
\# & Type
   & \multicolumn{5}{c}{\begin{tabular}[t]{l}Filters @\\ Patch size / stride\end{tabular}}
   & {Parameters} & {FLOPs} & \multicolumn{5}{c}{Output size}\\\midrule
   &    Input        &      & &                         & &       &     0                &        0                &    3 & @ & 32 & \times & 32 \\\cline{1-14}
\multicolumn{1}{|c}{1} & Convolution     & 32   &@& $3 \times 3 \times 3$   &/& 1     &   896                &  1736704                & \bfseries 32 & @ & \bfseries 32 & \times & \multicolumn{1}{c|}{\bfseries 32} \\
\multicolumn{1}{|c}{2} & BN + ELU        &      & &                         & &       &    64                &   163904                & \bfseries 32 & @ & \bfseries 32 & \times & \multicolumn{1}{c|}{\bfseries 32} \\
\multicolumn{1}{|c}{3} & Convolution     & 32   &@& $3 \times 3 \times32$   &/& 1     &  9248                & 18841600                & \bfseries 32 & @ & \bfseries 32 & \times & \multicolumn{1}{c|}{\bfseries 32} \\
\multicolumn{1}{|c}{4} & BN + ELU        &      & &                         & &       &    64                &   163904                & \bfseries 32 & @ & \bfseries 32 & \times & \multicolumn{1}{c|}{\bfseries 32} \\
\multicolumn{1}{|c}{} & Max pooling      &      & & $2 \times 2$            &/& 2     &     0                &    40960                &   32 & @ & 16 & \times & \multicolumn{1}{c|}{16} \\\cline{1-14}\\[-0.1cm]
 5 & Convolution     & 64   &@& $3 \times 3 \times32$   &/& 1     & 18496                &  9420800                &   64 & @ & 16 & \times & 16 \\
 6 & BN + ELU        &      & &                         & &       &   128                &    82048                &   64 & @ & 16 & \times & 16 \\
 7 & Convolution     & 64   &@& $3 \times 3 \times64$   &/& 1     & 36928                & \bfseries 18857984      &   64 & @ & 16 & \times & 16 \\
 8 & BN + ELU        &      & &                         & &       &   128                &    82048                &   64 & @ & 16 & \times & 16 \\
 ~ & Max pooling     &      & & $2 \times 2$            &/& 2     & ~                    &    20480                &   64 & @ &  8 & \times &  8 \\[0.1cm]
 9 & Convolution     & 64   &@& $3 \times 3 \times64$   &/& 1     & 36928                &  4714496                &   64 & @ &  8 & \times &  8 \\
10 & BN + ELU        &      & &                         & &       &   128                &    20608                &   64 & @ &  8 & \times &  8 \\
 ~ & Max pooling     &      & & $2 \times 2$            &/& 2     & ~                    &     5120                &   64 & @ &  4 & \times &  4 \\[0.1cm]
11 & Convolution (v) & 512  &@& $4 \times 4 \times64$   &/& 1     & \bfseries  524800    &  1048064                &  512 & @ &  1 & \times &  1 \\[0.1cm]
12 & BN + ELU        &      & &                         & &       &  1024                &     3584                &  512 & @ &  1 & \times &  1 \\
 ~ & Dropout 0.5     & ~    & &                         & &       &     0                &        0                &  512 & @ &  1 & \times &  1 \\
13 & Convolution     & 512  &@& $1 \times 1 \times 512$ &/& 1     &262656                &   523776                &  512 & @ &  1 & \times &  1 \\[0.1cm]
14 & BN + ELU        &      & &                         & &       &  1024                &     3584                &  512 & @ &  1 & \times &  1 \\
 ~ & Dropout 0.5     & ~    & &                         & &       &     0                &        0                &  512 & @ &  1 & \times &  1 \\
15 & Convolution     & k    &@& $1 \times 1 \times 512$ &/& 1     & {$k \cdot (512+1)$}  & {$1024 \cdot k$}        &    k & @ &  1 & \times &  1 \\[0.1cm]
   & Global avg Pooling &   & & $1 \times 1$            & &       &     0                &        {$k$}            &    k & @ &  1 & \times &  1 \\[0.1cm]
16 & BN + Softmax    & ~    &~&  ~                      & &       & {$2k$}               &     {$7k$}              &    k & @ &  1 & \times &  1 \\\midrule
   & $\sum$          & ~    & &                         & &       & {\makecell{$515k$\\+\num{892512}}} & {\makecell{$1032k$\\+\num{55729664}}} & \multicolumn{5}{r}{\num{103424}+$2k$}\\
\bottomrule
\end{tabular}
    \caption[Baseline architecture]{Baseline architecture with 3~input channels
             of size $32 \times 32$. All convolutional layers use \texttt{SAME}
             padding, except for layer~11 which used \texttt{VALID} padding in
             order to decrease the feature map size to $1\times 1$. If the
             input feature map is bigger than $32 \times 32$, for each power of
             two there are two \texttt{Convolution + BN + ELU} blocks and one
             \texttt{Max pooling} block added. This is the framed part in the
             table.}
    \label{table:baseline-architecture}
\end{table}

\begin{figure}[H]
    \hspace*{-0.8cm}
    \resizebox {1.05\columnwidth} {!} {
    \newcommand{\width}{0.2}
\newcommand{\height}{0.4}
\newcommand{\disty}{0.2}
\definecolor{colorbblue}{HTML}{0072B2}
\definecolor{colorbgreen}{HTML}{009E73}
\definecolor{colorborange}{HTML}{D55E00}

\tikzstyle{input}=[draw,fill=red!50]
\tikzstyle{conv}=[draw,fill=black!20]
\tikzstyle{max}=[draw,dashed,fill=black!10]
\tikzstyle{dropout}=[draw,dashed,fill=colorbgreen!30]
\tikzstyle{fc}=[draw,fill=green!10]
\tikzstyle{output}=[draw,fill=red!50]
\tikzstyle{act}=[draw,dashed,fill=colorbblue!30]
\def \coldist {2.3}
\def \widthb {1.9}

\begin{tikzpicture}[scale=2]
    \draw[->, -Latex, line width=5pt]
        (1.0+0*\coldist, 0.2) --(1.0+0*\coldist, -4*\disty-5*\height-0.3) --(2.1+0*\coldist, -4*\disty-5*\height-0.3) --(2.2+0*\coldist, 0.2)
     -- (1.0+1*\coldist, 0.2) --(1.0+1*\coldist, -4*\disty-5*\height-0.3) --(2.1+1*\coldist, -4*\disty-5*\height-0.3) --(2.2+1*\coldist, 0.2)
     -- (1.0+2*\coldist, 0.2) --(1.0+2*\coldist, -2*\disty-3*\height-0.3) --(2.1+2*\coldist, -2*\disty-3*\height-0.3) --(2.2+2*\coldist, 0.2)
     -- (1.0+3*\coldist, 0.2) --(1.0+3*\coldist, -3*\disty-4*\height-0.3) --(2.1+3*\coldist, -3*\disty-4*\height-0.3) --(2.2+3*\coldist, 0.2)
     -- (1.0+4*\coldist, 0.2) --(1.0+4*\coldist, -5*\disty-6*\height-0.6);

    \draw[draw=none] (0*\coldist,-1*\height)          rectangle (1.0,-1*\disty-1*\height) node[pos=.5] {$32 \times 32$};
    \draw[input]     (0*\coldist,-0*\height-0*\disty) rectangle (2.0,-0*\disty-1*\height) node[pos=.5] {Input};
    \draw[conv]      (0*\coldist,-1*\height-1*\disty) rectangle (2.0,-1*\disty-2*\height) node[pos=.5] {C $32@3 \times 3 / 1$};
    \draw[act]       (0*\coldist,-2*\height-2*\disty) rectangle (2.0,-2*\disty-3*\height) node[pos=.5] {BN + ELU};
    \draw[conv]      (0*\coldist,-3*\height-3*\disty) rectangle (2.0,-3*\disty-4*\height) node[pos=.5] {C $32@3 \times 3 / 1$};
    \draw[act]       (0*\coldist,-4*\height-4*\disty) rectangle (2.0,-4*\disty-5*\height) node[pos=.5] {BN + ELU};

    \draw[draw=none] (1*\coldist,-1*\height)          rectangle (1*\coldist+\widthb/2,-1*\disty-1*\height) node[pos=.5] {$16 \times 16$};
    \draw[max]       (1*\coldist,-0*\height-0*\disty) rectangle (1*\coldist+\widthb,-0*\disty-1*\height) node[pos=.5] {max pooling $2\times 2 / 2$};
    \draw[conv]      (1*\coldist,-1*\height-1*\disty) rectangle (1*\coldist+\widthb,-1*\disty-2*\height) node[pos=.5] {C $64@3 \times 3 / 1$};
    \draw[act]       (1*\coldist,-2*\height-2*\disty) rectangle (1*\coldist+\widthb,-2*\disty-3*\height) node[pos=.5] {BN + ELU};
    \draw[conv]      (1*\coldist,-3*\height-3*\disty) rectangle (1*\coldist+\widthb,-3*\disty-4*\height) node[pos=.5] {C $64@3 \times 3 / 1$};
    \draw[act]       (1*\coldist,-4*\height-4*\disty) rectangle (1*\coldist+\widthb,-4*\disty-5*\height) node[pos=.5] {BN + ELU};

    \draw[draw=none] (2*\coldist,-1*\height) rectangle (2*\coldist+\widthb/2,-1*\disty-1*\height) node[pos=.5] {$8 \times 8$};
    \draw[max]       (2*\coldist,-0*\height-0*\disty) rectangle (2*\coldist+\widthb,-0*\disty-1*\height) node[pos=.5] {max pooling $2\times 2 / 2$};
    \draw[conv]      (2*\coldist,-1*\height-1*\disty) rectangle (2*\coldist+\widthb,-1*\disty-2*\height) node[pos=.5] {C $64@3 \times 3 / 1$};
    \draw[act]       (2*\coldist,-2*\height-2*\disty) rectangle (2*\coldist+\widthb,-2*\disty-3*\height) node[pos=.5] {BN + ELU};

    \draw[draw=none] (3*\coldist,-1*\height) rectangle (3*\coldist+\widthb/2,-1*\disty-1*\height) node[pos=.5] {$4 \times 4$};
    \draw[max]       (3*\coldist,-0*\height-0*\disty) rectangle (3*\coldist+\widthb,-0*\disty-1*\height) node[pos=.5] {max pooling $2\times 2 / 2$};
    \draw[conv]      (3*\coldist,-1*\height-1*\disty) rectangle (3*\coldist+\widthb,-1*\disty-2*\height) node[pos=.5] {C $512@4 \times 4 / 1$ (V)};
    \draw[act]       (3*\coldist,-2*\height-2*\disty) rectangle (3*\coldist+\widthb,-2*\disty-3*\height) node[pos=.5] {BN + ELU};
    \draw[dropout]   (3*\coldist,-3*\height-3*\disty) rectangle (3*\coldist+\widthb,-3*\disty-4*\height) node[pos=.5] {Dropout, $p=0.5$};

    \draw[draw=none] (4*\coldist,-1*\height) rectangle (4*\coldist+\widthb/2,-1*\disty-1*\height) node[pos=.5] {$1 \times 1$};
    \draw[conv]      (4*\coldist,-0*\height-0*\disty) rectangle (4*\coldist+\widthb,-0*\disty-1*\height) node[pos=.5] {C $512@1 \times 1 / 1$};
    \draw[act]       (4*\coldist,-1*\height-1*\disty) rectangle (4*\coldist+\widthb,-1*\disty-2*\height) node[pos=.5] {BN + ELU};
    \draw[dropout]   (4*\coldist,-2*\height-2*\disty) rectangle (4*\coldist+\widthb,-2*\disty-3*\height) node[pos=.5] {Dropout, $p=0.5$};
    \draw[conv]      (4*\coldist,-3*\height-3*\disty) rectangle (4*\coldist+\widthb,-3*\disty-4*\height) node[pos=.5] {C $k@1 \times 1 / 1$};
    \draw[dropout]   (4*\coldist,-4*\height-4*\disty) rectangle (4*\coldist+\widthb,-4*\disty-5*\height) node[pos=.5] {Global AVG pooling};
    \draw[max]       (4*\coldist,-5*\height-5*\disty) rectangle (4*\coldist+\widthb,-5*\disty-6*\height) node[pos=.5] {BN + Softmax};
\end{tikzpicture}
    }
    \caption[Baseline architecture]{Architecture of the baseline model.
    \texttt{C $32@3 \times 3 / 1$} is a convolutional layer with 32~filters of
    kernel size $3 \times 3$ with stride~1.}
    \label{fig:baseline-architecture}
\end{figure}
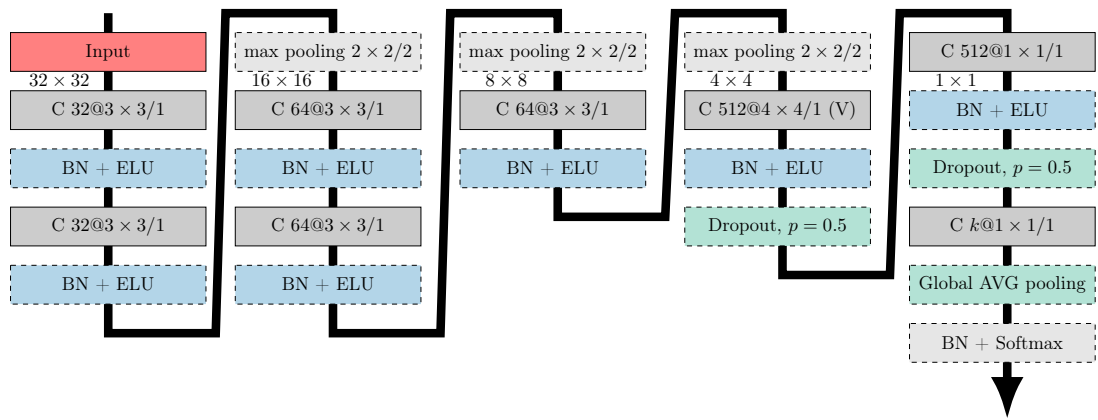

\clearpage
\subsection{Baseline Evaluation}
The results for the baseline model evaluated on eight datasets are given
in~\cref{table:baseline-model-performance}. The speed for inference for
different \glspl{GPU} is given in~\cref{table:baseline-model-time}.

\begin{table}[ht]
    \centering
    \begin{tabular}{lrlrlrr}
    \toprule
    \multirow{2}{*}{Dataset} & \multicolumn{4}{c}{Single Model Accuracy} &  \multicolumn{2}{c}{Ensemble of 10} \\
              & \multicolumn{2}{c}{Training Set}     & \multicolumn{2}{c}{Test Set}         & Training Set          & Test Set \\\midrule
    Asirra    & \SI{94.22}{\percent} & $\sigma=3.49$ & \SI{94.37}{\percent} & $\sigma=3.47$ &  \SI{97.07}{\percent} & \SI{97.37}{\percent} \\
    CIFAR-10  & \SI{91.23}{\percent} & $\sigma=1.10$ & \SI{85.84}{\percent} & $\sigma=0.87$ &  \SI{92.36}{\percent} & \SI{86.75}{\percent} \\
    CIFAR-100 & \SI{76.64}{\percent} & $\sigma=1.48$ & \SI{63.38}{\percent} & $\sigma=0.55$ &  \SI{78.30}{\percent} & \SI{64.70}{\percent} \\
    GTSRB     &\SI{100.00}{\percent} & $\sigma=0.00$ & \SI{99.18}{\percent} & $\sigma=0.11$ & \SI{100.00}{\percent} & \SI{99.46}{\percent} \\
    HASYv2    & \SI{89.49}{\percent} & $\sigma=0.42$ & \SI{85.35}{\percent} & $\sigma=0.10$ &  \SI{89.94}{\percent} & \SI{86.03}{\percent} \\
    MNIST     & \SI{99.93}{\percent} & $\sigma=0.07$ & \SI{99.53}{\percent} & $\sigma=0.06$ &  \SI{99.99}{\percent} & \SI{99.58}{\percent} \\
    STL-10    & \SI{94.12}{\percent} & $\sigma=0.87$ & \SI{75.67}{\percent} & $\sigma=0.34$ &  \SI{96.35}{\percent} & \SI{77.62}{\percent} \\
    SVHN      & \SI{99.02}{\percent} & $\sigma=0.07$ & \SI{96.28}{\percent} & $\sigma=0.10$ &  \SI{99.42}{\percent} & \SI{97.20}{\percent} \\
    \bottomrule
    \end{tabular}
    \caption[Baseline model evaluation]{\small Baseline model accuracy on eight
             datasets. The single model actuary is the 10~models used in the
             ensemble. The empirical standard deviation~$\sigma$ of the
             accuracy is also given. CIFAR-10, CIFAR-100 and STL-10 models use
             test-time transformations. None of the models uses unlabeled data
             or data from other datasets. For HASYv2 no test time
             transformations are used.}
    \label{table:baseline-model-performance}
\end{table}

\begin{table}[ht]
    \centering
    {\small
    \begin{tabular}{lllrrr}
    \toprule
    \multirow{2}{*}{Network}
             & \multirow{2}{*}{GPU}
                   & \multirow{2}{*}{Tensorflow} & \multicolumn{2}{c}{Inference per}                               & \multicolumn{1}{c}{Training} \\\cline{4-5}
                   &         &               & 1 Image               & 128 images              & \multicolumn{1}{c}{time / epoch} \\\midrule
    Baseline       & Default & Intel i7-4930K& \SI{3}{\milli\second} & \SI{244}{\milli\second} & \SI{231.0}{\second}           \\
    Baseline       & Optimized&Intel i7-4930K& \SI{2}{\milli\second} & \SI{143}{\milli\second} & \SI{149.0}{\second}           \\
    Baseline       & Default & GeForce~940MX & \SI{4}{\milli\second} & \SI{120}{\milli\second} & \SI{145.6}{\second}           \\
    Baseline       & Default & GTX 970       & \SI{6}{\milli\second} &  \SI{32}{\milli\second} & \SIrange{25.0}{26.3}{\second} \\
    Baseline       & Default & GTX 980       & \SI{3}{\milli\second} &  \SI{24}{\milli\second} & \SIrange{20.5}{21.1}{\second} \\
    Baseline       & Default & GTX 980 Ti    & \SI{5}{\milli\second} &  \SI{27}{\milli\second} & \SIrange{22.0}{22.1}{\second} \\
    Baseline       & Default & GTX 1070      & \textbf{\SI{2}{\milli\second}} &  \textbf{\SI{15}{\milli\second}} & \textbf{\SIrange{14.4}{14.5}{\second}} \\
    Baseline       & Default & Titan Black   & \SI{4}{\milli\second} &  \SI{25}{\milli\second} & \SIrange{28.1}{28.1}{\second} \\
    Baseline       & Optimized & Titan Black & \SI{3}{\milli\second} &  \SI{22}{\milli\second} & \SIrange{24.4}{24.4}{\second} \\
    DenseNet-40-12 & Default & GeForce~940MX &\SI{27}{\milli\second} &\SI{2403}{\milli\second} &  ---                          \\
    \bottomrule
    \end{tabular}
    }
    \caption[Baseline model speed comparison]{\small Speed comparison of the
             baseline model on CIFAR-10. The baseline model is evaluated on six
             Nvidia \glspl{GPU} and one CPU\@. The weights for DenseNet-40-12 are
             taken from~\cite{Majumdar2017-densenet-weights}. Weights the
             baseline model can be found at~\cite{thoma-msthesis-blog}. The
             optimized Tensorflow build makes use of SSE4.X, AVX, AVX2 and FMA
             instructions.}
    \label{table:baseline-model-time}
\end{table}
\clearpage

\subsection{Weight distribution}
\begin{figure}[ht]
    \centering
    \includegraphics[width=0.98\linewidth,keepaspectratio]{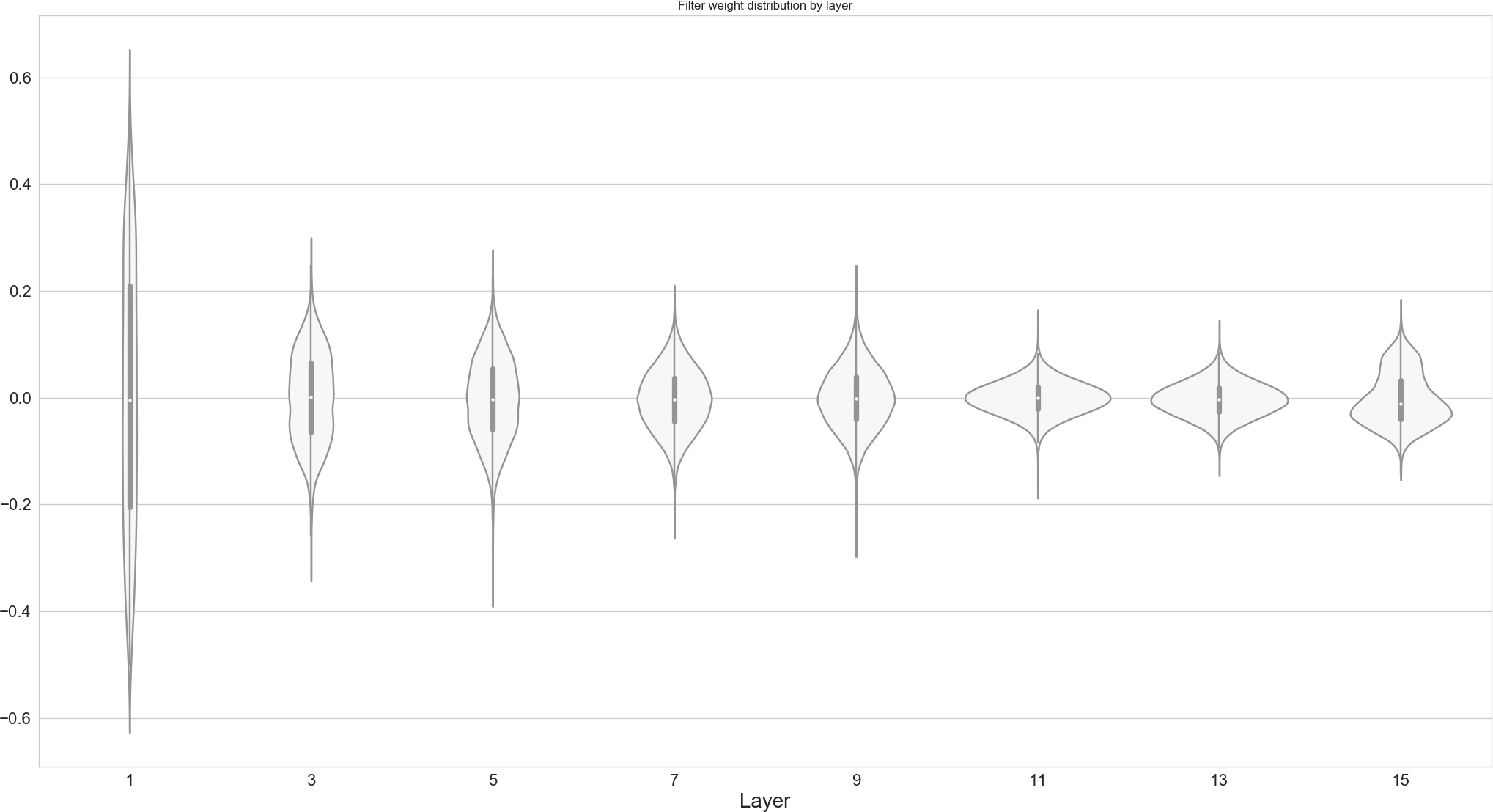}
    \caption[Baseline model filter weight distribution]{Violin plots of the
             distribution of filter weights of a baseline model trained on
             CIFAR-100. The weights of the first layer are relatively evenly
             spread in the interval $[-0.4, +0.4]$. With every layer the
             interval which contains \SI{95}{\percent} of the weights and is
             centered around the mean becomes smaller, especially with layer~11
             where the feature maps are of size $1
             \times 1$. In contrast to the other layers, the last convolutional
             layer has a bimodal distribution.\\
             This plot indicates that the network might benefit from bigger
             filters in the first layer, whereas the filters in layers 7 -- 11
             could potentially be smaller.}
    \label{fig:baseline-filter-weight-dist}
\end{figure}

\begin{figure}[ht]
    \centering
    \includegraphics[width=0.98\linewidth,keepaspectratio]{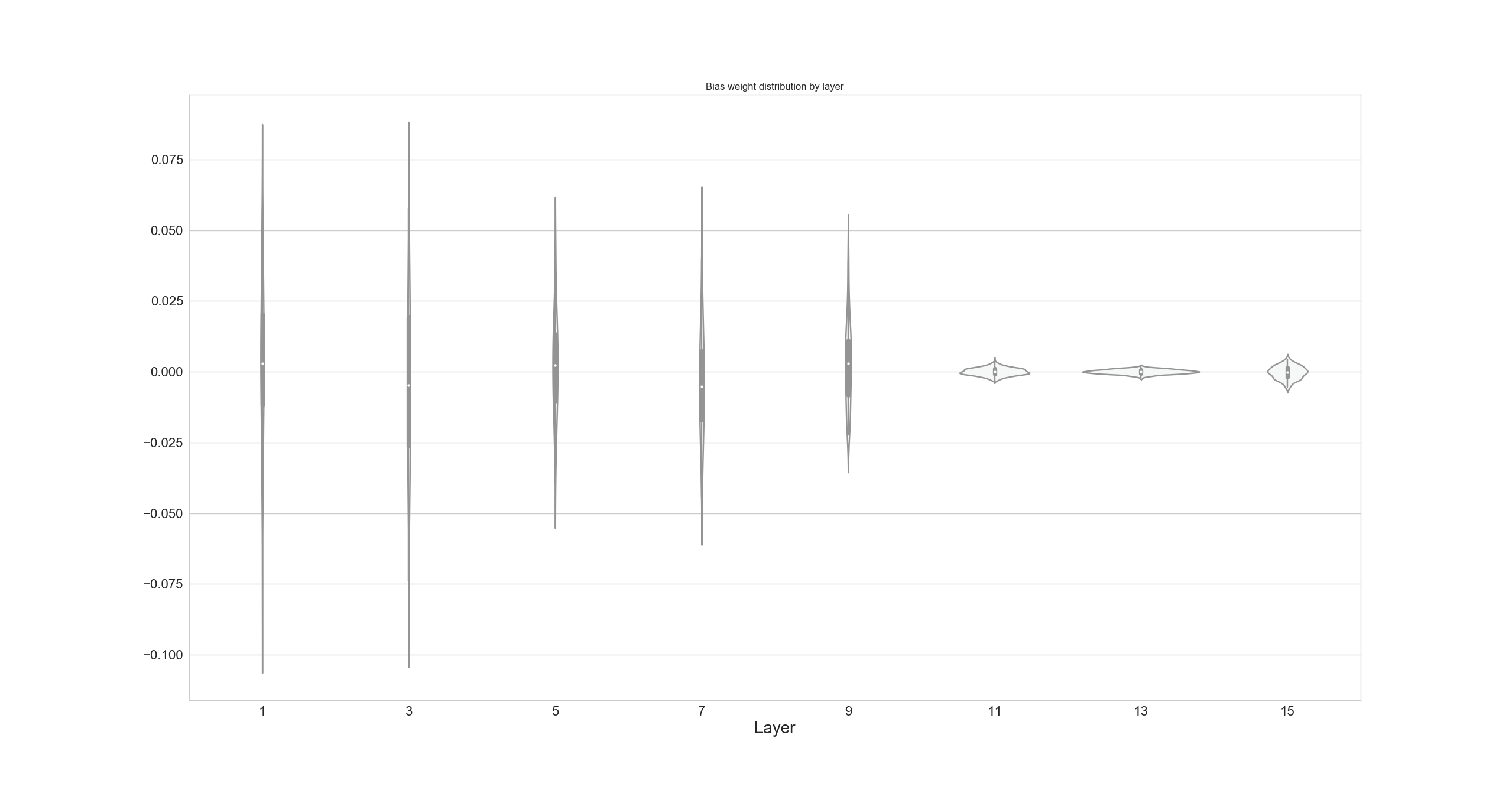}
    \caption[Baseline model bias weight distribution]{Violin plots of the
             distribution of bias weights of a baseline model trained on
             CIFAR-100. While the first layers biases are in $[-0.1, +0.1]$,
             after each max-pooling layer the interval which contains
             \SI{95}{\percent} of the weights and is centered around the mean
             becomes smaller. In the last three convolutional layer, most bias
             weights are in $[-0.005, +0.005]$.}
    \label{fig:baseline-bias-weight-dist}
\end{figure}

The distribution of filter weights by layer is visualized
in~\cref{fig:baseline-filter-weight-dist} and the distribution of bias weights
by layer is shown in~\cref{fig:baseline-bias-weight-dist}. Although both
figures only show the distribution for one specific model trained on CIFAR-100,
the following observed patterns are consistent for 70~models (7~datasets and 10~models
per dataset):
\begin{itemize}
    \item The empiric $[0.5-\text{percentile}, 99.5-\text{percentile}]$
          interval which contains $\SI{99}{\percent}$ of the filter weights is
          almost symmetric around zero. The same is true for the bias weights.
    \item The farther a layer is from the input away, the smaller the
          99-percentile interval is, except for the last layer
          (see~\cref{table:baseline-percentile-interval}).
    \item The 99-percentile interval of the first layers filter weights is
          about $[-0.5, +0.5]$, except for MNIST and HASYv2 where it is in
          $[-0.8, 0.8]$.
    \item The 99-percentile interval of the first layers bias weights is
          always in $[-0.2, 0.2]$.
    \item The distribution of filter weights of the last convolutional layer
          is not symmetric. In some cases the distribution is also not
          unimodal.
    \item The bias weights of the last three layers are very close to zero.
          The absolute value of most of them is smaller than $10^{-2}$.
\end{itemize}

\begin{figure}[ht]
    \centering
    \includegraphics[width=0.98\linewidth,keepaspectratio]{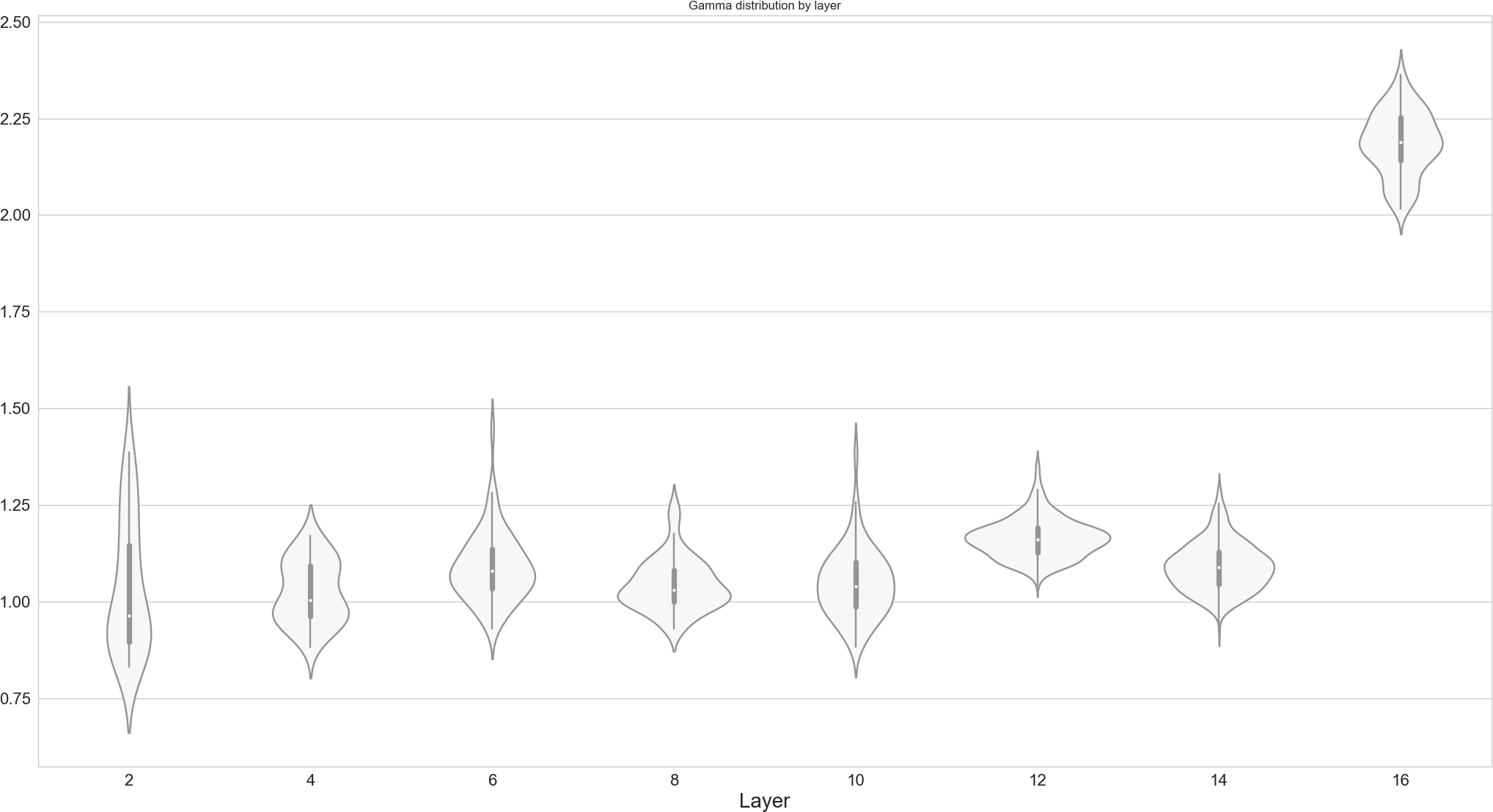}
    \caption[Baseline model $\gamma$ distribution]{Violin plots of the
             distribution of the $\gamma$ parameter of Batch Normalization
             layers of a baseline model trained on CIFAR-100.}
    \label{fig:baseline-CIFAR100-gamma-weight-dist}
\end{figure}

\begin{figure}[ht]
    \centering
    \includegraphics[width=0.98\linewidth,keepaspectratio]{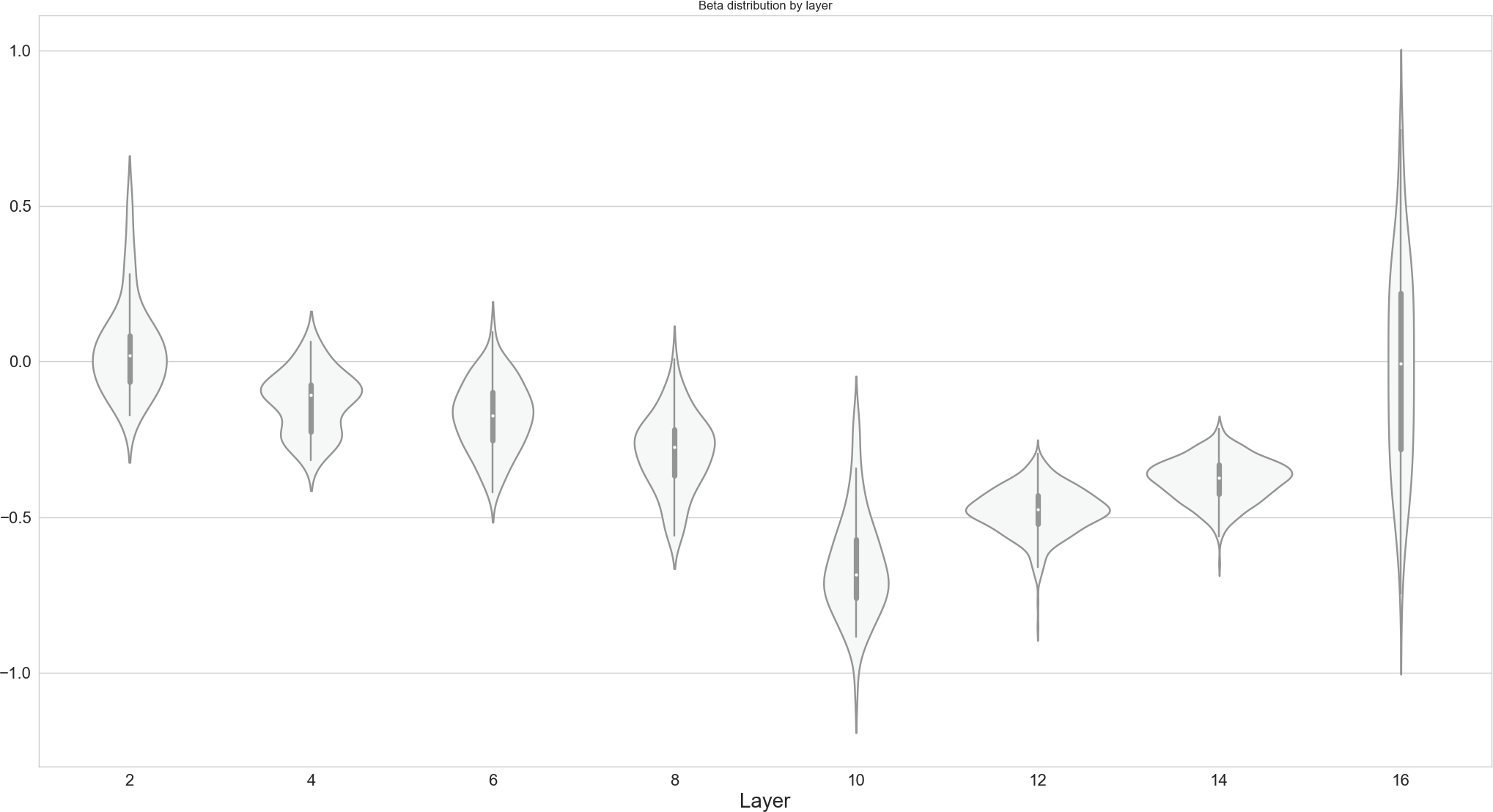}
    \caption[Baseline model $\beta$ distribution]{The distribution of the
             $\beta$ parameter of Batch Normalization layers of a baseline
             model trained on CIFAR-100.}
    \label{fig:baseline-CIFAR100-beta-weight-dist}
\end{figure}

Similarly, \cref{fig:baseline-CIFAR100-gamma-weight-dist} and
\cref{fig:baseline-CIFAR100-beta-weight-dist} show the distribution of the
$\gamma$ and the $\beta$ parameter of Batch Normalization. It is expected that
$\gamma$ is close to 1 and $\beta$ is close to 0. In those cases, the Batch
Normalization layer equals the identity and thus is only relevant for the
training. While $\gamma$ and $\beta$ do not show as clear patterns as the
filter and bias weights of convolutional layers, some observations are also
consistent through all models even for different datasets:
\begin{itemize}
    \item $\gamma$ of the last layer (layer~16) is bigger than 1.3.
    \item The 99-percentile interval for $\beta$ of the last layer is longer
          than the other 99-percentile intervals.
    \item The 99-percentile interval for $\beta$ of the fourth-last (layer~14
          for STL-10, layer~10 for all other models) is more negative then all
          other layers.
\end{itemize}

\begin{figure}[ht]
    \centering
    \includegraphics[width=0.98\linewidth,keepaspectratio]{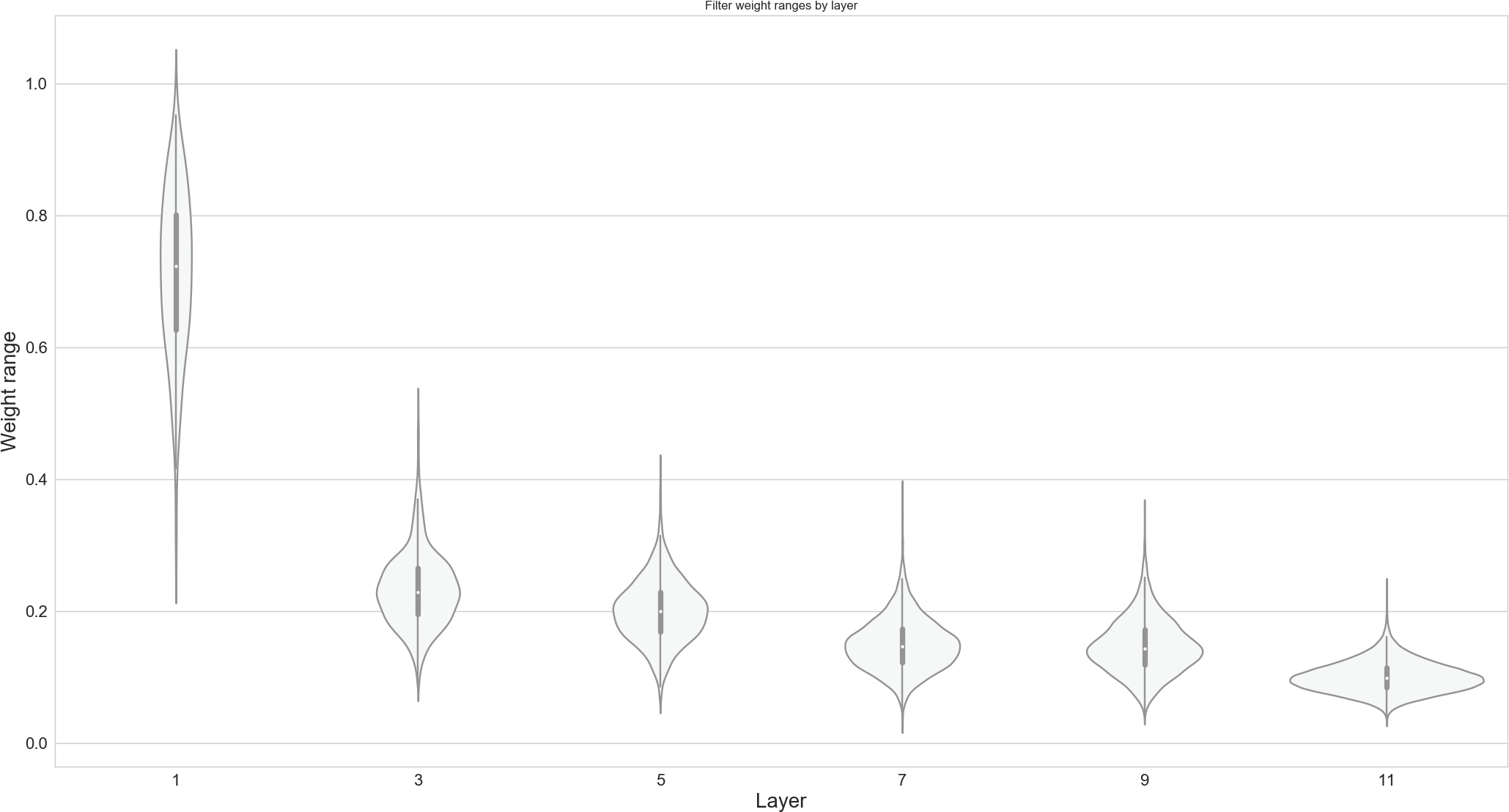}
    \caption[Baseline model filter weight range distribution]{The distribution
             of the range of values (max - min) of filters by channel and
             layer. For each filter, the range of values is recorded by
             channel. The smaller this range is, the less information is lost
             if a $n \times n$ filter is replaced by a $(n-1) \times (n-1)$
             filter.}
    \label{fig:baseline-filter-weight-range}
\end{figure}

Finally, the distribution of filter weight ranges is plotted
in~\cref{fig:baseline-filter-weight-range} for each convolutional layer. The
ranges are calculated for each channel and filter separately. The smaller the
values are, the less information is lost if the filters are replaced by smaller
filters.

\clearpage

\subsection{Training behavior}
Due to early stopping, the number of epochs which a model was trained differ.
The number of epochs trained with augmentation ranged from 133~epochs to
182~epochs with a standard deviation of 17.3~epochs for CIFAR-100.

\begin{figure}[H]
    \hspace*{-0.8cm}
    \resizebox {1.05\columnwidth} {!} {
    \definecolor{c1}{HTML}{0072B2}
\definecolor{c2}{HTML}{009E73}
\definecolor{c3}{HTML}{CB0000}
\begin{tikzpicture}
\pgfplotsset{
    scale only axis,
    xmin=0, xmax=145,
    width=15cm, height=8cm,     
}
    \begin{axis}[
            axis y line*=left,
            ymin=0.15,
            grid = major,
            grid style={dashed, gray!30},
            ylabel=validation accuracy,
            legend style={at={(0.5,0.1)},anchor=south},
            xlabel=epoch,
            xticklabel style=
            {/pgf/number format/1000 sep=,rotate=60,anchor=east,font=\scriptsize},
         ]
          \addplot[thick, c1, mark=., densely dashed] table [x=epoch, y=max_acc, col sep=comma] {graphics/baseline_cifar_test_acc.csv};\label{plot_one}\addlegendentry{maximum validation accuracy}
          \addplot[thick, c2, mark=.] table [x=epoch,y=min_acc, col sep=comma] {graphics/baseline_cifar_test_acc.csv};\label{plot_two}\addlegendentry{minimum validation accuracy}
    \end{axis}
    \begin{axis}[
      axis y line*=right,
      axis x line=none,
      ylabel=loss,
      legend style={at={(0.5,0.1)},anchor=south},
      y dir=reverse
    ]
          \addlegendimage{thick, c1, mark=., densely dashed}\addlegendentry{maximum validation accuracy}
          \addlegendimage{thick, c2, mark=.}\addlegendentry{minimum validation accuracy}
          \addplot[thick, c3, mark=., dotted] table [x=epoch,y=mean_loss, col sep=comma] {graphics/baseline_cifar_test_acc.csv};
          \addlegendentry{mean loss}
    \end{axis}
\end{tikzpicture}
    }
    \caption[Baseline model CIFAR-100 validation accuracy]{Minimum and maximum
             validation accuracy of the 10 trained models by epoch. The
             differences do not exceed \SI{1}{\percent} and does not increase
             by training epoch. Four models stopped the first training stage at
             epoch~133 which causes the shift in the loss and the maximum
             validation accuracy.}
    \label{fig:baseline-CIFAR-100-validation-accuracy}
\end{figure}
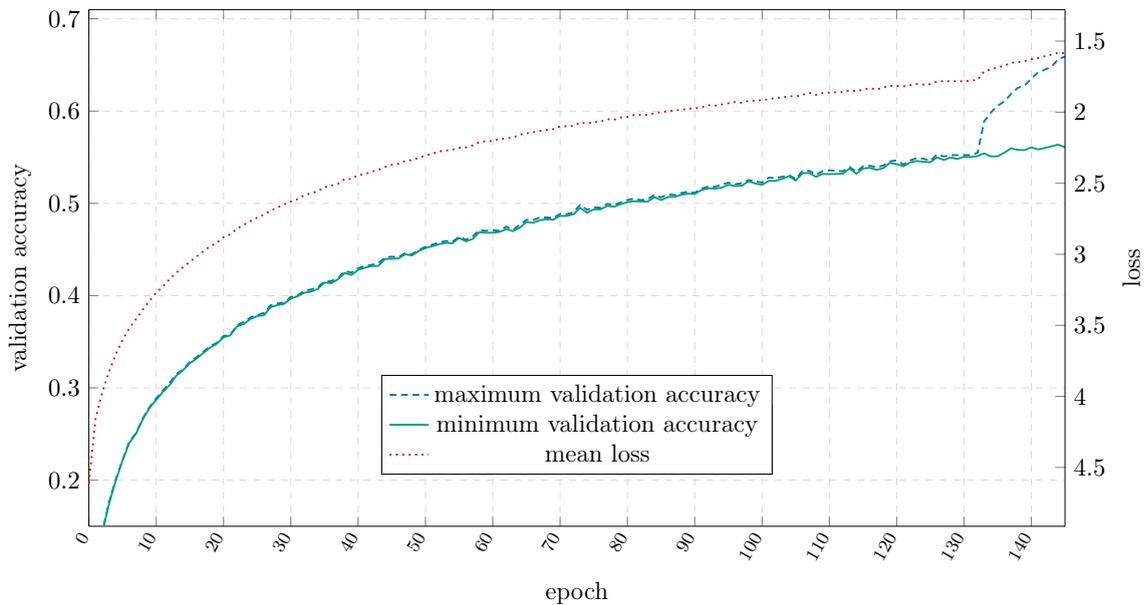

\Cref{fig:baseline-CIFAR-100-validation-accuracy} shows the worst and the best
validation accuracy during the training with augmented data. Different
initializations lead to very similar validation accuracies during training. The
image might lead to the wrong conclusion that models which are better at the
start are also better at the end. In order to check this hypothesis, the
relative order of validation accuracies for the 10~CIFAR-100 models was
examined. If the relative ordering stays approximately the same, then it can be
considered to run the first few epochs many times and only train the best
models to the end. For 10~models, there can be $\frac{10^2 - 10}{2} = 45$
pair-wise changes in the ordering at maximum if the relative order of
validation accuracies is reversed. For the baseline model, 21.8~changes in the
relative order of accuracies occurred in average for each pair of epochs $(i,
i+1)$. This means if one knows only the relative order of the validation accuracy
of two models $m$ and $m'$ in epoch $i$, it is doubtful if one can make any
statement about the ordering of $m$ and $m'$ in epoch $i+1$.

\begin{figure}[ht]
    \centering
    \includegraphics[width=0.98\linewidth,keepaspectratio]{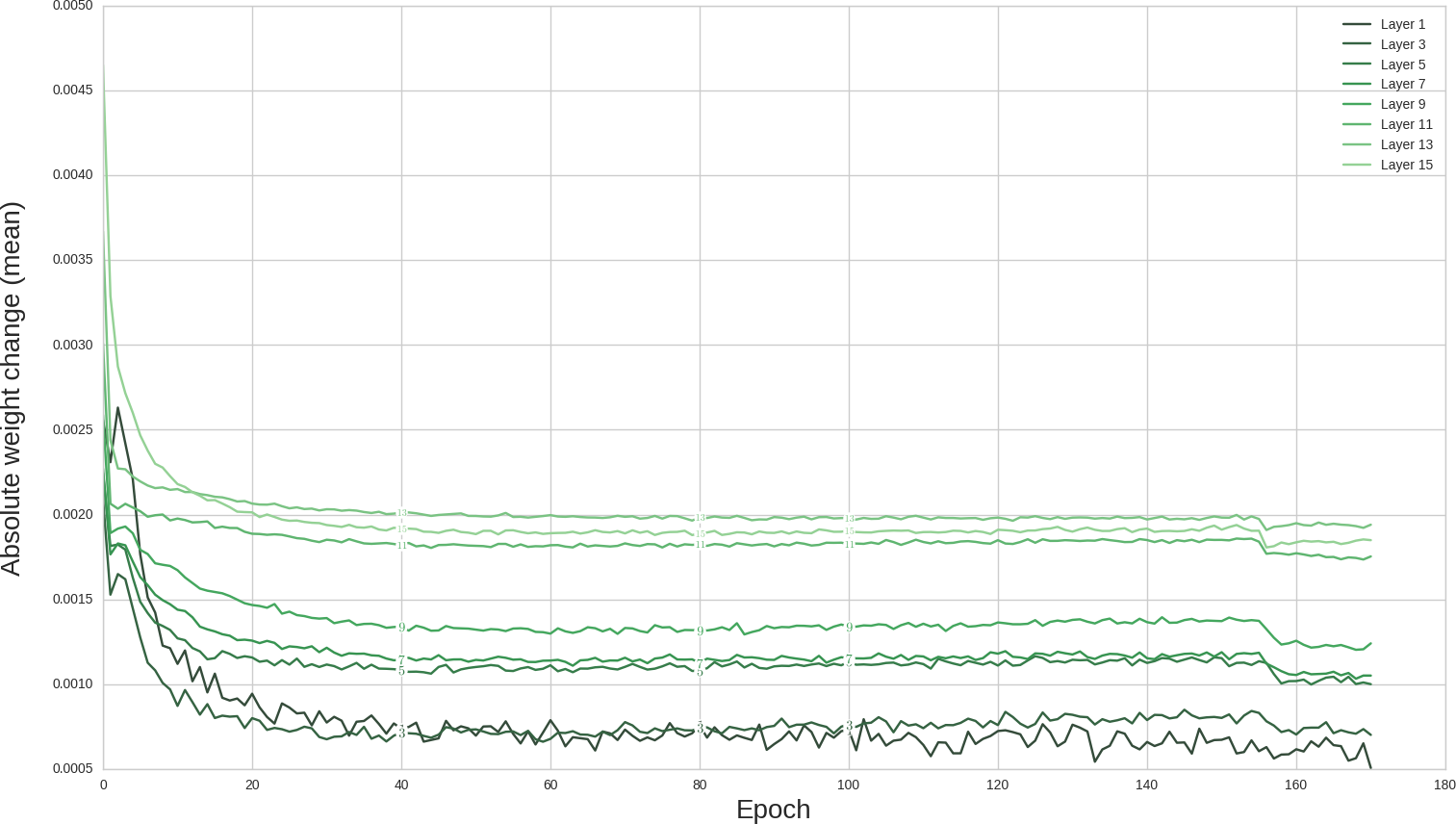}
    \caption[Baseline Weight updates (mean)]{Mean weight updates of the baseline model between epochs by layer.}
    \label{fig:baseline-weight-updates-mean}
\end{figure}

\begin{figure}[ht]
    \centering
    \includegraphics[width=0.98\linewidth,keepaspectratio]{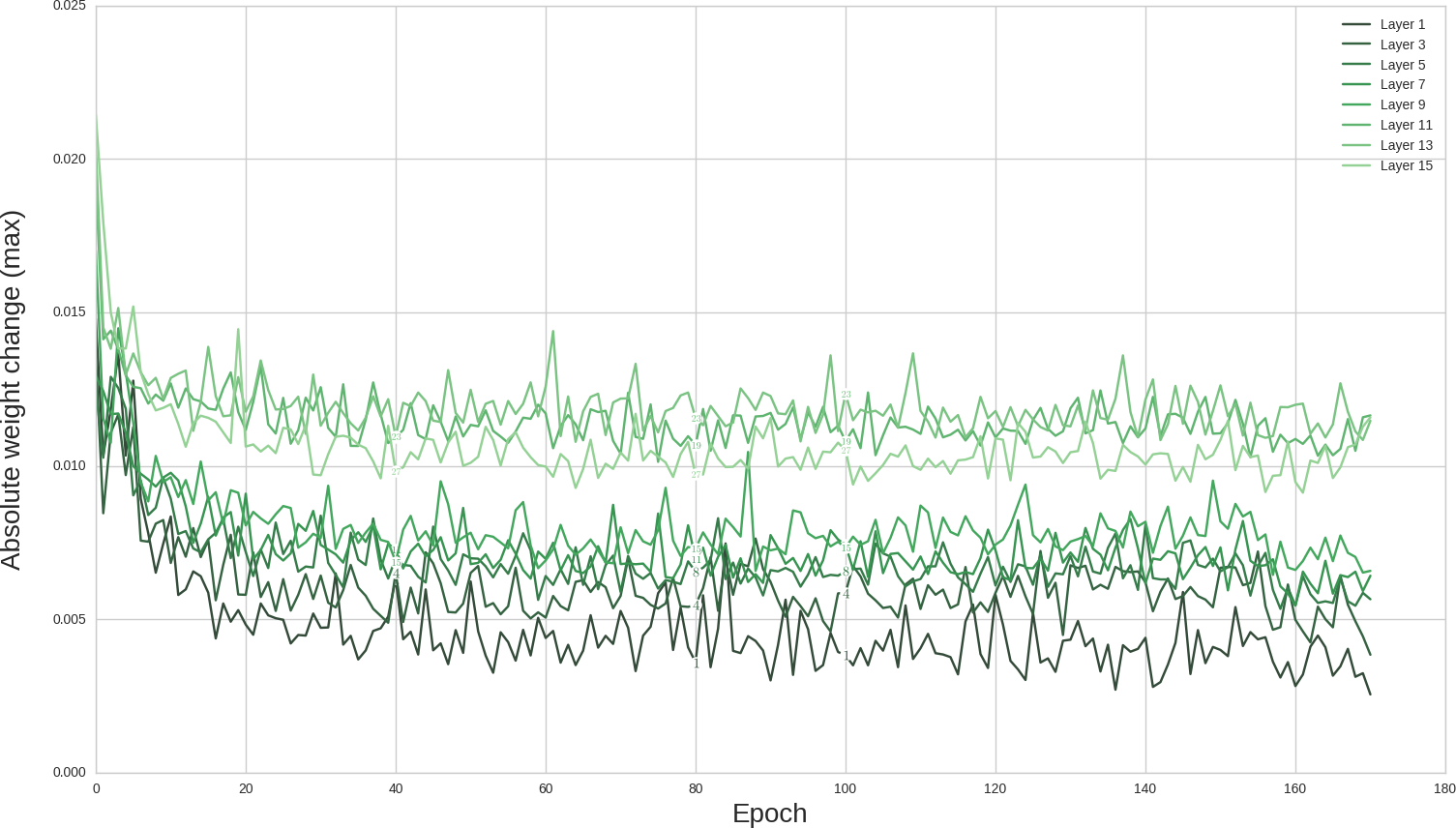}
    \caption[Baseline Weight updates (maximum)]{Maximum weight updates of the baseline model between epochs by layer.}
    \label{fig:baseline-weight-updates-max}
\end{figure}

\begin{figure}[ht]
    \centering
    \includegraphics[width=0.98\linewidth,keepaspectratio]{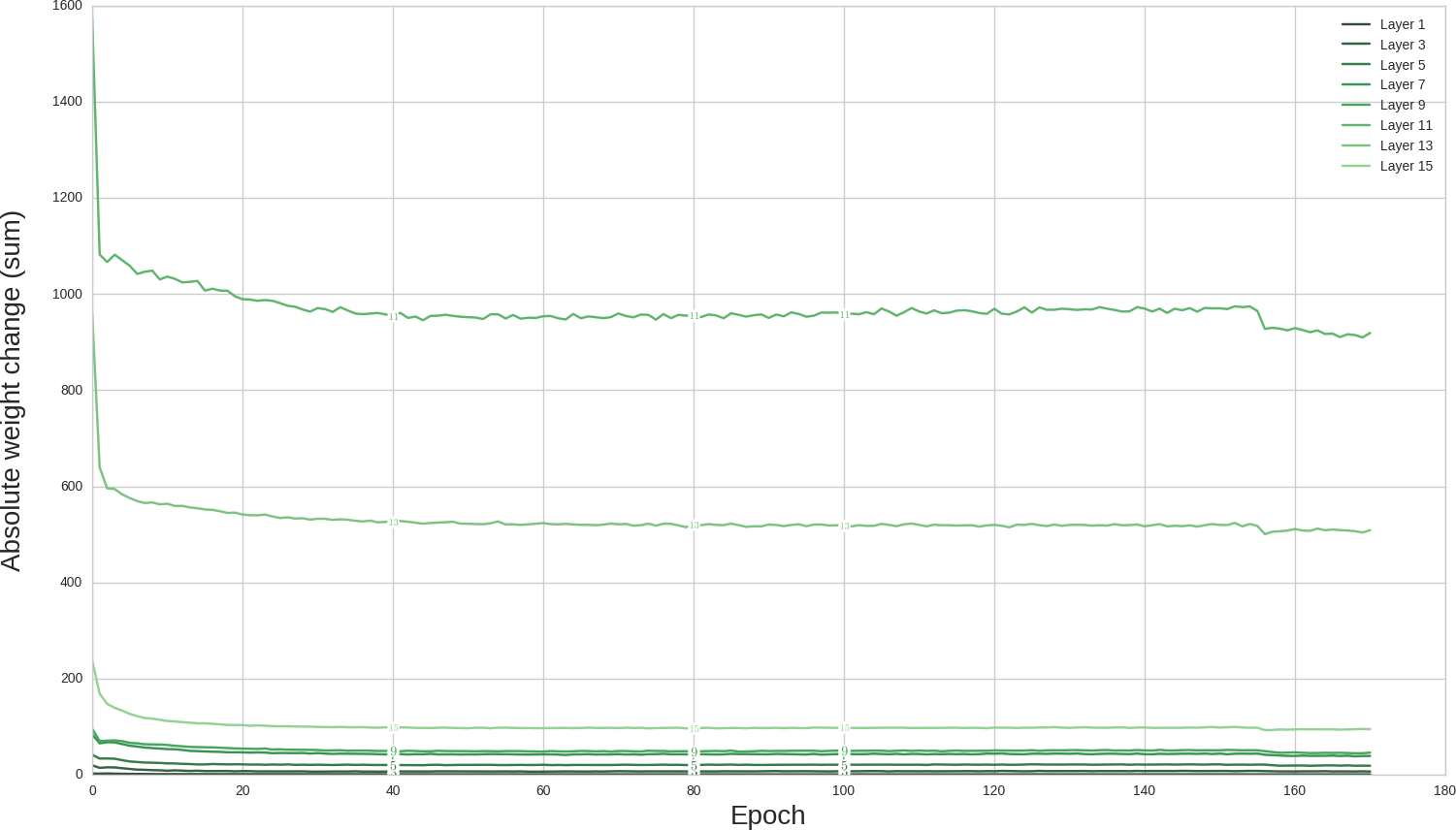}
    \caption[Baseline Weight updates (sum)]{Sum of weight updates of the baseline model between epochs by layer.}
    \label{fig:baseline-weight-updates-sum}
\end{figure}

\Cref{fig:baseline-weight-updates-mean,fig:baseline-weight-updates-max,fig:baseline-weight-updates-sum}
show how the weights changed while training on
CIFAR-100. It was expected that the absolute value of weight updates during
epochs (sum, max, and mean) decrease in later training stages. The intuition
was that weights need to be adjusted in a coarse way first. After that, the
intuition was that only slight modifications are applied by the \gls{SGD} based
training algorithm (ADAM). The mean, max and sum of weight updates as displayed in
\cref{fig:baseline-weight-updates-mean,fig:baseline-weight-updates-max,fig:baseline-weight-updates-sum},
however, do not show such a clear pattern. The biggest change happens as
expected in the first epoch after the weights are initialized. The change from
augmented training to non-augmented training was at epoch~156 to epoch~157

It can be observed, that layers which receive more input feature maps get
larger weight updates in mean. As layers which are closer to the output take
more input feature maps, their weight updates are larger. This pattern does not
occur when \gls{SGD} is used as the optimizer.


\clearpage
\section{Confusion Matrix Ordering}
The visualization of the confusion matrix can give valuable information about
which part of the task is hard. For more than about 10~classes, however, it
becomes hard to visualize and read.

\begin{figure}[ht]
    \centering
    \subfloat[CIFAR-10 Test set]{
        \includegraphics[width=0.45\textwidth]{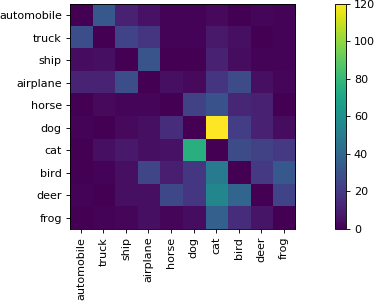}
        \label{fig:CIFAR-10-cm-test-sorted}
    }%
    \subfloat[Random]{
        \includegraphics[width=0.45\textwidth]{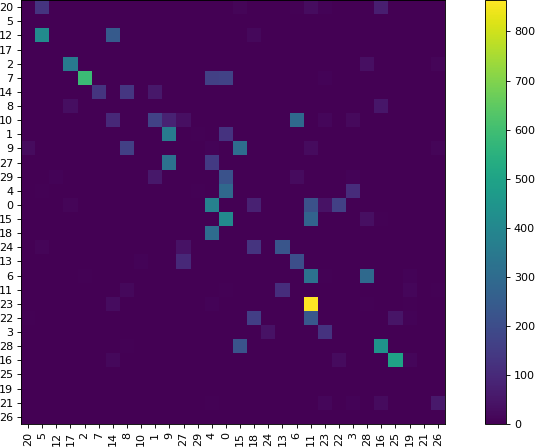}
        \label{fig:random-control}
    }
    \caption[Confusion matrices for
             CIFAR-10]{\cref{fig:CIFAR-10-cm-test-sorted} shows an ordered
             confusion matrix of the CIFAR-10 dataset. The diagonal elements
             are set to~0 in order to make other elements easier to see.\\
             \Cref{fig:random-control} shows a confusion matrix with random
             mistakes.}
    \label{fig:both}
\end{figure}

For CIFAR-10, the proposed method groups the four object classes and the six
animal classes together (see~\cref{fig:CIFAR-10-cm-test-sorted}).

\begin{figure}[ht]
    \centering
    \vspace*{-0.5cm}
    \includegraphics[width=0.98\paperwidth,height=0.38\paperheight,keepaspectratio]{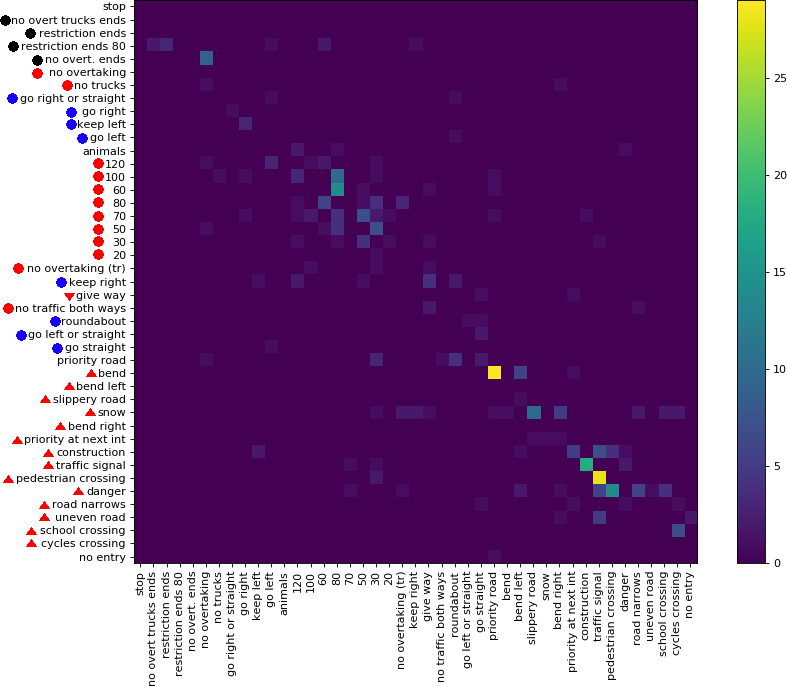}
    \includegraphics[width=0.98\paperwidth,height=0.38\paperheight,keepaspectratio]{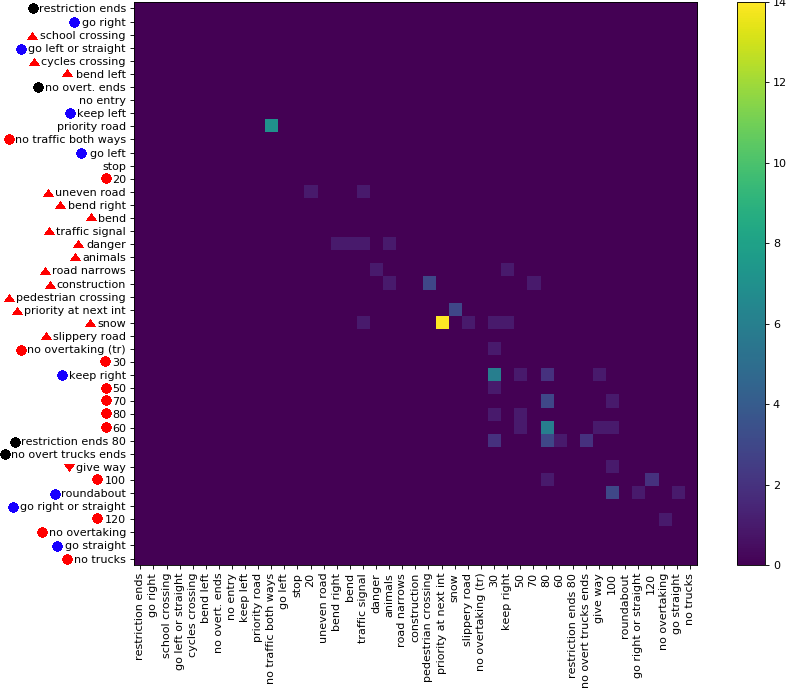}
    \caption[Confusion matrices for GTSRB]{The first image shows the confusion
             matrix for the test of GTSRB set after optimization
             to~\cref{eq:dist}. The diagonal elements are set to~0 in order to
             make other elements easier to see. The symbols next to the label
             on the vertical axis indicate the shape and the color of the
             signs.\\
             The second image shows the same, but with baseline model.\\
             Best viewed in electronic form.}
    \label{fig:GTSRB-test-cm}
\end{figure}

The first image of \cref{fig:GTSRB-test-cm} shows one example of a classifier with only
\SI{97.13}{\percent} test accuracy where a good permutation was found. Please
note that this is not the best classifier. The confusion matrix which resulted
from a baseline classifier with \SI{99.32}{\percent} test accuracy is displayed
in as the second image.

Those results suggest that the ordering of classes is a valuable tool to make
patterns easier to see. Humans, however, are good at finding patterns even if
they come from random noise. Hence, for comparison, a confusion matrix of a
classifier with 30~classes, $\SI{60}{\percent}$ accuracy and
$\SI{40}{\percent}$ uniformly random errors of a balanced dataset is created,
optimized according to \cref{eq:dist} and shown in~\cref{fig:random-control}.
It clearly looks different than \cref{fig:CIFAR-10-cm-test-sorted}.

\begin{figure}[ht]
    \centering
    \vspace*{-0.5cm}
    \includegraphics[width=0.98\paperwidth,height=0.4\paperheight,keepaspectratio]{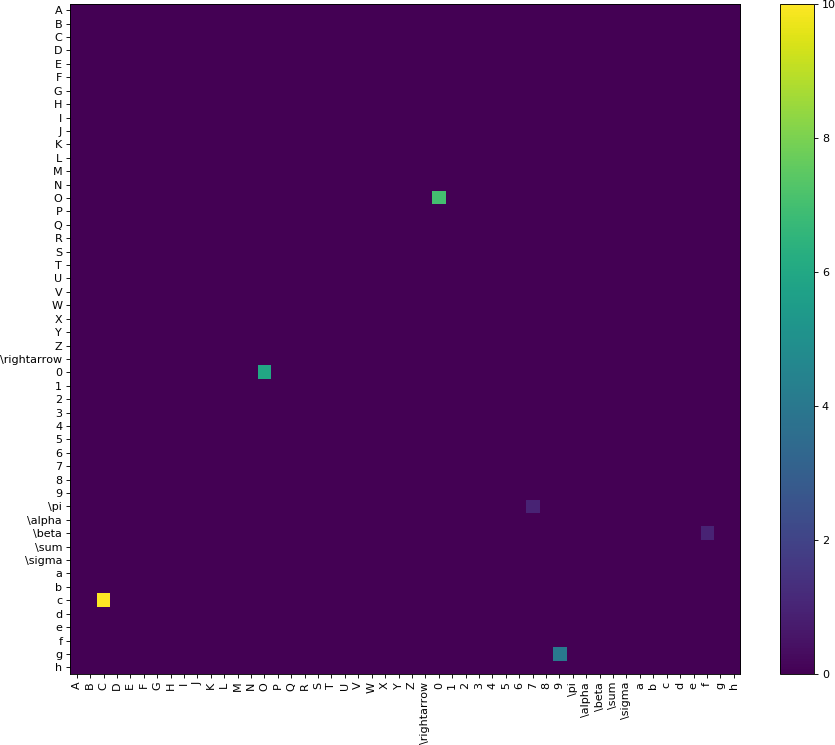}\\
    \includegraphics[width=0.98\paperwidth,height=0.4\paperheight,keepaspectratio]{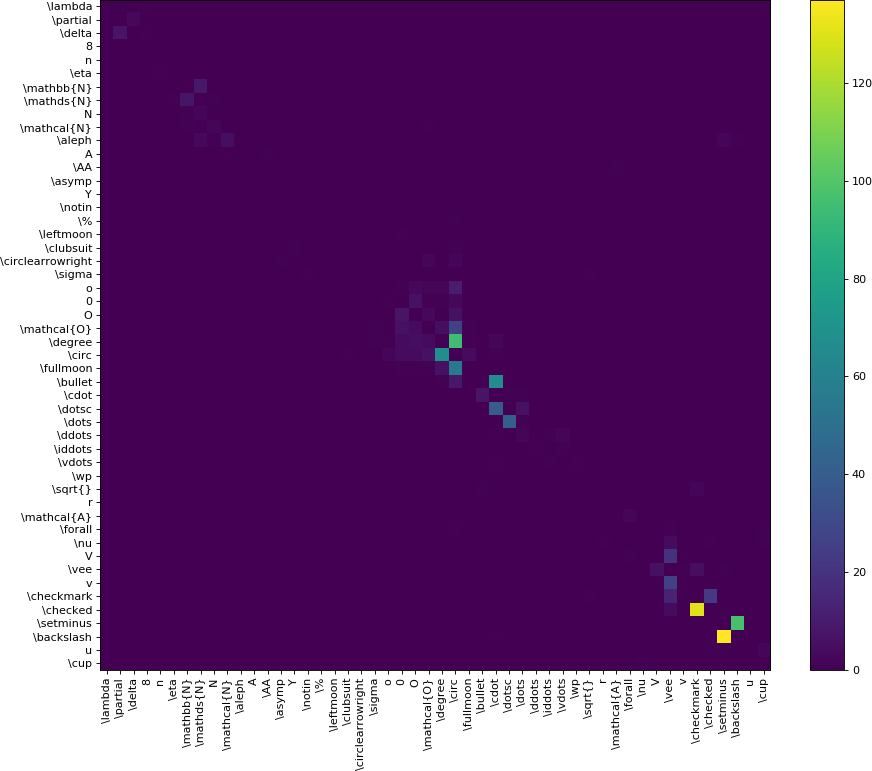}
    \caption[Confusion matrices for HASYv2]{The first 50~entries of the confusion matrix of the HASYv2
             dataset. The diagonal elements are set to~0 in order to make other
             elements easier to see. The top image shows arbitrary class
             ordering, the bottom image shows the optimized ordering. }
    \label{fig:HASY-cm-train-unsorted}
\end{figure}

On the HASYv2 dataset the class-ordering is necessary to see anything as most
possible confusions do not happen. See \cref{fig:HASY-cm-train-unsorted} for
comparison of the first 50~classes of the unsorted confusion matrix and the
sorted confusion matrix. If confusion matrices of a maximum size of $50 \times
50$ are displayed, the ordered method can show only 8~matrices because the
off-diagonal matrices are almost~0. Without sorting, 64~matrices have to be
displayed.
\clearpage
\thispagestyle{empty}
\clearpage

\section{Spectral Clustering vs CMO}\label{sec:cmo-clustering}
This section evaluates the clustering quality of \gls{CMO} in comparison to the
clustering quality of spectral clustering.

\begin{figure}[ht]
    \centering
    \includegraphics[width=0.98\paperwidth,height=0.4\paperheight,keepaspectratio]{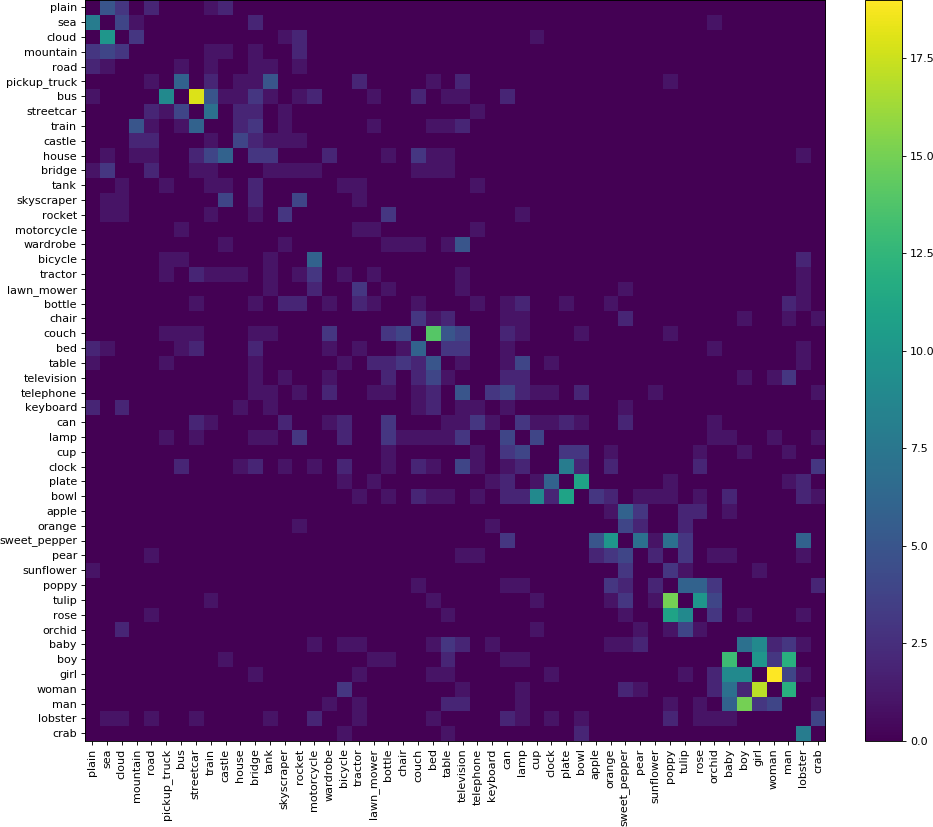}
    \caption[Confusion matrix of CIFAR-100]{The first 50~entries of the ordered
             confusion matrix of the CIFAR-100 dataset. The diagonal elements
             are set to~0 in order to make other elements easier to see. Best
             viewed in electronic form.}
    \label{fig:CIFAR-100-cm-test-sorted}
\end{figure}

The evaluated model achieves \SI{70.50}{\percent}~training accuracy and
\SI{53.16}{\percent}~test accuracy on CIFAR-100.
\Cref{fig:CIFAR-100-cm-test-sorted} shows the sorted confusion matrix.

CIFAR-100 has pre-defined coarse classes. Those are used as a ground truth for
the clusters which are to be found. The number of errors is determined by
\begin{enumerate*}[label=(\roman*)]
    \item Join all $n$ clusters which contain the classes of the coarse class
          $C$ to a set $M$. The error is $n$.
    \item Within $M$, find the set of classes $M^-$ which do not belong to $C$.
    \item The final error is $n$ + $|M^-|$.
\end{enumerate*}
As can be seen in \cref{table:clustering-differences-CIFAR-100}, both
clustering methods find reasonable clusters. \gls{CMO}, however, has only half
the error of spectral clustering.\\
The results for the HASYv2 dataset are qualitatively similar
(see~\cref{table:clustering-differences-HASYv2}). It should be noted that the
number of clusters was determined by using the semi-automatic method based on
\gls{CMO} as described in~\cref{sec:clustering-classes}.

\mathversion{normal2}
\begin{table}[H]
    \centering
    {
    \small
    \begin{tabular}{lp{0.32\linewidth}rp{0.32\linewidth}r}
    \toprule
    Cluster  & Spectral clustering                                                                & Errors & \gls{CMO}                                                                             & Errors \\\midrule
    fish     & aquarium fish, orchid + flatfish + ray, shark + trout, lion                        & 5      & aquarium fish, orchid + flatfish + ray + shark, trout                           & 4      \\
    flowers  & orchid, aquarium fish + sunflower + poppy, tulip + rose, train                     & 5      & orchid, aquarium fish + sunflower, poppy, tulip, rose                           & 2      \\
    people   & baby, boy, man + girl + woman                                                      & 2      & baby, boy, girl, woman, man                                                     & 0      \\
    reptiles & crocodile, plain, road, table, wardrobe + dinosaur + lizard + snake, worm + turtle & 9      & crocodile, lizard, lobster, caterpillar + dinosaur + snake + turtle, crab       & 6      \\
    trees    & maple, oak, pine + willow, forest + palm                                           & 3      & palm, willow, pine, maple, oak                                                  & 0      \\\midrule
    Total    & ~                                                                                  & 24     & ~                                                                               & 12     \\\bottomrule
    \end{tabular}
    }
    \caption[Clustering errors for spectral clustering and CMO on
             CIFAR-100]{Differences in spectral clustering and \gls{CMO}.
             Classes in a cluster are separated by \texttt{,} whereas clusters
             are separated by \texttt{+}.}
    \label{table:clustering-differences-CIFAR-100}
\end{table}

\mathversion{normal2}
\begin{table}[H]
    \centering
    {\small
    \begin{tabular}{clrlr}
    \toprule
    Cluster       & Spectral clustering                                                                      & Errors & \gls{CMO}                                                                               & Errors \\\midrule
    A             & $A$, $\mathcal{A}$, $\mathscr{A}$                                                        &      0 & $A$, $\mathcal{A}$, $\mathscr{A}$, \AA                                                  &      1 \\
    B             & $B$, $\mathcal{B}$                                                                       &      0 & $B$, $\mathcal{B}$                                                                      &      0 \\
    C             & $C$, $c$, $\subset$ and $\mathscr{C}$, $\xi$, $\mathscr{E}$ and $\mathcal{C}$            &      4 & $C$, $c$, $\subset$, $\mathcal{C}$ and $\mathscr{C}$                                    &      1 \\
    D             & $D$, $\mathcal{D}$, $\mathscr{D}$, $\triangleright$                                      &      1 & $D$, $\mathcal{D}$, $\mathscr{D}$                                                       &      0 \\
    E             & $E$ and $\mathcal{E}$, $\varepsilon$                                                     &      2 & $E$ and $\mathcal{E}$, $\varepsilon$, $\epsilon$, $\in$                                 &      4 \\
    F             & $F$ and $\mathcal{F}$, $\mathscr{F}$                                                     &      1 & $F$ and $\mathcal{F}$, $\mathscr{F}$                                                    &      1 \\
    H             & $H$ and $\mathscr{H}$, $\varkappa$ and $\mathcal{H}$                                     &      3 & $H$ and $\mathcal{H}$, $\mathscr{H}$                                                    &      1 \\
    K             & $K$, $\kappa$                                                                            &      0 & $K$, $\kappa$                                                                           &      0 \\
    L             & $L$, $\lfloor$ and $\mathcal{L}$, $\mathscr{L}$                                          &      1 & $L$, $\lfloor$ and $\mathcal{L}$, $\mathscr{L}$                                         &      1 \\
    M             & $M$ and $\mathcal{M}$ and $\mathfrak{M}$                                                 &      2 & $M$ and $\mu$, $\mathcal{M}$ and $\mathfrak{M}$                                         &      3 \\
    N             & $N$ and $\mathbb{N}$, $\mathds{N}$ and $\mathcal{N}$                                     &      2 & $N$ and $\mathbb{N}$, $\mathds{N}$ and $\mathcal{N}$, $\aleph$                          &      3 \\
    O             & $O$, $\mathcal{O}$, $0$, $\circ$, $\degree$, {\mbox {\wasyfamily \char 35}} and $o$      &      1 & $O$, $\mathcal{O}$, $0$, $\circ$, $\degree$  and {\mbox {\wasyfamily \char 35}} and $o$ &      2 \\
    P             & $P$, $\mathcal{P}$ and $p$, $\rho$ and $\mathscr{P}$ and $\wp$                           &      3 & $P$ and $\mathcal{P}$, $\mathscr{P}$, $\wp$ and $p$, $\rho$                             &      2 \\
    Q             & $Q$, $\mathbb{Q}$, $\mathds{Q}$, $\iota$, $\sqcup$, $\gtrsim$, $\ell$, $\Im$, \AE, $1$   &      7 & $Q$ and $\mathds{Q}$, $\mathbb{Q}$                                                      &      1 \\
    R             & $R$, $\mathcal{R}$ and $\mathbb{R}$, $\mathds{R}$, $k$ and $\Re$                         &      3 & $R$ and $\Re$, $\mathcal{R}$, $\mathds{R}$, $\mathbb{R}$                                &      1 \\
    S             & $S$, $s$, $\mathcal{S}$                                                                  &      0 & $S$, $s$, $\mathcal{S}$                                                                 &      0 \\
    T             & $T$, $\top$ and $\mathcal{T}$, $\tau$                                                    &      1 & $T$, $\top$ and $\mathcal{T}$, $\tau$                                                   &      1 \\
    U             & $U$, $\cup$ and $u$, $\mathcal{U}$, $\mathfrak{A}$                                       &      1 & $U$, $u$, $\mathcal{U}$, $\mathfrak{A}$ and $\cup$                                      &      2 \\
    V             & $V$, $v$, $\vee$                                                                         &      0 & $V$, $v$, $\vee$                                                                        &      0 \\
    W             & $W$, $w$, $\omega$                                                                       &      0 & $W$, $w$ and $\omega$                                                                   &      1 \\
    X             & $X$, $x$, $\mathcal{X}$, $\chi$, $\times$                                                &      0 & $X$, $x$, $\mathcal{X}$, $\chi$, $\times$                                               &      0 \\
    Y             & $Y$ and $y$                                                                              &      1 & $Y$, $y$                                                                                &      0 \\
    Z             & $Z$, $z$, $\mathcal{Z}$ and $\mathbb{Z}$, $\mathds{Z}$                                   &      1 & $Z$, $z$, $\mathbb{Z}$, $\mathcal{Z}$, $\mathds{Z}$                                     &      0 \\\midrule
    Total         &                                                                                          &     34 &                                                                                         &     25 \\\bottomrule
    \end{tabular}
    }
    \caption{Differences in spectral clustering and \gls{CMO}.}
    \label{table:clustering-differences-HASYv2}
\end{table}
\mathversion{normal}
\clearpage

\section{Hierarchy of Classifiers}\label{sec:hierarchy-of-classifiers}
In a first step, a classifier is trained on the 100~classes of CIFAR-100. The
fine-grained root classifier achieves an accuracy of \SI{65.29}{\percent} with
test-time transformations. The accuracy on the found sub-classes are listed
in~\cref{table:CIFAR-100-accuracies-subclasses}. The fact that the root
classifier achieves better results within a cluster than the specialized leaf
classifiers in 13 of 14~cases could either be due to limited training data,
overfitting or the small size of $\SI{32}{\pixel} \times \SI{32}{\pixel}$ of
the data. The experiment also shows that most of the errors are due to not
identifying the correct cluster. Hence, in this case, more work in improving
the root classifier is necessary rather than improving the discrimination of
classes within a cluster.\\
Although the classes within a cluster capture most of the classifications,
many misclassifications happen outside of the clusters. For example, in
cluster~3, a perfect leaf classifier would push the accuracy in the
\textit{full} column only to \SI{63.50}{\percent} due to errors of the root
classifier where the root classifier does not predict the correct cluster.\\
The leaf classifiers use the same topology as the root classifier. By
initializing them with the root classifiers weights their performance can be
pushed at about the \textit{inner} accuracy. They are, however, only useful
if their accuracy is well above the \textit{inner} accuracy of the root
classifier. Hence, for CIFAR-100, building hierarchies of classifiers is not
useful.
\begin{table}[H]
    \centering
    \begin{tabular}{rrrrl}
    \toprule
  \multirow{3}{*}{Cluster} & \multirow{3}{*}{Classes} & \multicolumn{3}{c}{accuracy} \\
          &         & \multicolumn{2}{c}{root classifier}             & leaf classifier \\\cline{3-4}
          &         & cluster identified   & class identified | cluster & class identified | cluster \\\midrule
    1     & 3       & \SI{69.67}{\percent} & \SI{84.27}{\percent}     & \SI{72.98}{\percent} \\
    2     & 5       & \SI{46.60}{\percent} & \SI{58.54}{\percent}     & \SI{43.47}{\percent} \\
    3     & 2       & \SI{58.50}{\percent} & \SI{92.13}{\percent}     & \SI{83.46}{\percent}\\
    4     & 2       & \SI{50.50}{\percent} & \SI{87.83}{\percent}     & \SI{81.74}{\percent}\\
    5     & 3       & \SI{44.67}{\percent} & \SI{79.29}{\percent}     & \SI{71.01}{\percent}\\
    6     & 2       & \SI{29.50}{\percent} & \SI{78.67}{\percent}     & \SI{72.00}{\percent}\\
    7     & 2       & \SI{52.50}{\percent} & \SI{92.11}{\percent}     & \SI{87.72}{\percent}\\
    8     & 2       & \SI{59.50}{\percent} & \SI{86.23}{\percent}     & \SI{81.88}{\percent}\\
    9     & 2       & \SI{59.00}{\percent} & \SI{90.08}{\percent}     & \SI{87.79}{\percent}\\
    10    & 2       & \SI{62.00}{\percent} & \SI{85.52}{\percent}     & \SI{73.10}{\percent}\\
    11    & 2       & \SI{67.00}{\percent} & \SI{87.01}{\percent}     & \SI{75.32}{\percent}\\
    12    & 2       & \SI{72.50}{\percent} & \SI{94.77}{\percent}     & \SI{76.77}{\percent}\\
    13    & 2       & \SI{64.00}{\percent} & \SI{82.58}{\percent}     & \SI{86.27}{\percent}\\
    14    & 2       & \SI{79.67}{\percent} & \SI{89.85}{\percent}     & \SI{89.10}{\percent}\\
    \bottomrule
    \end{tabular}
    \caption[Accuracies for hierarchy of classifiers on CIFAR-100]{Accuracies
             of the root classifier trained on the full set of 100~classes
             evaluated on 14~clusters of classes. Each class has 100~elements
             to test. The column \textit{cluster identified} gives the percentage
             that the root classifiers argmax prediction is within the correct cluster,
             but not necessarily the correct class. The columns
             \textit{class identified | cluster} only consider data points
             where the root classifier correctly identified the cluster.}
    \label{table:CIFAR-100-accuracies-subclasses}
\end{table}

\section{Increased width for faster learning}\label{sec:05-network-width}
\vspace{-1cm}
More filters in one layer could simplify the optimization problem as each
filter needs smaller updates. Hence a \gls{CNN} $N$ with $n_i$~filters in
layer~$i$ is expected to take more epochs than a \gls{CNN}~$N'$ with $2 \cdot
n_i$ filters in layer~$i$ to achieve the same validation
accuracy.

This hypothesis can be falsified by training a \gls{CNN}~$N$ and a
\gls{CNN}~$N'$ and comparing the trained number of epochs. As more filters can
lead to different results depending on the layer where they are added, five
models are trained. The details about those models are given
in~\cref{table:increased-model-width-models}

\begin{table}[ht]
    \centering
    \begin{tabular}{lrrrr}
    \toprule
    \multirow{2}{*}{Name} & \multirow{2}{*}{Layer} &  \multicolumn{2}{c}{Filter count} & Total \\
             &       & Baseline              &  New & parameters \\\midrule
    $m_9$    &  9    &  64                   &  638 & \num{5978566}    \\
    $m_9'$   &  9    &  64                   &  974 & \num{8925622}    \\
    $m_{11}$ & 11    & 512                   & 3786 & \num{5982698}    \\
    $m_{11}'$& 11    & 512                   & 1024 & \num{1731980}    \\
    $m_{13}$ & 13    & 512                   & 8704 & \num{5982092}    \\
    \bottomrule
    \end{tabular}
    \caption[Parameters of models with increased capacity]{Models which are
             identical to the baseline, except that the number of filters of
             one layer was increased.}
    \label{table:increased-model-width-models}
\end{table}

The detailed results are given in~\cref{table:CIFAR-100-increased-model-size}.
As expected, the number of training epochs of the models with increased numbers
of parameters is lower. The wall-clock time, however, is higher due to the
increase in computation per forward- and backward-pass.

For $m_9$, $m_{11}$ and $m_{13}$, the filter weight range of the layer with
increased capacity decreases compared
to~\cref{fig:baseline-filter-weight-range}, the filter weights of the layer
with increased capacity are more concentrated around zero compared
to~\cref{fig:baseline-filter-weight-dist}. For model $m_{13}$, the distribution
of weight of the output layer changed to a more bell-shaped distribution.
Except for this, the distribution of filter weights in other layers did not
change for all three models compared to the baseline.

\begin{table}[H]
    \centering
    \begin{tabular}{lrrrrrr}
    \toprule
    \multirow{3}{*}{Model} & \multirow{3}{*}{Parameters}     & \multicolumn{3}{c}{Accuracy} & \multicolumn{2}{c}{Training}\\
               &                & \multicolumn{2}{c}{Single Model}& Ensemble & Mean Epochs & Mean Time\\
               &                & Mean  & std  &          &  & \\\midrule
    baseline   &  \num{944012}  & \SI{63.38}{\percent} & 0.55 & \SI{64.70}{\percent}     & 154.7       & \SI{3856}{\second} \\
    $m_9$      & \num{5978566}  & \SI{65.53}{\percent} & \textbf{0.37} & \SI{66.72}{\percent} & 105.7  & \SI{4472}{\second} \\
    $m_9'$     & \num{8925622}  & \SI{65.10}{\percent} & 1.09 & \SI{66.54}{\percent}    & \textbf{95.6}& \SI{5261}{\second} \\
    $m_{11}$   & \num{5982698}  & \textbf{\SI{65.73}{\percent}} & 0.77 & \textbf{\SI{67.38}{\percent}} & 149.2       & \SI{5450}{\second}\\
    $m_{11}'$  & \num{1731980}  & \SI{62.12}{\percent} & 0.48 & \SI{62.89}{\percent}    & 143.6        & \textbf{\SI{3665}{\second}} \\
    $m_{13}$   & \num{5982092}  & \SI{62.39}{\percent} & 0.66 & \SI{63.77}{\percent}    & 147.8        & \SI{4485}{\second} \\
    \bottomrule
    \end{tabular}
    \caption[Training time for models with increased capacity]{Training time in
             epochs and wall-clock time for the baseline and models $m_9$,
             $m_{11}$, $m_{13}$ as well as their accuracies.}
    \label{table:CIFAR-100-increased-model-size}
\end{table}

\section{Weight updates}\label{sec:weight-updates}
\Cref{sec:05-network-width} shows that wider networks learn faster. One
hypothesis why this happens is that every single weight updates can be smaller
to learn the same function. Thus the loss function is smoother and thus
gradient descent based optimization algorithms lead to more consistent weight
updates.

Consequently, it is expected that layers with fewer filters have more erratic
updates. If there are many filters, the weights of a filter which does not
contribute much to the end results or is even harmful filter can gradually be
set to zero, essentially removing one path in the network.

\begin{figure}[ht]
    \centering
    \includegraphics[width=0.98\linewidth,keepaspectratio]{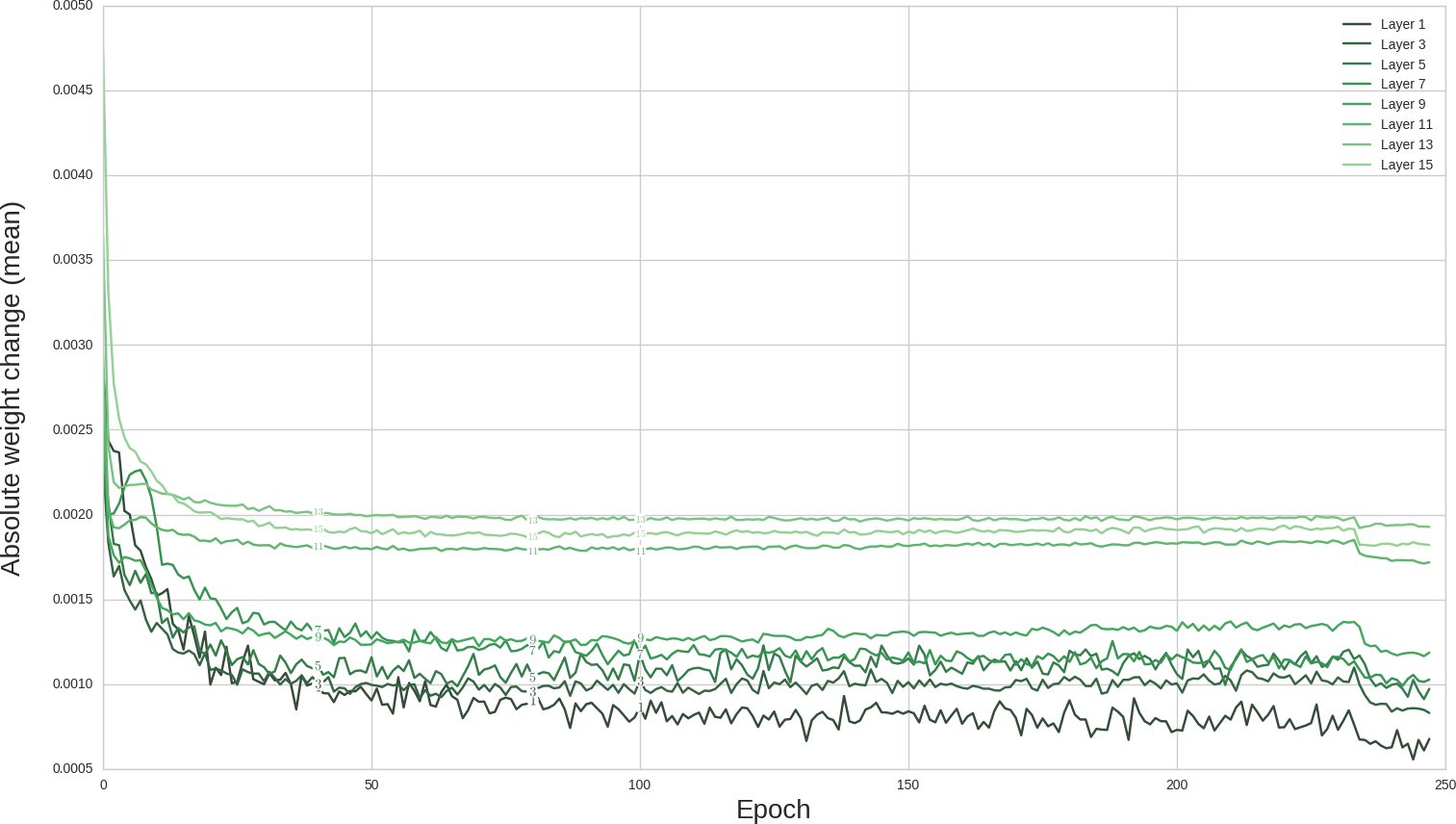}
    \caption[Mean weight updates of model with bottleneck]{Mean weight updates between epochs by layer. The model is the
             baseline model, but with layer~5 reduced to 3~filters.}
    \label{fig:erratic-mean-weight-updates-mean}
\end{figure}

In order to test the hypothesis, the baseline model was adjusted.
The number of filters in layer~5 was reduced from 64~filters to 3~filters.
As one can see in~\cref{fig:erratic-mean-weight-updates-mean}, the mean weight
update of the layers 1, 3, 5, 7 and 9 have a far bigger range than the layers
11, 13 and 15 after epoch~50. Compared to the baseline models mean updates
(\cref{fig:baseline-weight-updates-mean},
\cpageref{fig:baseline-weight-updates-mean}), the mean weight updates of layers~1
and 3 are higher, the range of the mean weight update from epoch~50 is higher
for layer~5 and the range of mean updates of layer~7 is higher.

For the maximum and the sum, no similar pattern could be observed (see \cref{fig:erratic-weight-updates-max,fig:erratic-weight-updates-sum}).

\section{Multiple narrow layers vs One wide layer}\label{sec:multiple-narrow-one-wide}
On a given feature map size one can have an arbitrary number of convolutional
layers with \texttt{SAME} padding and each layer can have an arbitrary number
of filters. A convolutional layer with more filters is called
\textit{wider}~\cite{zagoruyko2016wide}, a convolutional layer with fewer
filters is thus called narrower and the number of filters in a convolutional
layer is the layers \textit{width}.

If the number of parameters which may be used for the feature map scale is
fixed and high enough, there are still many combinations. If $n_i$ with $i = 0,
\dots, k$ is the number of output feature maps of layer~$i$ where $i = 0$ is
the input layer and all filters are $3 \times 3$ filters without a bias, then
the number of parameters is
\[\text{Parameters} = \sum_{i=1}^k \left ((n_{i-1} \cdot 3^2 + 1) \cdot n_{i}\right )\]
Hence the width of one layer does not only influence the parameters in this
layer, but also in the next layer.

The number of possible subsequent layers of one feature map size is enormous,
even if constraints are placed on the number of parameters. For example, the
first convolutional layer of the baseline model has 896~parameters. If one
assumes that less than 3~filters per layer are not desirable, one keeps all
layers having a bias and all layers only use $3\times 3$ filters, then the
maximum depth is~10. If one furthermore assumes that at least 800~parameters
should be used, there are still 120~possible layer combinations. As
experimentally evaluating one layer combination takes about 10~hours on a GTX
970 for CIFAR-100 it is not possible to evaluate all layer combinations. In
the following, a couple of changes to the network width / depth will be
evaluated.

Each layer expands the perceptive field. Hence deeper layer can use more of the
input for every single output value. But deeper networks need more time for
inference as the output of layer~$i$ has to be computed before the output
of~$i+1$ can be computed. Hence there is less potential to parallelize
computations. Each filter can be seen as a concept which can be learned. The
deeper the filter is in the network, the higher is the abstraction level of the
concept. In most cases, both is necessary: Many different concepts (width) and
high-level concepts (depth).

Reducing the two first convolutional layers of the baseline model
(see~\cpageref{fig:baseline-architecture}) to one convolutional layer of
48~filters (\num{944396}~parameters in total, whereas the baseline model
has \num{944012}~parameters) resulted in a mean accuracy of \SI{61.64}{\percent}
(-\SI{1.74}{\percent}) and a standard deviation of $\sigma=1.12$ (+0.57).
The ensemble achieved \SI{63.18}{\percent} (-\SI{1.52}{\percent}). As expected,
the training time per epoch was reduced. For the GTX~980, it was reduced from
\SI{22.0}{\second} of the baseline model to \SI{15}{\second} of the model
with one less convolutional layer, one less Batch Normalization and one less
activation layer. The inference time was also reduced from
\SI{6}{\milli\second} to \SI{4}{\milli\second} for 1~image and from
\SI{32}{\milli\second} to \SI{23}{\milli\second} for 128~images. Due to the
loss in accuracy of more then one percentage point of the mean model and the
increased standard deviation of the models performance, at least two
convolutional layers are on the $\SI{32}{\pixel} \times \SI{32}{\pixel}$
feature map scale are recommendable for CIFAR-100.

Changing the baseline to have less filters but more layers is another option.
This was tried for the first block at the $\SI{32}{\pixel} \times \SI{32}{\pixel}$
feature map scale. The two convolutional layers (layers~1 -- 4 in~\cpageref{table:baseline-architecture})
were replaced by two convolutional layers with~27 filters and one convolutional
layer with 26~filters in the \texttt{convolution - BN - ELU} pattern. The
model has \num{944132} parameters. Compared to the baseline model, the time
for inference was the same. This is unexpected, because the inference time
changed when a layer was removed at this scale. The mean test accuracy was
\SI{63.66}{\percent} (+0.28) and the standard deviation was $\sigma=1.03$ (+0.48).
The ensemble achieved \SI{64.91}{\percent} test accuracy (+0.21).

Having two nonlinearities at each feature map scale could be important to
learn nonlinear transformations at that scale. As the baseline model does only
have one nonlinearity at the $8 \times 8$ feature maps scale, another
convolutional layer with 64~filters, Batch Normalization and \gls{ELU} was
added. To keep the number of parameters constant, layer~11 of the baseline
model was reduced from 512~filters to 488~filters. The new model achieves a
mean accuracy of \SI{63.09}{\percent} (-0.29) with a standard deviation of
$\sigma=0.70$ (+0.15). The ensemble achieves an accuracy of
\SI{64.39}{\percent} (+0.31). This could indicate that having two convolutional
layers is more important for layers close to the input than intermediate layer.
Alternatively, the parameters could be more important in layer~11 than having
a new convolutional layer after layer~9.

In order to control the hypothesis that having two convolutional layers are
less important in the middle of a network, the second convolutional layer at
the $16 \times 16$ feature map scale is removed. The first convolutional layer
was increased from 32~filters to 59~filters, the second convolutional layer was
increased from 32~filter s to 58~filters in order to keep the amount of
parameters of the model constant. The adjusted model achieved
\SI{62.72}{\percent} (-0.66) mean test accuracy with a standard deviation of
$\sigma=0.84$ (+0.29). The ensemble achieved \SI{63.88}{\percent} test accuracy
(-0.66).

Even more extreme, if both convolutional layers are removed from the $16 \times
16$ feature map scale, the mean test accuracy drops to \SI{61.21}{\percent}
(-2.17) with a standard deviation of $\sigma = 0.51$ (-0.04). The ensemble
achieves a test accuracy of \SI{63.07}{\percent} (-1.63). Thus it is very
important to have at least one convolutional layer at this feature map scale.

\section{Batch Normalization}\label{sec:05-batch-normalizaiton}
In~\cite{clevert2015fast}, the authors write that Batch Normalization does not
improve \gls{ELU} networks. Hence the effect of removing Batch Normalization
from the baseline is investigated in this experiment.

As before, 10~models are trained on CIFAR-100. The training setup and the model
$m_{\text{no-bn}}$ are identical to the baseline model $m$, except that in
$m_{\text{no-bn}}$ the Batch Normalization layers are removed.

One notable difference is the training time: While $m$ needs
\SI{21}{\milli\second} per epoch in average on a GTX~980, $m_{\text{no-bn}}$
only needs \SI{21}{\milli\second} per epoch. The number of epochs used for
training, however, also increased noticeably from 149~epochs to 178~epochs in
average. The standard deviation of trained epochs is 17.3~epochs for the
baseline model and 23.4~epochs for $m_{\text{no-bn}}$.

The mean accuracy of $m_{\text{no-bn}}$ is \SI{62.86}{\percent} and hence
0.52~percentage points worse. The standard deviation between models increased
from 0.55 to 0.61. This is likely a result of the early stopping policy and the
differences in training epochs. This can potentially be fixed by retraining the
models which stopped earlier than the model which was trained for the biggest
amount of epochs. The ensemble test accuracy is \SI{63.88}{\percent} and hence
0.82~percentage points worse than the baseline.

The filter weight range and distribution is approximately the same
as~\cref{fig:baseline-filter-weight-range} and
\cref{fig:baseline-filter-weight-dist}, but the distribution of bias weights
changed noticeably: While the bias weights of the baseline are spread out in
the first layer and much more concentrated in subsequent layers
(see~\cref{fig:baseline-bias-weight-dist}), the model without Batch
Normalization has rather concentrated weights in the first layers and only the
bias weights of the last layer is spread out (see~\cref{fig:no-bn-bias-weight-dist}).

Another model $m_{\text{no-bn}}'$ which has one more filter in the
convolutional layer~1, 3, 5, and 7 to compensate for the loss of parameters in
Batch Normalization. The mean test accuracy of 10 such models is
\SI{62.87}{\percent} which is 0.51~percentage points worse than the baseline.
The ensemble of $m_{\text{no-bn}}'$ achieves \SI{64.33}{\percent} which is
0.37~percentage points worse than the baseline. The mean training time was
\SI{14}{\second} per epoch and 157.4 epochs with a standard deviation of
20.7~epochs.

Hence it is not advisable to remove Batch Normalization for the final model. It
could, however, be possible to remove Batch Normalization for the experiments
to iterate quicker through different ideas if the relative performance changes
behave the same with or without Batch Normalization.

\clearpage
\section{Batch size}\label{sec:05-batch-size}
\vspace*{-0.5cm}The mini-batch size $m \in \mathbb{N}_{\geq 1}$ influences
\begin{itemize}
    \item \textbf{Epochs until convergence}: The smaller $m$, the more often
          the model is updated in one epoch. Those updates, however, are based
          on fewer samples of the dataset. Hence the gradients of different
          mini-batches can noticeably differ. In the literature, this is
          referred to as gradient noise~\cite{keskar2016large}.
    \item \textbf{Training time per epoch}: The smaller the batch size, the
          higher the training time per epoch as the hardware is not optimally
          utilized.
    \item \textbf{Resulting model quality}: The choice of the
          hyperparameter~$m$ influences the accuracy of the classifier when
          training is finished. \cite{keskar2016large}~supports the view that
          smaller $m$ result in less sharp minima. Hence smaller~$m$ lead to
          better generalization.
\end{itemize}

Empiric evaluation results can be found
in~\cref{table:CIFAR-100-accuracies-batchsize}. Those results confirm the claim
of~\cite{keskar2016large} that lower batch sizes generalize better.

\begin{table}[H]
    \centering
    \begin{tabular}{rrrrrrr}
    \toprule
    \multirow{2}{*}{$m$}
       &\multicolumn{1}{c}{Training}& \multirow{2}{*}{Epochs} & \multicolumn{1}{c}{Mean total} & \multicolumn{2}{c}{Single model} & \multicolumn{1}{c}{Ensemble} \\\cline{5-6}
       &\multicolumn{1}{c}{time}    &            & \multicolumn{1}{c}{training time} & \multicolumn{1}{c}{Accuracy} & \multicolumn{1}{c}{std} & \multicolumn{1}{c}{Accuracy} \\\midrule
     8 &\SI{118}{\second\per\epoch} & \textbf{81} -- \textbf{153} & \SI{14131}{\second} & \SI{61.93}{\percent} & $\sigma=1.03$ & \SI{65.68}{\percent}\\
    16 & \SI{62}{\second\per\epoch} & 103 -- 173 &  \SI{8349}{\second} & \textbf{\SI{64.16}{\percent}} & $\sigma=0.81$ & \textbf{\SI{66.98}{\percent}}\\
    32 & \SI{35}{\second\per\epoch} & 119 -- 179 &  \SI{5171}{\second} & \SI{64.11}{\percent} & $\sigma=0.75$ & \SI{65.89}{\percent}\\
    64 & \SI{25}{\second\per\epoch} & 133 -- 195 & \textbf{\SI{2892}{\second}} & \SI{63.38}{\percent} & $\boldsymbol{\sigma=0.55}$ & \SI{64.70}{\percent}\\
   128 & \SI{18}{\second\per\epoch} & 145 -- 239 &  \SI{3126}{\second} & \SI{62.23}{\percent} & $\sigma=0.73$ & \SI{63.55}{\percent}\\\bottomrule
    \end{tabular}
    \caption[Baseline model training time]{Training time per epoch and single
             model test set accuracy (mean and standard deviation) of baseline
             models trained with different mini-batch sizes~$m$ on GTX~970
             \glspl{GPU} on CIFAR-100.}
    \label{table:CIFAR-100-accuracies-batchsize}
\end{table}

\section{Bias}\label{sec:05-bias}
\vspace*{-0.5cm}\Cref{fig:baseline-bias-weight-dist} suggests that the bias is not important
for the layers~11, 13 and 15. Hence a model $m_{\text{no-bias}}$ is created
which is identical to the baseline model $m$, except that the bias of layers
11, 13 and 15 is removed.

The mean test accuracy of 10~trained $m_{\text{no-bias}}$ is
\SI{63.74}{\percent} which is an improvement of 0.36~percentage points over the
baseline. The ensemble achieves a test accuracy of \SI{65.13}{\percent} which
is 0.43 percentage points better than the baseline. Hence the bias can safely
be removed.

Removing the biases did not have a noticeable effect on the filter weight range,
the filter weight distribution or the distribution of the remaining biases.
Also, the $\gamma$ and $\beta$ parameters of the Batch Normalization layers did
not noticeably change.
\goodbreak

\section{Learned Color Space Transformation}\label{sec:05-color-space-transformations}
In~\cite{mishkin2016systematic} it is described that placing one convolutional
layer with 10~filters of size $1 \times 1$ directly after the input and then
another convolutional layer with 3~filters of size $1 \times 1$ acts as a
learned transformation in another color space and boosts the accuracy.

This approach was evaluated on CIFAR-100 by adding a convolutional layer with
ELU~activation and 10~filters followed by another convolutional layer with ELU
activation and 3~filters. The mean accuracy of 10~models was
\SI{63.31}{\percent} with a standard deviation of~1.37. The standard deviation
is noticeable higher than the standard deviation of the baseline model (0.55)
and the accuracy also decreased by 0.07~percentage points. The accuracy of
the ensemble is at \SI{64.77}{\percent} and hence 0.07~percentage points higher
than the accuracy of the baseline models.

The inference time for 1~image and for 128~images did not change compared to
the baseline. The training time per epoch increased from \SI{26}{\second} to
\SI{30}{\second} on the GTX~970.

Hence it is not advisable to use the learned color space transformation.

\section{Pooling}\label{sec:05-pooling}
An alternative to max pooling with stride~2 with a $2\times 2$ kernel is using
a $3 \times 3$ kernel with stride~2.

This approach was evaluated on CIFAR-100 by replacing all max pooling layers
with the $3 \times 3$ kernel max pooling (and \texttt{SAME} padding). The mean
accuracy of 10~models was \SI{63.32}{\percent} ($-0.06$) and the
standard deviation was~0.57 ($+0.02$). The ensemble achieved
$\SI{65.15}{\percent}$ test accuracy ($+0.45$).

The training time per epoch decreased from \SIrange{20.5}{21.1}{\second} to
\SI{18.6}{\second} (mean of 10 training runs) on the Nvidia GTX~970. The time
for inference increased from \SI{25}{\milli\second} to \SI{26}{\milli\second}
for a batch of 128~images.

\section{Activation Functions}
Nonlinear, differentiable activation functions are important for neural
networks to allow them to learn nonlinear decision boundaries. One of the
simplest and most widely used activation functions for \glspl{CNN} is
\gls{ReLU}~\cite{AlexNet-2012}, but others such as
\gls{ELU}~\cite{clevert2015fast}, \gls{PReLU}~\cite{he2015delving}, softplus~\cite{7280459}
and softsign~\cite{bergstra2009quadratic} have been proposed. The baseline uses
\gls{ELU}.

Activation functions differ in the range of values and the derivative. The
definitions and other comparisons of eleven activation functions are given
in~\cref{table:activation-functions-overview}.

Theoretical explanations why one activation function is preferable to another
in some scenarios are the following:
\begin{itemize}
    \item \textbf{Vanishing Gradient}: Activation functions like tanh and the
          logistic function saturate outside of the interval $[-5, 5]$. This
          means weight updates are very small for preceding neurons, which is
          especially a problem for very deep or recurrent networks as described
          in~\cite{bengio1994learning}. Even if the neurons learn eventually,
          learning is slower~\cite{AlexNet-2012}.
    \item \textbf{Dying ReLU}: The dying \gls{ReLU} problem is similar to the
          vanishing gradient problem. The gradient of the \gls{ReLU} function
          is~0 for all non-positive values. This means if all elements of the
          training set lead to a negative input for one neuron at any point in
          the training process, this neuron does not get any update and hence
          does not participate in the training process. This problem is
          addressed in~\cite{maas2013rectifier}.
    \item \textbf{Mean unit activation}: Some publications
          like~\cite{clevert2015fast,BatchNormalization-2015} claim that mean
          unit activations close to 0 are desirable. They claim that this
          speeds up learning by reducing the bias shift effect. The speedup
          of learning is supported by many experiments. Hence the possibility
          of negative activations is desirable.
\end{itemize}

Those considerations are listed
in~\cref{table:properties-of-activation-functions} for 11~activation functions.
Besides the theoretical properties, empiric results are provided
in~\cref{table:CIFAR-100-accuracies-activation-functions,table:CIFAR-100-timing-activation-functions}.
The baseline network was adjusted so that every activation function except the
one of the output layer was replaced by one of the 11~activation functions.

As expected, \gls{PReLU} and \gls{ELU} performed best. Unexpected was that the
logistic function, tanh and softplus performed worse than the identity and it
is unclear why the pure-softmax network performed so much better than the
logistic function.
One hypothesis why the logistic function performs so bad is that it cannot
produce negative outputs. Hence the logistic$^-$ function was developed:
\[\text{logistic}^-(x) = \frac{1}{1+ e^{-x}} - 0.5\]
The logistic$^-$ function has the same derivative as the logistic function and
hence still suffers from the vanishing gradient problem.
The network with the logistic$^-$ function achieves an accuracy which is
\SI{11.30}{\percent} better than the network with the logistic function, but is
still \SI{5.54}{\percent} worse than the \gls{ELU}.

Similarly, \gls{ReLU} was adjusted to have a negative output:
\[\text{ReLU}^-(x) = \max(-1, x) = \text{ReLU}(x+1) - 1\]
The results of \gls{ReLU}$^-$ are much worse on the training set, but perform
similar on the test set. The result indicates that the possibility of hard zero
and thus a sparse representation is either not important or similar important as
the possibility to produce negative outputs. This
contradicts~\cite{glorot2011deep,srivastava2014understanding}.

A key difference between the logistic$^-$ function and \gls{ELU} is that
\gls{ELU} does neither suffers from the vanishing gradient problem nor is its
range of values bound. For this reason, the S2ReLU activation function, defined
as
\[\text{S2ReLU}(x) = ReLU(\frac{x}{2} + 1) - ReLU(-\frac{x}{2} + 1) =
  \begin{cases}-\frac{x}{2} + 1 &\text{if } x \le -2\\
               x &\text{if } -2\le x \le 2\\
               \frac{x}{2} + 1&\text{if } x > -2\end{cases}\]
This function is similar to SReLUs as introduced in~\cite{jin2016deep}. The
difference is that S2ReLU does not introduce learnable parameters. The S2ReLU
was designed to be symmetric, be the identity close to zero and have a smaller
absolute value than the identity farther away. It is easy to compute and easy to
implement.

Those results --- not only the absolute values, but also the relative
comparison --- might depend on the network architecture, the training
algorithm, the initialization and the dataset. Results for MNIST can be found
in~\cref{table:MNIST-accuracies-activation-functions} and for HASYv2
in~\cref{table:HASYv2-accuracies-activation-functions}. For both datasets, the
logistic function has a much shorter training time and a noticeably lower test
accuracy.

\begin{table}[H]
    \centering
    \begin{tabular}{lccc}
    \toprule
    Function      & Vanishing Gradient        & Negative Activation possible & Bound activation \\\midrule
    Identity      & \cellcolor{green!25}No    & \cellcolor{green!25}  Yes    & \cellcolor{green!25}No  \\
    Logistic      & \cellcolor{red!25} Yes    & \cellcolor{red!25}   No      & \cellcolor{red!25}  Yes \\
    Logistic$^-$  & \cellcolor{red!25} Yes    & \cellcolor{green!25}  Yes    & \cellcolor{red!25}  Yes \\
    Softmax        & \cellcolor{red!25} Yes    & \cellcolor{green!25}  Yes    & \cellcolor{red!25}  Yes \\
    tanh          & \cellcolor{red!25} Yes    & \cellcolor{green!25}  Yes    & \cellcolor{red!25}  Yes \\
    Softsign      & \cellcolor{red!25} Yes    & \cellcolor{green!25}Yes      & \cellcolor{red!25}   Yes \\
    ReLU          & \cellcolor{yellow!25}Yes\footnotemark & \cellcolor{red!25} No & \cellcolor{yellow!25}Half-sided \\
    Softplus      & \cellcolor{green!25}No    & \cellcolor{red!25}   No      & \cellcolor{yellow!25}Half-sided \\
    S2ReLU        & \cellcolor{green!25}No    & \cellcolor{green!25}Yes      & \cellcolor{green!25} No \\
    LReLU/PReLU   & \cellcolor{green!25}No    & \cellcolor{green!25}Yes      & \cellcolor{green!25} No \\
    ELU           & \cellcolor{green!25}No    & \cellcolor{green!25}Yes      & \cellcolor{green!25} No \\
    \bottomrule
    \end{tabular}
    \caption[Activation function properties]{Properties of activation functions.}
    \label{table:properties-of-activation-functions}
\end{table}
\footnotetext{The dying ReLU problem is similar to the vanishing gradient problem.}

\glsunset{LReLU}
\begin{table}[H]
    \centering
    \begin{tabular}{@{\extracolsep{4pt}}lcccccc@{}}
    \toprule
    \multirow{2}{*}{Function} & \multicolumn{4}{c}{Single model}                                                    & \multicolumn{2}{c}{Ensemble of 10} \\\cline{2-3}\cline{4-5}\cline{6-7}
                   & \multicolumn{2}{c}{Training set}     &\multicolumn{2}{c}{Test set}                  & Training set         & Test set \\\midrule
    Identity       & \SI{66.25}{\percent} & $\boldsymbol{\sigma=0.77}$ &\SI{56.74}{\percent} & \textbf{$\sigma=0.51$} & \SI{68.77}{\percent} & \SI{58.78}{\percent}\\
    Logistic       & \SI{51.87}{\percent} & $\sigma=3.64$ &\SI{46.54}{\percent} & $\sigma=3.22$          & \SI{61.19}{\percent} & \SI{54.58}{\percent}\\
    Logistic$^-$   & \SI{66.49}{\percent} & $\sigma=1.99$ &\SI{57.84}{\percent} & $\sigma=1.15$          & \SI{69.04}{\percent} & \SI{60.10}{\percent}\\
    Softmax        & \SI{75.22}{\percent} & $\sigma=2.41$ &\SI{59.49}{\percent} & $\sigma=1.25$          & \SI{78.87}{\percent} & \SI{63.06}{\percent}\\
    Tanh           & \SI{67.27}{\percent} & $\sigma=2.38$ &\SI{55.70}{\percent} & $\sigma=1.44$          & \SI{70.21}{\percent} & \SI{58.10}{\percent}\\
    Softsign       & \SI{66.43}{\percent} & $\sigma=1.74$ &\SI{55.75}{\percent} & $\sigma=0.93$          & \SI{69.78}{\percent} & \SI{58.40}{\percent}\\
    \gls{ReLU}     & \SI{78.62}{\percent} & $\sigma=2.15$ &\SI{62.18}{\percent} & $\sigma=0.99$          & \SI{81.81}{\percent} & \SI{64.57}{\percent}\\
    \gls{ReLU}$^-$ & \SI{76.01}{\percent} & $\sigma=2.31$ &\SI{62.87}{\percent} & $\sigma=1.08$          & \SI{78.18}{\percent} & \SI{64.81}{\percent}\\
    Softplus       & \SI{66.75}{\percent} & $\sigma=2.45$ &\SI{56.68}{\percent} & $\sigma=1.32$          & \SI{71.27}{\percent} & \SI{60.26}{\percent}\\
    S2ReLU         & \SI{63.32}{\percent} & $\sigma=1.69$ &\SI{56.99}{\percent} & $\sigma=1.14$          & \SI{65.80}{\percent} & \SI{59.20}{\percent}\\
    \gls{LReLU}    & \SI{74.92}{\percent} & $\sigma=2.49$ &\SI{61.86}{\percent} & $\sigma=1.23$          & \SI{77.67}{\percent} & \SI{64.01}{\percent}\\
    \gls{PReLU}    & \textbf{\SI{80.01}{\percent}} & $\sigma=2.03$ &\SI{62.16}{\percent} & $\sigma=0.73$ & \textbf{\SI{83.50}{\percent}} & \textbf{\SI{64.79}{\percent}}\\
    \gls{ELU}      & \SI{76.64}{\percent} & $\sigma=1.48$ &\textbf{\SI{63.38}{\percent}} & $\sigma=0.55$ & \SI{78.30}{\percent} & \SI{64.70}{\percent}\\
    \bottomrule
    \end{tabular}
    \caption[Activation function evaluation results on CIFAR-100]{Training and
             test accuracy of adjusted baseline models trained with different
             activation functions on CIFAR-100. For LReLU, $\alpha = 0.3$ was
             chosen.}
    \label{table:CIFAR-100-accuracies-activation-functions}
\end{table}

\begin{table}[H]
    \centering
    \begin{tabular}{lccclllll}
    \toprule
    \multirow{2}{*}{Function} & \multicolumn{2}{c}{Inference per}                                & Training                            & \multirow{2}{*}{Epochs} & Mean total        \\\cline{2-3}
                              & 1 Image                        & 128                             & time                                &                         & training time     \\\midrule
    Identity                  & \SI{8}{\milli\second}          & \SI{42}{\milli\second}          & \SI{31}{\second\per\epoch}          & 108 -- \textbf{148}     &\SI{3629}{\second} \\
    Logistic                  & \SI{6}{\milli\second}          & \textbf{\SI{31}{\milli\second}} & \SI{24}{\second\per\epoch}          & \textbf{101} -- 167     &\textbf{\SI{2234}{\second}} \\
    Logistic$^-$              & \SI{6}{\milli\second}          & \textbf{\SI{31}{\milli\second}} & \textbf{\SI{22}{\second\per\epoch}} & 133 -- 255              &\SI{3421}{\second} \\
    Softmax                   & \SI{7}{\milli\second}          & \SI{37}{\milli\second}          & \SI{33}{\second\per\epoch}          & 127 -- 248              &\SI{5250}{\second} \\
    Tanh                      & \SI{6}{\milli\second}          & \textbf{\SI{31}{\milli\second}} & \SI{23}{\second\per\epoch}          & 125 -- 211              &\SI{3141}{\second} \\
    Softsign                  & \SI{6}{\milli\second}          & \textbf{\SI{31}{\milli\second}} & \SI{23}{\second\per\epoch}          & 122 -- 205              &\SI{3505}{\second} \\
    \gls{ReLU}                & \SI{6}{\milli\second}          & \textbf{\SI{31}{\milli\second}} & \SI{23}{\second\per\epoch}          & 118 -- 192              &\SI{3449}{\second} \\
    Softplus                  & \SI{6}{\milli\second}          & \textbf{\SI{31}{\milli\second}} & \SI{24}{\second\per\epoch}          & \textbf{101} -- 165     &\SI{2718}{\second} \\
    S2ReLU                    & \textbf{\SI{5}{\milli\second}} & \SI{32}{\milli\second}          & \SI{26}{\second\per\epoch}          & 108 -- 209              &\SI{3231}{\second} \\
    \gls{LReLU}               & \SI{7}{\milli\second}          & \SI{34}{\milli\second}          & \SI{25}{\second\per\epoch}          & 109 -- 198              &\SI{3388}{\second} \\
    \gls{PReLU}               & \SI{7}{\milli\second}          & \SI{34}{\milli\second}          & \SI{28}{\second\per\epoch}          & 131 -- 215              &\SI{3970}{\second} \\
    \gls{ELU}                 & \SI{6}{\milli\second}          & \textbf{\SI{31}{\milli\second}} & \SI{23}{\second\per\epoch}          & 146 -- 232              &\SI{3692}{\second} \\
    \bottomrule
    \end{tabular}
    \caption[Activation function timing results on CIFAR-100]{Training time and
             inference time of adjusted baseline models trained with different
             activation functions on GTX~970 \glspl{GPU} on CIFAR-100. It was
             expected that the identity is the fastest function. This result is
             likely an implementation specific problem of Keras~2.0.4 or
             Tensorflow~1.1.0.}
    \label{table:CIFAR-100-timing-activation-functions}
\end{table}

\begin{table}[H]
    \centering
    \begin{tabular}{lccccc}
    \toprule
    \multirow{2}{*}{Function} & \multicolumn{2}{c}{Single model}              & Ensemble & \multicolumn{2}{c}{Epochs}\\\cline{2-3}\cline{5-6}
                              & Accuracy             & std                    & Accuracy & Range & Mean \\\midrule
    Identity                  & \SI{99.45}{\percent} & $\sigma=0.09$          & \SI{99.63}{\percent} & 55 -- \hphantom{0}77  & 62.2\\
    Logistic                  & \SI{97.27}{\percent} & $\sigma=2.10$          & \SI{99.48}{\percent} & \textbf{37} -- \hphantom{0}76  & \textbf{54.5}\\
    Softmax                   & \SI{99.60}{\percent} & $\boldsymbol{\sigma=0.03}$& \SI{99.63}{\percent} & 44 -- \hphantom{0}73  & 55.6\\
    Tanh                      & \SI{99.40}{\percent} & $\sigma=0.09$          & \SI{99.57}{\percent} & 56 -- \hphantom{0}80  & 67.6\\
    Softsign                  & \SI{99.40}{\percent} & $\sigma=0.08$          & \SI{99.57}{\percent} & 72 -- 101             & 84.0\\
    \gls{ReLU}                & \textbf{\SI{99.62}{\percent}} & \textbf{$\sigma=0.04$} & \textbf{\SI{99.73}{\percent}} & 51 -- \hphantom{0}94 & 71.7\\
    Softplus                  & \SI{99.52}{\percent} & $\sigma=0.05$          & \SI{99.62}{\percent} & 62 -- \hphantom{0}\textbf{70}  & 68.9\\
    \gls{PReLU}               & \SI{99.57}{\percent} & $\sigma=0.07$          & \textbf{\SI{99.73}{\percent}} & 44 -- \hphantom{0}89 & 71.2\\
    \gls{ELU}                 & \SI{99.53}{\percent} & $\sigma=0.06$          & \SI{99.58}{\percent} & 45 -- 111 & 72.5\\
    \bottomrule
    \end{tabular}
    \caption[Activation function evaluation results on MNIST]{Test accuracy of
             adjusted baseline models trained with different activation
             functions on MNIST.}
    \label{table:MNIST-accuracies-activation-functions}
\end{table}
\glsreset{LReLU}

\section{Label smoothing}
Ensembles consisting of $n$~models trained by the same procedure on the
same data but initialized with different weights and trained with a different
order of the training data perform consistently better than single models. One
drawback of ensembles in applications such as self-driving cars is that they
increase the computation by a factor of~$n$. One idea why they improve the
test accuracy is by reducing the variance.

The idea of label smoothing is to use the ensemble prediction of the training
data as labels for another classifier. For every element $x$ of the training
set, the one-hot encoded target $t(x)$ is smoothed by the ensemble prediction
$y_E(x)$
\[t'(x) = \alpha \cdot t(x) + (1-\alpha) y_E(x)\]
where $\alpha \in [0, 1]$ is the smoothing factor.

\pagebreak[3]
There are three reasons why label smoothing could be beneficial:
\begin{itemize}
    \item \textbf{Training speed}: The ensemble prediction contains more
          information about the image than binary class decisions. Classifiers
          in computer vision predict how similar the input looks to other input
          of the classes they are trained on. By smoothing the labels, the
          information that one image could also belong to another class is
          passed to the optimizer. In early stages of the optimization this
          could lead to a lower loss on the non-smoothed validation set.
    \item \textbf{Higher accuracy}: Using smoothed labels for the optimization
          could lead to a higher accuracy of the base-classifier due to a
          smoothed error surface. It might be less likely that the classifier
          gets into bad local minima.
    \item \textbf{Label noise}: Depending on the way how the labels are
           obtained, it might not always be clear which label is the correct
           one. Also, labeling errors can be present in training datasets.
           Those errors severely harm the training. By smoothing the labels
           errors could be relaxed.
\end{itemize}

10~models $m_{\text{smooth}}$ are trained with the $\alpha=0.5$ smoothed labels from the prediction
of an ensemble of 10~baseline models. The mean accuracy of the models trained
on the smoothed training set labels was \SI{63.61}{\percent} (+\SI{0.23}{\percent})
and the standard deviation was $\sigma=0.72$ (+\SI{0.17}{\percent}).
The ensemble of 10~$m_{\text{smooth}}$ models achieved \SI{64.79}{\percent}
accuracy ($+\SI{0.09}{\percent}$). Hence the effect of this kind of label
smoothing on the final accuracy is questionable.

The training speed didn't noticeably change either: The number of trained
epochs ranged from 144 to 205, the mean number of epochs was 177. The baseline
training ranged from 146 to 232 epochs with a mean of 174~epochs. After 10,
30 and 80~epochs both training methods accuracy differed by less than one
percentage point. Hence it is unlikely that label smoothing has a positive
effect on the training speed.

Hinton et al.\ called this method \textit{distillation}
in~\cite{hinton2015distilling}. Hinton et al.\ used smooth and hard labels for
training, this work only used smoothed labels.

\clearpage
\section{Optimized Classifier}
In comparison to the baseline classifier, the following changes are applied to
the optimized classifier:
\begin{itemize}
    \item \textbf{Remove the bias for the last layers}: For all layers which
          output a $1 \times 1$ feature map, the bias is removed
    \item \textbf{Increase the max pooling kernel to $3 \times 3$}
    \item \textbf{More filters in the first layers}
\end{itemize}

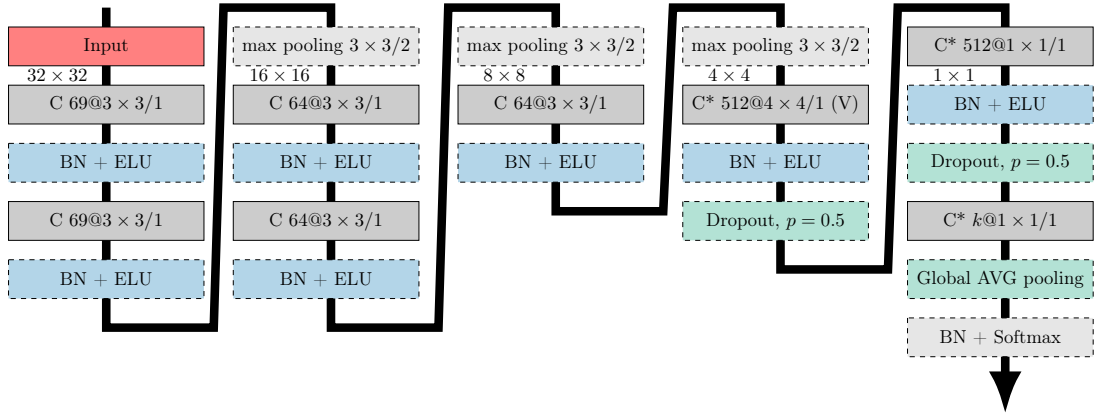
\begin{figure}[H]
    \hspace*{-0.8cm}
    \resizebox {1.05\columnwidth} {!} {
    \newcommand{\width}{0.2}
\newcommand{\height}{0.4}
\newcommand{\disty}{0.2}
\definecolor{colorbblue}{HTML}{0072B2}
\definecolor{colorbgreen}{HTML}{009E73}
\definecolor{colorborange}{HTML}{D55E00}

\tikzstyle{input}=[draw,fill=red!50]
\tikzstyle{conv}=[draw,fill=black!20]
\tikzstyle{max}=[draw,dashed,fill=black!10]
\tikzstyle{dropout}=[draw,dashed,fill=colorbgreen!30]
\tikzstyle{fc}=[draw,fill=green!10]
\tikzstyle{output}=[draw,fill=red!50]
\tikzstyle{act}=[draw,dashed,fill=colorbblue!30]
\def \coldist {2.3}
\def \widthb {1.9}

\begin{tikzpicture}[scale=2]
    \draw[->, -Latex, line width=5pt]
        (1.0+0*\coldist, 0.2) --(1.0+0*\coldist, -4*\disty-5*\height-0.3) --(2.1+0*\coldist, -4*\disty-5*\height-0.3) --(2.2+0*\coldist, 0.2)
     -- (1.0+1*\coldist, 0.2) --(1.0+1*\coldist, -4*\disty-5*\height-0.3) --(2.1+1*\coldist, -4*\disty-5*\height-0.3) --(2.2+1*\coldist, 0.2)
     -- (1.0+2*\coldist, 0.2) --(1.0+2*\coldist, -2*\disty-3*\height-0.3) --(2.1+2*\coldist, -2*\disty-3*\height-0.3) --(2.2+2*\coldist, 0.2)
     -- (1.0+3*\coldist, 0.2) --(1.0+3*\coldist, -3*\disty-4*\height-0.3) --(2.1+3*\coldist, -3*\disty-4*\height-0.3) --(2.2+3*\coldist, 0.2)
     -- (1.0+4*\coldist, 0.2) --(1.0+4*\coldist, -5*\disty-6*\height-0.6);

    \draw[draw=none] (0*\coldist,-1*\height)          rectangle (1.0,-1*\disty-1*\height) node[pos=.5] {$32 \times 32$};
    \draw[input]     (0*\coldist,-0*\height-0*\disty) rectangle (2.0,-0*\disty-1*\height) node[pos=.5] {Input};
    \draw[conv]      (0*\coldist,-1*\height-1*\disty) rectangle (2.0,-1*\disty-2*\height) node[pos=.5] {C $69@3 \times 3 / 1$};
    \draw[act]       (0*\coldist,-2*\height-2*\disty) rectangle (2.0,-2*\disty-3*\height) node[pos=.5] {BN + ELU};
    \draw[conv]      (0*\coldist,-3*\height-3*\disty) rectangle (2.0,-3*\disty-4*\height) node[pos=.5] {C $69@3 \times 3 / 1$};
    \draw[act]       (0*\coldist,-4*\height-4*\disty) rectangle (2.0,-4*\disty-5*\height) node[pos=.5] {BN + ELU};

    \draw[draw=none] (1*\coldist,-1*\height)          rectangle (1*\coldist+\widthb/2,-1*\disty-1*\height) node[pos=.5] {$16 \times 16$};
    \draw[max]       (1*\coldist,-0*\height-0*\disty) rectangle (1*\coldist+\widthb,-0*\disty-1*\height) node[pos=.5] {max pooling $3\times 3 / 2$};
    \draw[conv]      (1*\coldist,-1*\height-1*\disty) rectangle (1*\coldist+\widthb,-1*\disty-2*\height) node[pos=.5] {C $64@3 \times 3 / 1$};
    \draw[act]       (1*\coldist,-2*\height-2*\disty) rectangle (1*\coldist+\widthb,-2*\disty-3*\height) node[pos=.5] {BN + ELU};
    \draw[conv]      (1*\coldist,-3*\height-3*\disty) rectangle (1*\coldist+\widthb,-3*\disty-4*\height) node[pos=.5] {C $64@3 \times 3 / 1$};
    \draw[act]       (1*\coldist,-4*\height-4*\disty) rectangle (1*\coldist+\widthb,-4*\disty-5*\height) node[pos=.5] {BN + ELU};

    \draw[draw=none] (2*\coldist,-1*\height) rectangle (2*\coldist+\widthb/2,-1*\disty-1*\height) node[pos=.5] {$8 \times 8$};
    \draw[max]       (2*\coldist,-0*\height-0*\disty) rectangle (2*\coldist+\widthb,-0*\disty-1*\height) node[pos=.5] {max pooling $3\times 3 / 2$};
    \draw[conv]      (2*\coldist,-1*\height-1*\disty) rectangle (2*\coldist+\widthb,-1*\disty-2*\height) node[pos=.5] {C $64@3 \times 3 / 1$};
    \draw[act]       (2*\coldist,-2*\height-2*\disty) rectangle (2*\coldist+\widthb,-2*\disty-3*\height) node[pos=.5] {BN + ELU};

    \draw[draw=none] (3*\coldist,-1*\height) rectangle (3*\coldist+\widthb/2,-1*\disty-1*\height) node[pos=.5] {$4 \times 4$};
    \draw[max]       (3*\coldist,-0*\height-0*\disty) rectangle (3*\coldist+\widthb,-0*\disty-1*\height) node[pos=.5] {max pooling $3\times 3 / 2$};
    \draw[conv]      (3*\coldist,-1*\height-1*\disty) rectangle (3*\coldist+\widthb,-1*\disty-2*\height) node[pos=.5] {C* $512@4 \times 4 / 1$ (V)};
    \draw[act]       (3*\coldist,-2*\height-2*\disty) rectangle (3*\coldist+\widthb,-2*\disty-3*\height) node[pos=.5] {BN + ELU};
    \draw[dropout]   (3*\coldist,-3*\height-3*\disty) rectangle (3*\coldist+\widthb,-3*\disty-4*\height) node[pos=.5] {Dropout, $p=0.5$};

    \draw[draw=none] (4*\coldist,-1*\height) rectangle (4*\coldist+\widthb/2,-1*\disty-1*\height) node[pos=.5] {$1 \times 1$};
    \draw[conv]      (4*\coldist,-0*\height-0*\disty) rectangle (4*\coldist+\widthb,-0*\disty-1*\height) node[pos=.5] {C* $512@1 \times 1 / 1$};
    \draw[act]       (4*\coldist,-1*\height-1*\disty) rectangle (4*\coldist+\widthb,-1*\disty-2*\height) node[pos=.5] {BN + ELU};
    \draw[dropout]   (4*\coldist,-2*\height-2*\disty) rectangle (4*\coldist+\widthb,-2*\disty-3*\height) node[pos=.5] {Dropout, $p=0.5$};
    \draw[conv]      (4*\coldist,-3*\height-3*\disty) rectangle (4*\coldist+\widthb,-3*\disty-4*\height) node[pos=.5] {C* $k@1 \times 1 / 1$};
    \draw[dropout]   (4*\coldist,-4*\height-4*\disty) rectangle (4*\coldist+\widthb,-4*\disty-5*\height) node[pos=.5] {Global AVG pooling};
    \draw[max]       (4*\coldist,-5*\height-5*\disty) rectangle (4*\coldist+\widthb,-5*\disty-6*\height) node[pos=.5] {BN + Softmax};
\end{tikzpicture}
    }
    \caption[Optimized architecture]{Architecture of the optimized model.
    \texttt{C $32@3 \times 3 / 1$} is a convolutional layer with 32~filters of
    kernel size $3 \times 3$ with stride~1. The * indicates that no bias is
    used.}
    \label{fig:optimized-architecture}
\end{figure}

The detailed architecture is given in \cref{table:optimized-architecture} and
visualized in \cref{fig:optimized-architecture}. The evaluation is given
in~\cref{table:optimized-model-performance} and the timing comparison is given
in~\cref{table:optimized-model-time}.

\begin{table}[ht]
    \renewrobustcmd{\bfseries}{\fontseries{b}\selectfont}
    \sisetup{detect-weight,mode=text,group-minimum-digits = 4}
    \centering
\small
\addtolength{\tabcolsep}{-1.5pt}
\begin{tabular}{
  @{}cl
  S[table-format=3.0]@{\,}c@{\,}l@{\,}c@{\,}l
  S[table-format=5.0]
  S[table-format=8.0]
  S[table-format=4.0]@{\,}c@{\,}S[table-format=2.0]@{}>{${}}c<{{}$}@{}S[table-format=2.0]
  @{}
    }
\toprule
\# & Type
   & \multicolumn{5}{c}{\begin{tabular}[t]{l}Filters @\\ Patch size / stride\end{tabular}}
   & {Parameters} & {FLOPs} & \multicolumn{5}{c}{Output size}\\\midrule
   &    Input        &      & &                         & &       &     0                &        0                &    3 & @ & 32 & \times & 32 \\\cline{1-14}
\multicolumn{1}{|c}{1} & Convolution     & 69   &@& $3 \times 3 \times 3$   &/& 1     &  1932                &  3744768                & \bfseries 69 & @ & \bfseries 32 & \times & \multicolumn{1}{c|}{\bfseries 32} \\
\multicolumn{1}{|c}{2} & BN + ELU        &      & &                         & &       &   138                &   353418                & \bfseries 69 & @ & \bfseries 32 & \times & \multicolumn{1}{c|}{\bfseries 32} \\
\multicolumn{1}{|c}{3} & Convolution     & 69   &@& $3 \times 3 \times32$   &/& 1     & 42918                & 37684096                & \bfseries 69 & @ & \bfseries 32 & \times & \multicolumn{1}{c|}{\bfseries 32} \\
\multicolumn{1}{|c}{4} & BN + ELU        &      & &                         & &       &   138                &   353418                & \bfseries 69 & @ & \bfseries 32 & \times & \multicolumn{1}{c|}{\bfseries 32} \\
\multicolumn{1}{|c}{} & Max pooling      &      & & $2 \times 2$            &/& 2     &     0                &    40960                &   32 & @ & 16 & \times & \multicolumn{1}{c|}{16} \\\cline{1-14}\\[-0.1cm]
 5 & Convolution     & 64   &@& $3 \times 3 \times32$   &/& 1     & 39808                & 20332544                &   64 & @ & 16 & \times & 16 \\
 6 & BN + ELU        &      & &                         & &       &   128                &    82048                &   64 & @ & 16 & \times & 16 \\
 7 & Convolution     & 64   &@& $3 \times 3 \times64$   &/& 1     & 36928                & \bfseries 18857984      &   64 & @ & 16 & \times & 16 \\
 8 & BN + ELU        &      & &                         & &       &   128                &    82048                &   64 & @ & 16 & \times & 16 \\
 ~ & Max pooling     &      & & $2 \times 2$            &/& 2     & ~                    &    20480                &   64 & @ &  8 & \times &  8 \\[0.1cm]
 9 & Convolution     & 64   &@& $3 \times 3 \times64$   &/& 1     & 36928                &  4714496                &   64 & @ &  8 & \times &  8 \\
10 & BN + ELU        &      & &                         & &       &   128                &    20608                &   64 & @ &  8 & \times &  8 \\
 ~ & Max pooling     &      & & $2 \times 2$            &/& 2     & ~                    &     5120                &   64 & @ &  4 & \times &  4 \\[0.1cm]
11 & Convolution (v) & 512  &@& $4 \times 4 \times64$   &/& 1     & \bfseries  524288    &  1048064                &  512 & @ &  1 & \times &  1 \\[0.1cm]
12 & BN + ELU        &      & &                         & &       &  1024                &     3584                &  512 & @ &  1 & \times &  1 \\
 ~ & Dropout 0.5     & ~    & &                         & &       &     0                &        0                &  512 & @ &  1 & \times &  1 \\
13 & Convolution     & 512  &@& $1 \times 1 \times 512$ &/& 1     &262144                &   523776                &  512 & @ &  1 & \times &  1 \\[0.1cm]
14 & BN + ELU        &      & &                         & &       &  1024                &     3584                &  512 & @ &  1 & \times &  1 \\
 ~ & Dropout 0.5     & ~    & &                         & &       &     0                &        0                &  512 & @ &  1 & \times &  1 \\
15 & Convolution     & k    &@& $1 \times 1 \times 512$ &/& 1     & {$512 \cdot k$}      &  {$512 \cdot k$}        &    k & @ &  1 & \times &  1 \\[0.1cm]
   & Global avg Pooling &   & & $1 \times 1$            & &       &     0                &        {$k$}            &    k & @ &  1 & \times &  1 \\[0.1cm]
16 & BN + Softmax    & ~    &~&  ~                      & &       & {$2k$}               &     {$7k$}              &    k & @ &  1 & \times &  1 \\\midrule
   & $\sum$          & ~    & &                         & &       & {\makecell{$514k$\\+\num{947654}}}& {\makecell{$520k$\\+\num{87870996}}} & \multicolumn{5}{r}{\num{179200}+$2k$}\\
\bottomrule
\end{tabular}
    \caption[Optimized architecture]{Optimized architecture with 3~input
             channels of size $32 \times 32$. All convolutional layers use
             \texttt{SAME} padding, except for layer~11 which used
             \texttt{VALID} padding in order to decrease the feature map size
             to $1\times 1$. If the input feature map is bigger than $32 \times
             32$, for each power of two there are two \texttt{Convolution + BN
             + ELU} blocks and one \texttt{Max pooling} block added. This is
             the framed part in the table.}
    \label{table:optimized-architecture}
\end{table}

\begin{table}[ht]
    \centering
    \begin{tabular}{lrlrlrr}
    \toprule
    \multirow{2}{*}{Dataset} & \multicolumn{4}{c}{Single Model Accuracy} &  \multicolumn{2}{c}{Ensemble of 10} \\
              & \multicolumn{2}{c}{Training Set} & \multicolumn{2}{c}{Test Set} & Training Set   & Test Set \\\midrule
    Asirra    & \SI{95.83}{\percent} & $\sigma = 4.70$ & \SI{90.75}{\percent}  & $\sigma = 4.73$ & \SI{98.78}{\percent} & \SI{93.09}{\percent} \\
    CIFAR-10  & \SI{94.58}{\percent} & $\sigma = 0.70$ & \SI{87.92}{\percent}  & $\sigma = 0.46$ & \SI{96.47}{\percent} & \SI{89.86}{\percent} \\
    CIFAR-100 & \SI{77.96}{\percent} & $\sigma = 2.18$ & \SI{64.42}{\percent}  & $\sigma = 0.73$ & \SI{81.44}{\percent} & \SI{67.03}{\percent} \\
    GTSRB     &\SI{100.00}{\percent} & $\sigma = 0.00$ & \SI{99.28}{\percent}  & $\sigma = 0.10$ &\SI{100.00}{\percent} & \SI{99.51}{\percent} \\
    HASYv2    & \SI{88.79}{\percent} & $\sigma = 0.45$ & \SI{85.36}{\percent}  & $\sigma = 0.15$ & \SI{89.36}{\percent} & \SI{85.92}{\percent} \\
    MNIST     & \SI{99.88}{\percent} & $\sigma = 0.10$ & \SI{99.48}{\percent}  & $\sigma = 0.13$ & \SI{99.99}{\percent} & \SI{99.67}{\percent} \\
    STL-10    & \SI{95.43}{\percent} & $\sigma = 3.57$ & \SI{75.09}{\percent}  & $\sigma = 2.39$ & \SI{98.54}{\percent} & \SI{78.66}{\percent} \\
    SVHN      & \SI{99.08}{\percent} & $\sigma = 0.07$ & \SI{96.37}{\percent}  & $\sigma = 0.12$ & \SI{99.50}{\percent} & \SI{97.47}{\percent} \\
    \bottomrule
    \end{tabular}
    \caption[Optimized model evaluation results]{\small Optimized model
             accuracy on eight datasets. The single model actuary is the
             10~models used in the ensemble. The empirical standard
             deviation~$\sigma$ of the accuracy is also given. CIFAR-10,
             CIFAR-100 and STL-10 models use test-time transformations. None of
             the models uses unlabeled data or data from other datasets. For
             MNIST, GTSRB, SVHN and HASY, no test time transformations are
             used.}
    \label{table:optimized-model-performance}
\end{table}

\begin{table}[ht]
    \centering
    {\small
    \begin{tabular}{lllrrr}
    \toprule
    \multirow{2}{*}{Network}
             & \multirow{2}{*}{GPU} & \multirow{2}{*}{Tensorflow}
                                & \multicolumn{2}{c}{Inference per}                               & \multicolumn{1}{c}{Training} \\\cline{4-5}
                    &           &               & 1 Image               & 128 images              & \multicolumn{1}{c}{time / epoch} \\\midrule
    Optimized       & Default   & Intel i7-4930K& \SI{5}{\milli\second} & \SI{432}{\milli\second} & \SI{386}{\second} \\
    Optimized       & Optimized & Intel i7-4930K& \SI{4}{\milli\second} & \SI{307}{\milli\second} & \SI{315}{\second} \\
    Optimized       & Default   & GeForce~940MX & \SI{4}{\milli\second} & \SI{205}{\milli\second} & \SI{192}{\second} \\
    Optimized       & Default   & GTX 970       & \SI{6}{\milli\second} &  \SI{41}{\milli\second} &  \SI{35}{\second} \\
    Optimized       & Default   & GTX 980       & \SI{3}{\milli\second} &  \SI{35}{\milli\second} &  \SI{27}{\second} \\
    Optimized       & Default   & GTX 980 Ti    & \SI{6}{\milli\second} &  \SI{36}{\milli\second} &  \SI{26}{\second} \\
    Optimized       & Default   & GTX 1070      & \SI{2}{\milli\second} &  \SI{24}{\milli\second} &  \SI{21}{\second} \\
    Optimized       & Default   & Titan Black   & \SI{4}{\milli\second} &  \SI{46}{\milli\second} &  \SI{43}{\second} \\
    \bottomrule
    \end{tabular}
    }
    \caption[Optimized model speed comparison]{\small Speed comparison of the
             optimized model on CIFAR-10. The baseline model is evaluated on
             six Nvidia \glspl{GPU} and one CPU\@. The weights for DenseNet-40-12
             are taken from~\cite{Majumdar2017-densenet-weights}. Weights the
             baseline model can be found at~\cite{thoma-msthesis-blog}. The
             optimized Tensorflow build makes use of SSE4.X, AVX, AVX2 and FMA
             instructions.}
    \label{table:optimized-model-time}
\end{table}

\section{Early Stopping vs More Data}
A separate validation set is necessary for two reasons:
\begin{enumerate*}[label=(\arabic*)]
    \item Early stopping and
    \item preventing overfitting due to many experiments.
\end{enumerate*}
To prevent overfitting, a different dataset can be used. For example, all
decisions about hyperparameters in this thesis are based on CIFAR-100, but the
network is finally trained and evaluated with the same hyperparameters on all
datasets.\footnote{Except data augmentation and test time transformations.} The
validation set can hence be removed if early stopping is removed. Instead, the
validation data is used in a first run to determine the number of epochs
necessary for training. In a second training run the validation data is added
to the training set. The number of used epochs for the second run is given
in~\cref{table:optimized-model-training-epochs}.

\begin{table}[ht]
    \centering
    \begin{tabular}{lrrrr}
    \toprule
    Dataset   & Mean epochs    & Train data       & classes   & average data / class \\\midrule
    Asirra    & 60             & \num{15075}      & 2         & \num{7538} \\
    MNIST     & 41             & \num{54000}      & 10        & \num{5400}  \\
    SVHN      & 45             & \num{543949}     & 10        & \num{54395} \\
    CIFAR-10  & 84             & \num{45000}      & 10        & \num{4500}  \\
    HASYv2    & 92             & \num{136116}     & 369       & \num{369}   \\
    GTSRB     & 97             & \num{35288}      & 43        & \num{821}   \\
    STL-10    & 116            & \num{4500}       & 10        & \num{450}   \\
    CIFAR-100 & 155            & \num{45000}      & 100       & \num{450}   \\
    \bottomrule
    \end{tabular}
    \caption[Optimized model mean training epochs]{Mean number of training
             epochs for the optimized model. For comparison, the total amount
             of used training data, the number of classes of the dataset and
             the average amount of data per class is given.}
    \label{table:optimized-model-training-epochs}
\end{table}

Alternatively, the model can be trained with \gls{ES} purely on the training
loss. All three methods -- early stopping on the validation set accuracy, early
stopping on the training loss and training a fixed number of epochs are
evaluated. While having more data helped with Asirra and CIFAR-100, the results
as shown in~\cref{table:optimized-model-comparison-es-data} on the other
datasets are only marginally different. For CIFAR-10, training with more data
did not improve the results when the number of epochs is fixed, but notably
improved the results when the training loss was used as the early stopping
criterion.

\begin{table}[ht]
    \centering
    \begin{tabular}{lrlr}
    \toprule
    \multirow{2}{*}{Dataset} & \multicolumn{2}{c}{Early Stopping} & Fixed epochs\\
              & val. acc             & train loss                 & \\\midrule
    Asirra    & \SI{93.09}{\percent} & \SI{96.01}{\percent}\footnotemark & \SI{96.01}{\percent}\\
    CIFAR-10  & \SI{89.86}{\percent} & \SI{91.75}{\percent} & \SI{88.88}{\percent}\\
    CIFAR-100 & \SI{67.03}{\percent} & \SI{71.01}{\percent} & \SI{69.08}{\percent}\\
    HASYv2    & \SI{85.92}{\percent} & \SI{82.89}{\percent}\footnotemark & \SI{85.05}{\percent}\\
    MNIST     & \SI{99.67}{\percent} & \SI{99.64}{\percent} & \SI{99.57}{\percent}\\
    STL-10    & \SI{78.66}{\percent} & \SI{83.25}{\percent} & \SI{78.64}{\percent}\\
    \bottomrule
    \end{tabular}
    \caption[Optimized model trained with early stopping vs training with more
    data]{Comparisons of trained optimized models with early stopping on the
    validation accuracy compared training setups without a validation set and
    thus more training data. The second column uses the training loss as a
    stopping criterion, the third column uses a fixed number of epochs which
    is equal to the mean number of training epochs of the models with early
    stopping on the validation set accuracy.}
    \label{table:optimized-model-comparison-es-data}
\end{table}
\addtocounter{footnote}{-1}
\footnotetext{Only 1~model is trained due to the long training time of 581~epochs and 12~hours for this model.}
\addtocounter{footnote}{1}
\footnotetext{Only 3~models are in this ensemble due to the long training time of more than 8~hours per model.}

\section{Regularization}
Stronger regularization might even improve the results when using the training
loss as an early stopping criterion. $\ell_2$ regularization with a weighting
factor of $\lambda = 0.0001$ is used in all other experiments. While the
accuracy as shown in~\cref{table:opt-model-regularization} does not show a
clear pattern, the number of epochs increases with lower model regularization
(see~\cref{table:opt-model-regularization-epochs}).

\begin{table}[ht]
    \centering
    \begin{tabular}{lrlrlrr}
    \toprule
    \multirow{2}{*}{$\lambda $} & \multicolumn{4}{c}{Single Model Accuracy} &  \multicolumn{2}{c}{Ensemble of 10} \\
              & \multicolumn{2}{c}{Training Set}     & \multicolumn{2}{c}{Test Set}         & Training Set          & Test Set \\\midrule
    $\lambda = 0.01$   & \SI{73.83}{\percent} & $\sigma=1.78$ & \SI{58.94}{\percent} & $\sigma=1.33$ &  \SI{87.78}{\percent} & \SI{69.98}{\percent} \\
    $\lambda = 0.001$  & \SI{82.86}{\percent} & $\sigma=0.89$ & \SI{63.03}{\percent} & $\sigma=0.67$ &  \SI{91.86}{\percent} & \SI{71.02}{\percent} \\
    $\lambda = 0.0001$ & \SI{77.96}{\percent} & $\sigma=2.18$ & \SI{64.42}{\percent} & $\sigma=0.73$ &  \SI{81.44}{\percent} & \SI{67.03}{\percent} \\
    \bottomrule
    \end{tabular}
    \caption[Model regularization with early stopping on training
    loss]{Different choices of $\ell_2$ model regularization applied to the
    optimized model.}
    \label{table:opt-model-regularization}
\end{table}

\begin{table}[ht]
    \centering
    \begin{tabular}{lllll}
    \toprule
    $\lambda $         & min & max & mean  & std  \\\midrule
    $\lambda = 0.01$   & 457 & 503 & 404.6 & 37.2 \\
    $\lambda = 0.001$  & 516 & 649 & 588.4 & 41.6 \\
    $\lambda = 0.0001$ & 579 & 833 & 696.1 & 79.1 \\
    \bottomrule
    \end{tabular}
    \caption[Model regularization with early stopping on training
    loss - Training time]{Training time in epochs of models with early stopping on
                          training loss by different choices of $\ell_2$ model
                          regularization applied to the optimized model.}
    \label{table:opt-model-regularization-epochs}
\end{table}

\chapter{Conclusion and Outlook}
This master thesis gave an extensive overview over the design patterns of
\glspl{CNN} in~\cref{ch:CNN}, the methods how \glspl{CNN} can be analyzed and
the principle directions of topology learning algorithms
in~\cref{ch:topology-learning}.

\Glsfirst{CMO}, originally developed as a method to make visualizations of
confusion matrices easier to read (see~\cref{fig:HASY-cm-train-unsorted}), was
introduced as a class clustering algorithm
in~\cref{ch:hierarchical-classification} and evaluated
in~\cref{sec:clustering-classes,sec:hierarchy-of-classifiers}. The important
insights are:
\begin{itemize}
    \item Ordering the classes in the confusion matrix allows to display the
          relevant parts even for several hundred classes.
    \item A hierarchy of classifiers based on the classes does not improve
          the results on CIFAR-100. There are three possible reasons for this:
          \begin{itemize}
              \item $\SI{32}{\pixel} \times \SI{32}{\pixel}$ is too low
                    dimensional
              \item 100~classes are not enough for this approach
              \item More classes are always easier to distinguish if each new
                    class comes with more data. One reason why this might be
                    the case is that distinguishing the object from background
                    has similar properties even for different classes.
          \end{itemize}
    \item Label smoothing had only a minor effect on the accuracy and no effect
          on the training time when a single base classifier was used to train
          with the smoothed labels by an ensemble of base classifiers.
\end{itemize}

A baseline model was defined and evaluated on eight publicly available
datasets. The baselines topology and training setup are described in detail as
well as its behavior during training and properties of the weights of the
trained model.

The influence of various hyperparameters is examined in~\cref{sec:05-network-width,sec:weight-updates,sec:multiple-narrow-one-wide,sec:05-batch-normalizaiton,sec:05-batch-size,sec:05-bias,sec:05-color-space-transformations,sec:05-pooling}
for CIFAR-100. The insights of those experiments are:
\begin{itemize}
    \item Averaging ensembles of 10~base classifiers of the same architecture
          and trained with the same setup consistently improve the accuracy.
          The amount of improvement depends on the base classifiers, but the
          ensemble tends to improve the test accuracy by about one percentage
          point.
    \item Wider networks learn in fewer epochs. This, however, does not mean
          that the wall-clock time is lower due to increased computation in
          forward- and backward passes.
    \item Batch Normalization increases the training time noticeably. For
          the described \gls{ELU} baseline model it also increases accuracy,
          which contradicts~\cite{clevert2015fast}.
    \item The lower the batch size, the longer the time for each epoch of
          training and the less epochs need to be trained. Higher accuracy by
          lower batch sizes was empirically confirmed. The batch size, however,
          can also be too low.
    \item An analysis of the weights of the baseline indicated that the bias
          of layers close to the output layer can be removed. This was
          experimentally confirmed.
    \item It could not be confirmed that learned color space transformation, as
          described in~\cite{mishkin2016systematic}, improves the network.
          Neither with \gls{ELU} nor with \gls{LReLU} and $\alpha=0.3$.
    \item It could be confirmed that \gls{ELU} networks gives better results
          than any other activation function on CIFAR-100. For the character
          datasets MNIST and HASYv2, however, \gls{ReLU}, \gls{LReLU},
          \gls{PReLU}, Softplus and \gls{ELU} all performed similar.
    \item Changing the activation functions to the identity had very little
          impact on the HASYv2 and MNIST classifiers. Note that those networks
          are still able to learn nonlinear decision boundaries due to
          max-pooling and \texttt{SAME} padding. For CIFAR-100, however, the
          accuracy drops by \SI{6.64}{\percent} when \gls{ELU} is replaced by
          the identity.
\end{itemize}

Based on the results of those experiments, an optimized classifier was
developed and evaluated on all eight datasets.

The state of the art of STL-10 was improved from
\SI{74.80}{\percent}~\cite{zhao2015stacked} to \SI{78.66}{\percent} without using
the unlabeled part of the dataset. The state of the art of HASYv2 was improved
from \SI{81.00}{\percent}~\cite{thoma2017hasyv2} to \SI{85.92}{\percent}, for
GTSRB the state of the art was improved from
\SI{99.46}{\percent}~\cite{6033589} to \SI{99.51}{\percent}, for Asirra it was
improved from \SI{82.7}{\percent}~\cite{golle2008machine} to
\SI{93.09}{\percent}.\footnote{The baseline is better than the optimized model on Asirra and on HASYv2.} This was mainly achieved by the combination of \Gls{ELU},
Dropout, ensembles, training data augmentation and test-time transformations.
The removal of the bias of layers close to the output and re-usage of those
parameters in layers close to the input as well as using $3\times 3$ pooling
instead of $2 \times 2$ pooling improved the baseline.

While writing this masters thesis, several related questions could not be
answered:
\begin{itemize}
    \item Deeper \glspl{CNN} have generally higher accuracy, if trained long
          enough and if overfitting is not a problem. But at which
          subsampling-level does having more layers have the biggest effect?
          Can this question be answered before a deeper network is trained?
    \item Is label smoothing helpful for noisy labels?
    \item How does the choice of activation functions influence residual
          architectures? Could the results be the same for different activation
          functions in architectures with hundreds of layers?
    \item The results for the pooling kernel were inconclusive. Larger pooling
          kernels might be advantageous as well as fractional max
          pooling~\cite{graham2014fractional}.
    \item Why is the mean weight update (see~\cref{fig:baseline-weight-updates-mean})
          not decreasing? Is this an effect that can and should be fixed?
    \item Why is softmax so much better than the logistic function? Can the
          reason be used to further improve \gls{ELU}?
\end{itemize}

Besides those questions, the influence of optimizers on time per epoch, epochs
until convergence, total training time, memory consumption, accuracy of the
models and standard deviation of the models was not evaluated. This, and the
stopping criterion for training might be crucial for the models quality.

\pagestyle{scrplain}
\appendix

\chapter{Figures, Tables and Algorithms}
\begin{figure}[ht]
    \centering
    \subfloat[Original image]{
        \includegraphics[width=0.32\textwidth]{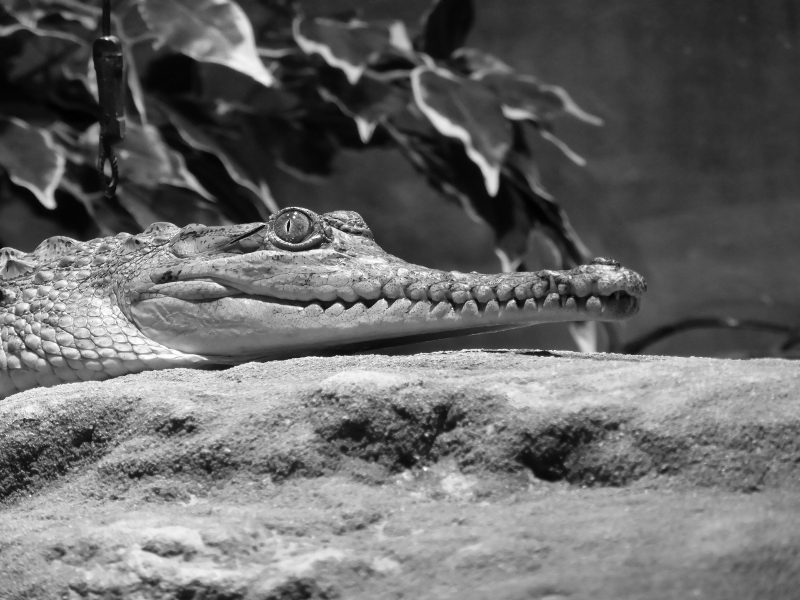}
        \label{fig:image-filters-original}
    }%
    \subfloat[Smoothing filter]{
        \includegraphics[width=0.32\textwidth]{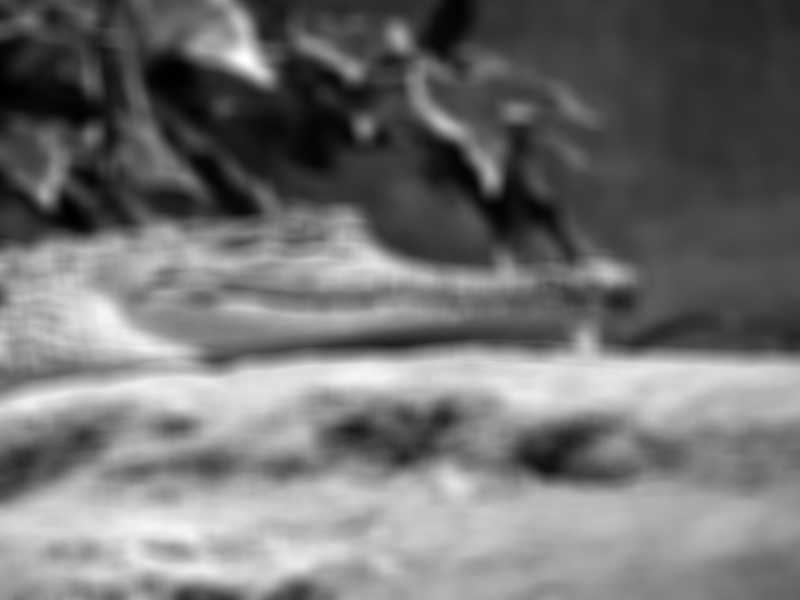}
        \label{fig:image-filters-uniform-smoothing}
    }%
    \subfloat[Laplace edge detection filter]{
        \includegraphics[width=0.32\textwidth]{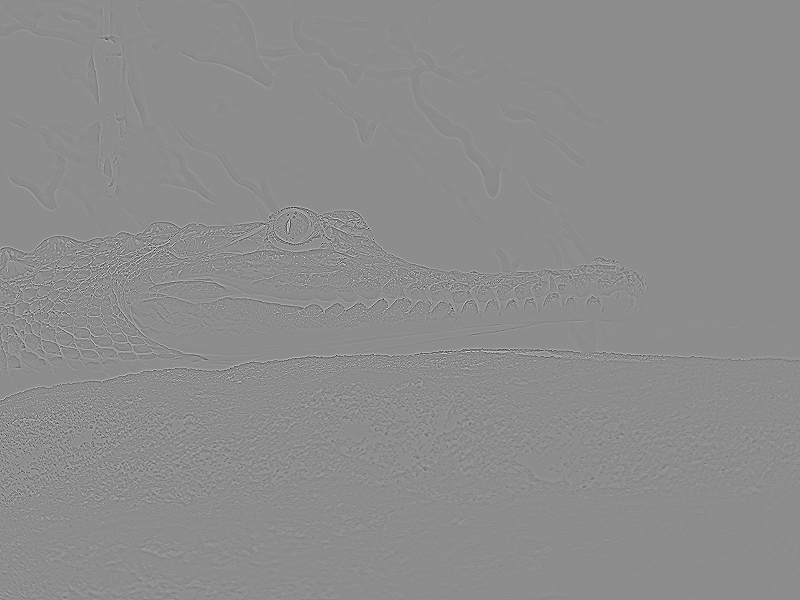}
        \label{fig:image-filters-laplace}
    }\\
    \subfloat[Sobel edge detection filter]{
        \includegraphics[width=0.32\textwidth]{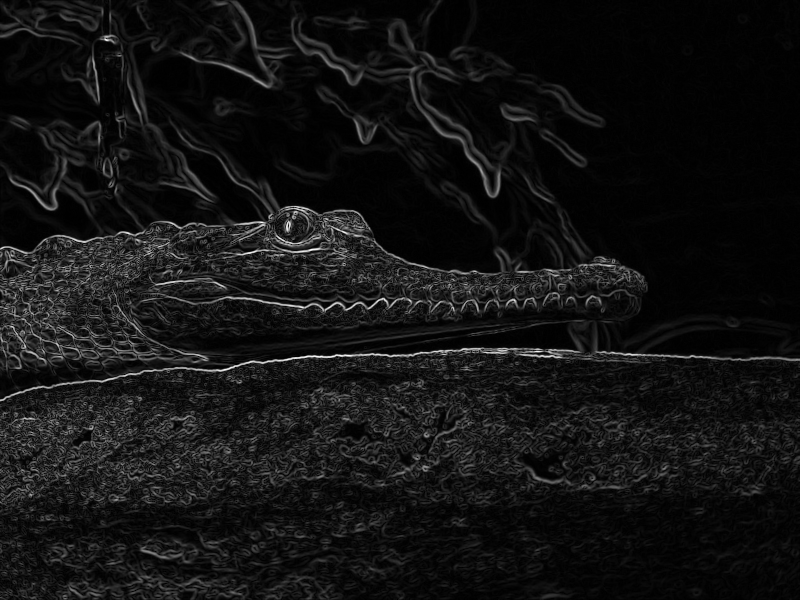}
        \label{fig:image-filters-sobel}
    }%
    \subfloat[Prewitt edge detection filter]{
        \includegraphics[width=0.32\textwidth]{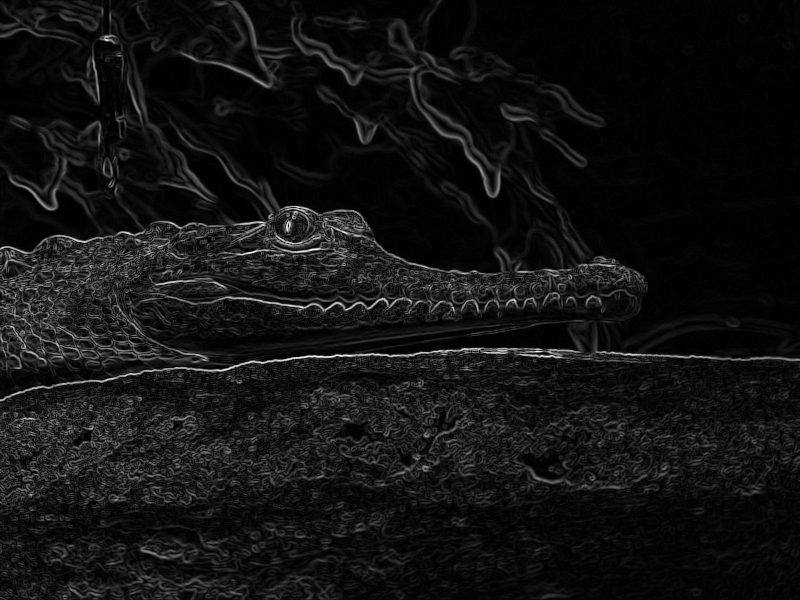}
        \label{fig:image-filters-prewitt}
    }%
    \subfloat[Canny filter]{
        \includegraphics[width=0.32\textwidth]{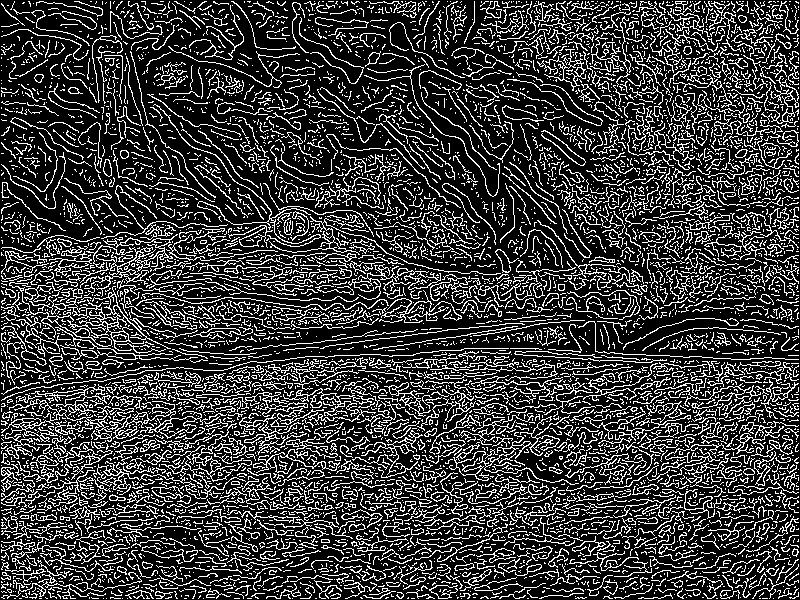}
        \label{fig:image-filters-canny}
    }%
    \caption[Image Filters]{Examples of image filters. Best viewed in electronic form.}
    \label{fig:image-filters}
\end{figure}

\begin{figure}[ht]
    \centering
    \includegraphics[width=0.98\linewidth,keepaspectratio]{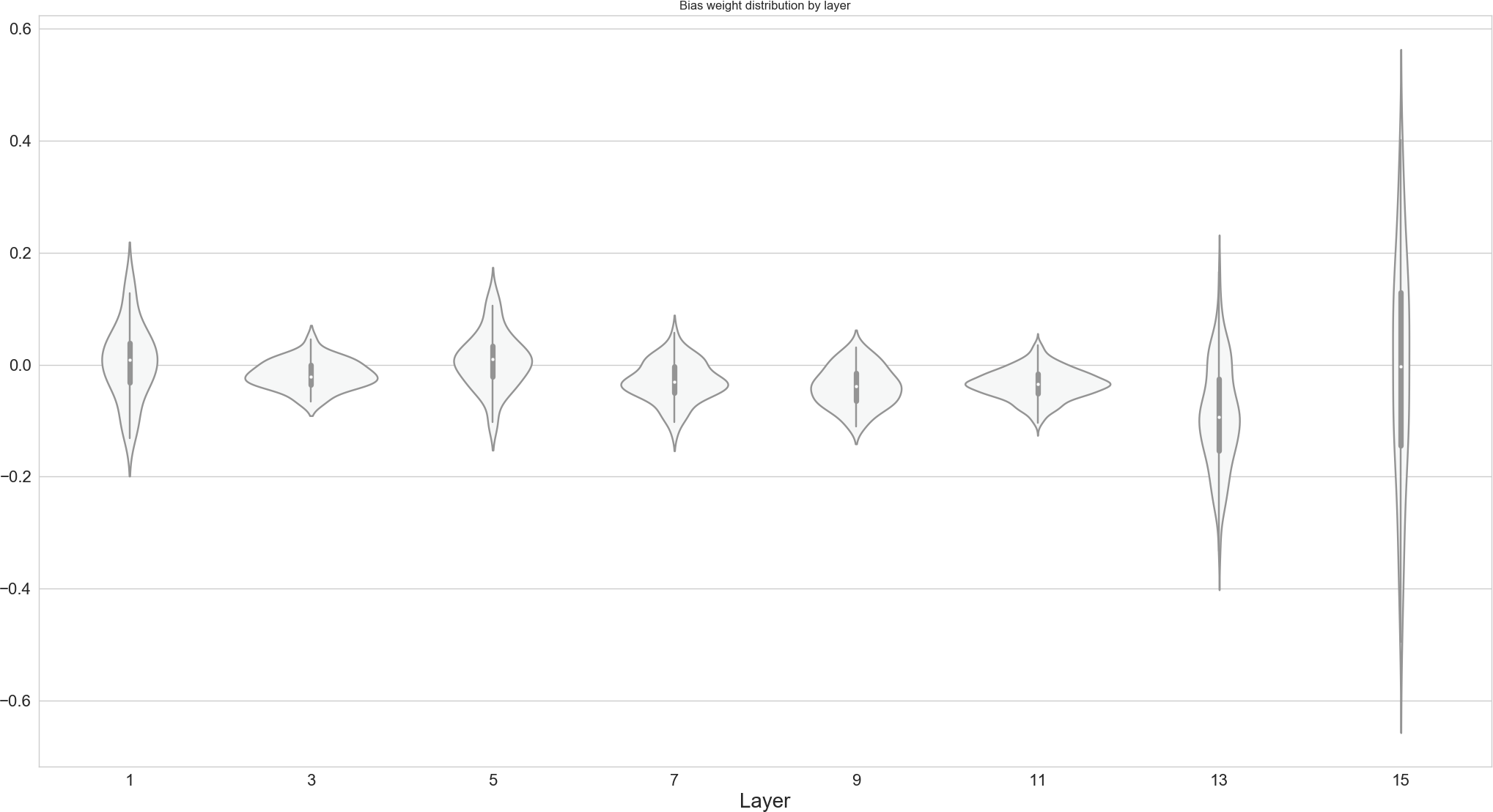}
    \caption[Bias weight distribution without BN]{The distribution of bias
             weights of a model without batch normalization trained on
             CIFAR-100.}
    \label{fig:no-bn-bias-weight-dist}
\end{figure}

\begin{figure}[ht]
    \centering
    \includegraphics[width=0.98\linewidth,keepaspectratio]{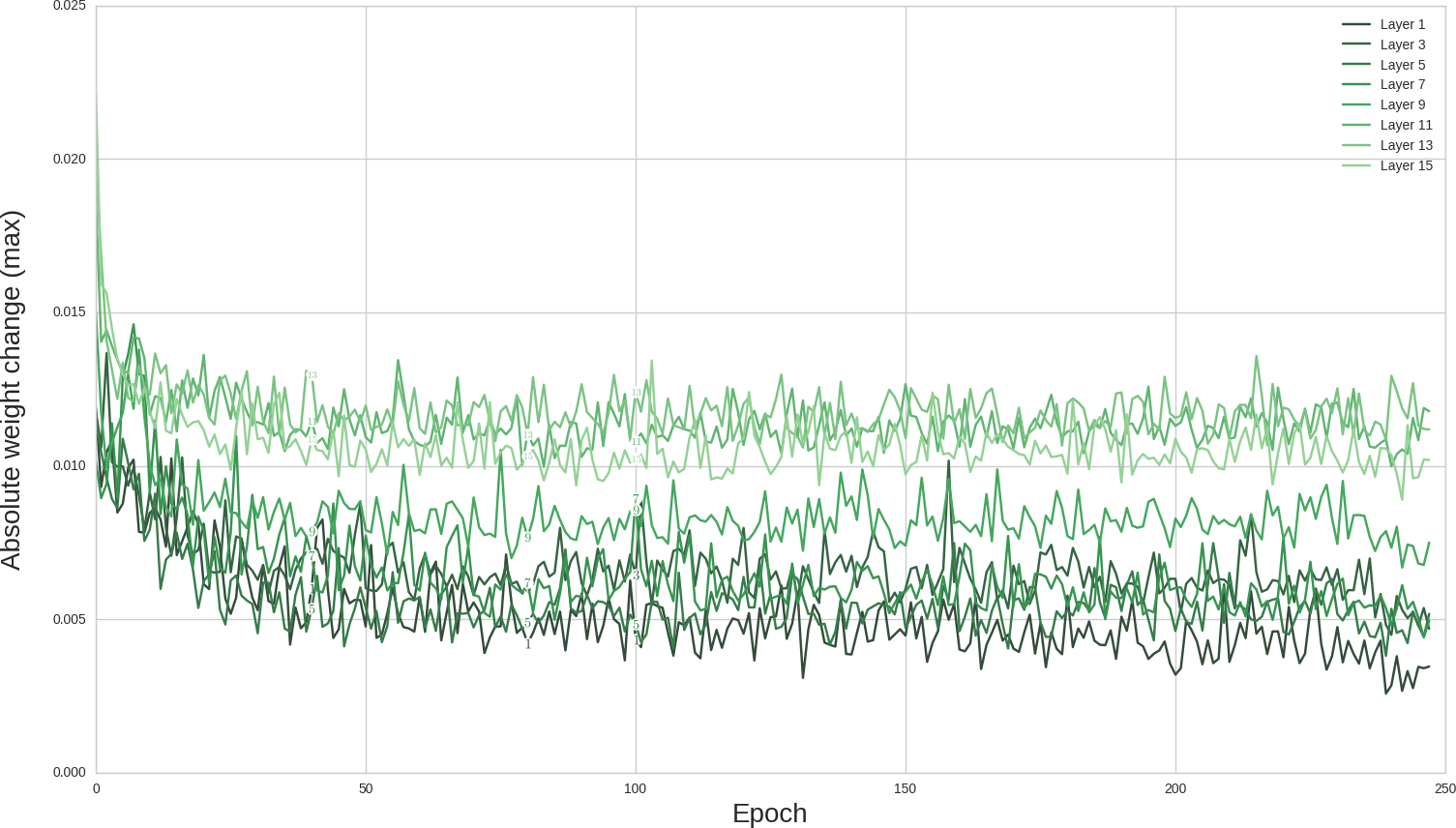}
    \caption[Maximum weight updates of baseline with bottleneck]{Maximum weight
             updates between epochs by layer. The model is the baseline model,
             but with layer~5 reduced to 3~filters.}
    \label{fig:erratic-weight-updates-max}
\end{figure}

\begin{figure}[ht]
    \centering
    \includegraphics[width=0.98\linewidth,keepaspectratio]{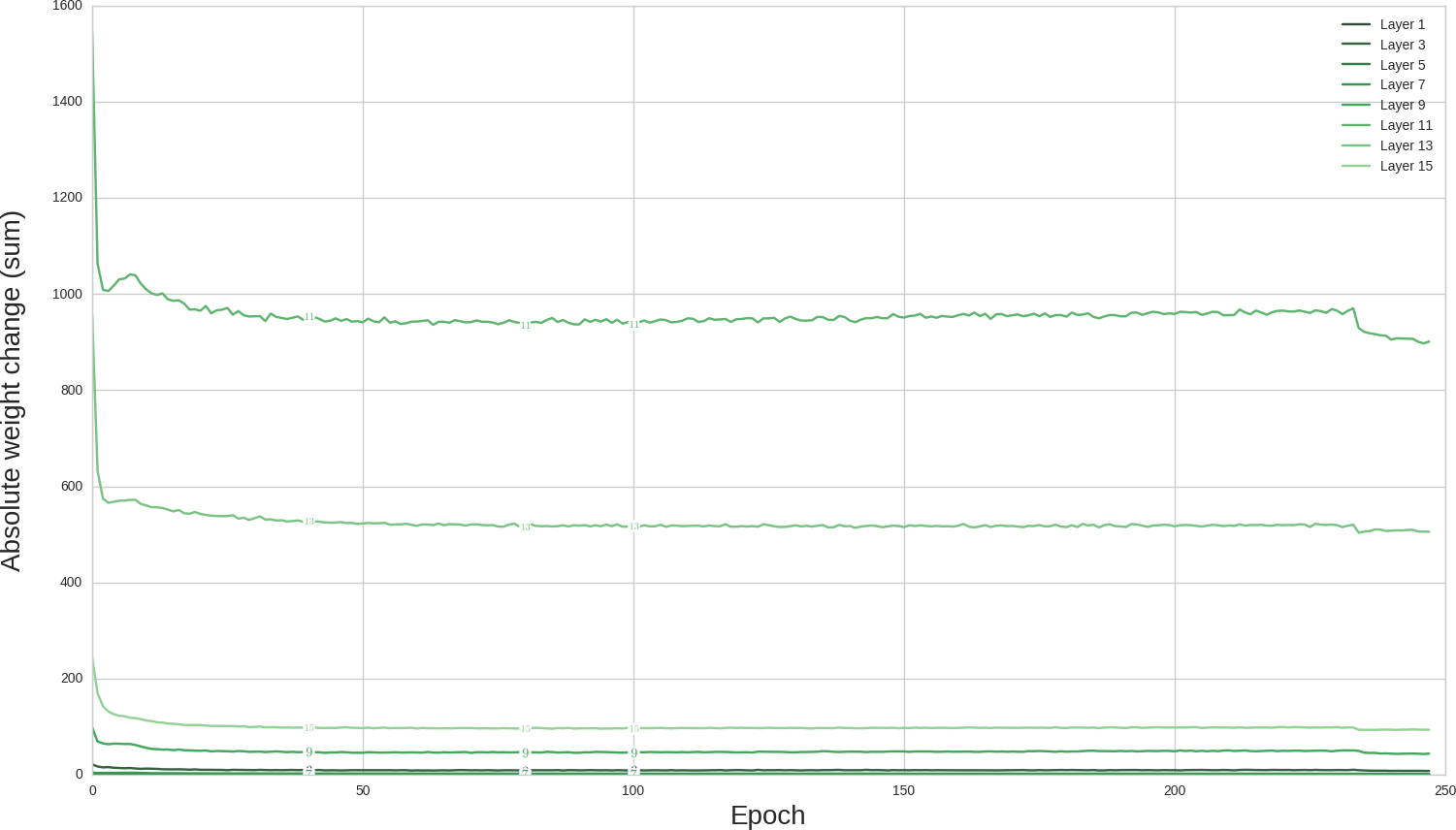}
    \caption[Sum of weight updates of baseline with bottleneck]{Sum of weight
             updates between epochs by layer.  The model is the baseline model,
             but with layer~5 reduced to 3~filters.}
    \label{fig:erratic-weight-updates-sum}
\end{figure}

\begin{table}[ht]
    \centering
    {\small
    \begin{tabular}{rcc}
    \toprule
    \multirow{2}{*}{Layer} & \multicolumn{2}{c}{99-percentile interval} \\
          & filter                  & bias          \\\midrule
    1     & [-0.50, 0.48]           & [-0.06, 0.07] \\
    3     & [-0.21, 0.19]           & [-0.07, 0.07] \\
    5     & [-0.20, 0.17]           & [-0.07, 0.05] \\
    7     & [-0.15, 0.14]           & [-0.05, 0.06] \\
    9     & [-0.14, 0.15]           & [-0.04, 0.03] \\
    11    & [-0.08, 0.08]           & [-0.00, 0.00] \\
    13    & [-0.08, 0.08]           & [-0.00, 0.00] \\
    15    & [-0.10, 0.11]           & [-0.01, 0.01] \\
    \bottomrule
    \end{tabular}
    }
    \caption[99-percentile intervals for filter weights on
    CIFAR-100]{99-percentile intervals for filter weights and bias weights by
    layer of a baseline model trained on CIFAR-100.}
    \label{table:baseline-percentile-interval}
\end{table}

\begin{algorithm}[H]
    \begin{algorithmic}
    \Require $C \in \mathbb{N}^{n \times n}$, steps $\in \mathbb{N}$, $T \in \mathbb{R}^+$, $c \in (0, 1)$
    \Procedure{SimulatedAnnealing}{$C$, steps, $T$, $c$}
        \State $\text{bestScore} \gets \Call{accuracy}{C}$
        \State $\text{bestC} \gets C$
        \For{$i=0$; $i < \text{steps}$; $i \gets i + 1$}
            \State $p \gets \Call{randomFloat}{0, 1}$
            \If{$p < 0.5$}
                \Comment{Swap rows}
                \State $i \gets \Call{randomInteger}{1, \dots, n}$
                \State $j \gets \Call{randomInteger}{1, \dots, n} \setminus \Set{i}$
                \State $p \gets \Call{randomUniform}{0, 1}$
                \State $C' \gets \Call{swap}{C, i, j}$
                \State $s \gets \Call{accuracy}{C'}$
                \If{$p < \operatorname{exp}(\frac{s-\text{bestScore}}{T})$}
                    \State $C \gets C'$
                    \If{$s > \text{bestScore}$}
                        \State $\text{bestScore} \gets s$
                        \State $\text{bestC} \gets C$
                    \EndIf
                \EndIf
                \State $T \gets T \cdot c$
            \Else
                \Comment{Move Block}
                \State $s \gets \Call{randomInteger}{1, \dots, n}$ \Comment{Block start}
                \State $e \gets \Call{randomInteger}{s, \dots, n}$ \Comment{Block end}
                \State $i \gets \Call{randomInteger}{1, \dots, n - (e - s)}$ \Comment{Block insert position}
                \State Move Block (s, \dots, e) to position $i$
            \EndIf
        \EndFor
        \State \Return bestM
    \EndProcedure
    \end{algorithmic}
    \caption[Simulated Annealing]{Simulated Annealing for minimizing~\cref{eq:dist}.}
    \label{alg:simulated-annealing}
\end{algorithm}

\begin{table}[H]
    \centering
    \setlength\tabcolsep{1.5pt}
    \begin{tabular}{@{\extracolsep{4pt}}lcccccccr@{}}
    \toprule
    \multirow{2}{*}{Function} & \multicolumn{4}{c}{Single model}              & \multicolumn{2}{c}{Ensemble of 10} & \multicolumn{2}{c}{Epochs}\\\cline{2-5}\cline{6-7}\cline{8-9}
                              & \multicolumn{2}{c}{Training set}     &\multicolumn{2}{c}{Test set}                  & Train                & Test                 & Range     & \multicolumn{1}{c}{Mean} \\\midrule
    Identity                  & \SI{87.92}{\percent} & $\sigma=0.40$ & \SI{84.69}{\percent} & $\sigma=0.08$         & \SI{88.59}{\percent} & \SI{85.43}{\percent} & \hphantom{0}92 -- 140 & 114.5\\
    Logistic                  & \SI{81.46}{\percent} & $\sigma=5.08$ & \SI{79.67}{\percent} & $\sigma=4.85$         & \SI{86.38}{\percent} & \SI{84.60}{\percent} & \hphantom{0}\textbf{58} -- \hphantom{0}\textbf{91}  & \textbf{77.3}\\
    Softmax                   & \SI{88.19}{\percent} & $\sigma=0.31$ & \SI{84.70}{\percent} & $\sigma=0.15$         & \SI{88.69}{\percent} & \SI{85.43}{\percent} & 124 -- 171& 145.8\\
    Tanh                      & \SI{88.41}{\percent} & $\sigma=0.36$ & \SI{84.46}{\percent} & $\sigma=0.27$         & \SI{89.24}{\percent} & \SI{85.45}{\percent} & \hphantom{0}89 -- 123 & 108.7\\
    Softsign                  & \SI{88.00}{\percent} & $\sigma=0.47$ & \SI{84.46}{\percent} & $\sigma=0.23$         & \SI{88.77}{\percent} & \SI{85.33}{\percent} & \hphantom{0}77 -- 119 & 104.1\\
    \gls{ReLU}                & \SI{88.93}{\percent} & $\sigma=0.46$ & \textbf{\SI{85.35}{\percent}} & $\sigma=0.21$         & \SI{89.35}{\percent} & \SI{85.95}{\percent} & \hphantom{0}96 -- 132 & 102.8\\
    Softplus                  & \SI{88.42}{\percent} & $\boldsymbol{\sigma=0.29}$ & \SI{85.16}{\percent} & $\sigma=0.15$         & \SI{88.90}{\percent} & \SI{85.73}{\percent} &            108 -- 143 & 121.0\\
    \gls{LReLU}               & \SI{88.61}{\percent} & $\sigma=0.41$ & \SI{85.21}{\percent} & $\boldsymbol{\sigma=0.05}$         & \SI{89.07}{\percent} & \SI{85.83}{\percent} & \hphantom{0}87 -- 117 & 104.5\\
    \gls{PReLU}               & \textbf{\SI{89.62}{\percent}} & $\sigma=0.41$ & \textbf{\SI{85.35}{\percent}} & $\sigma=0.17$& \textbf{\SI{90.10}{\percent}} & \SI{86.01}{\percent} & \hphantom{0}85 -- 111 & 100.5\\
    \gls{ELU}                 & \SI{89.49}{\percent} & $\sigma=0.42$ & \textbf{\SI{85.35}{\percent}} & $\sigma=0.10$         & \SI{89.94}{\percent} & \textbf{\SI{86.03}{\percent}} & \hphantom{0}73 -- 113 &  92.4\\
    \bottomrule
    \end{tabular}
    \caption[Activation function evaluation results on HASYv2]{Test accuracy of
             adjusted baseline models trained with different activation
             functions on HASYv2. For LReLU, $\alpha = 0.3$ was chosen.}
    \label{table:HASYv2-accuracies-activation-functions}
\end{table}

\begin{table}[H]
    \centering
    \setlength\tabcolsep{1.5pt}
    \begin{tabular}{@{\extracolsep{4pt}}lcccccccr@{}}
    \toprule
    \multirow{2}{*}{Function} & \multicolumn{4}{c}{Single model}              & \multicolumn{2}{c}{Ensemble of 10} & \multicolumn{2}{c}{Epochs}\\\cline{2-5}\cline{6-7}\cline{8-9}
                              & \multicolumn{2}{c}{Training set}     &\multicolumn{2}{c}{Test set}                  & Train                & Test                 & Range     & \multicolumn{1}{c}{Mean} \\\midrule
    Identity                  & \SI{87.49}{\percent} & $\sigma=2.50$ & \SI{69.86}{\percent} & $\sigma=1.41$         & \SI{89.78}{\percent} & \SI{71.90}{\percent} & \hphantom{0}51 -- \hphantom{0}65  &  53.4\\
    Logistic                  & \SI{45.32}{\percent} & $\sigma=14.88$& \SI{40.85}{\percent} & $\sigma=12.56$        & \SI{51.06}{\percent} & \SI{45.49}{\percent} & \hphantom{0}38 -- \hphantom{0}93  &  74.6\\
    Softmax                   & \SI{87.90}{\percent} & $\sigma=3.58$ & \SI{67.91}{\percent} & $\sigma=2.32$         & \SI{91.51}{\percent} & \SI{70.96}{\percent} & 108 -- 150           & 127.5\\
    Tanh                      & \SI{85.38}{\percent} & $\sigma=4.04$ & \SI{67.65}{\percent} & $\sigma=2.01$         & \SI{90.47}{\percent} & \SI{71.29}{\percent} & 48 -- \hphantom{0}92 & 65.2\\
    Softsign                  & \SI{88.57}{\percent} & $\sigma=4.00$ & \SI{69.32}{\percent} & $\sigma=1.68$         & \SI{93.04}{\percent} & \SI{72.40}{\percent} & 55 -- 117            & 83.2\\
    \gls{ReLU}                & \SI{94.35}{\percent} & $\sigma=3.38$ & \SI{71.01}{\percent} & $\sigma=1.63$         & \SI{98.20}{\percent} & \SI{74.85}{\percent} & 52 -- \hphantom{0}98 & 75.5\\
    Softplus                  & \SI{83.03}{\percent} & $\sigma=2.07$ & \SI{68.28}{\percent} & $\sigma=1.74$         & \SI{93.04}{\percent} & \SI{75.99}{\percent} & 56 -- \hphantom{0}89 & 68.9\\
    \gls{LReLU}               & \SI{93.83}{\percent} & $\sigma=3.89$ & \SI{74.66}{\percent} & $\sigma=2.11$         & \SI{97.56}{\percent} & \SI{78.08}{\percent} & 52 -- 120 & 80.1\\
    \gls{PReLU}               & \SI{95.53}{\percent} & $\sigma=1.92$ & \SI{71.69}{\percent} & $\sigma=1.37$         & \SI{98.17}{\percent} & \SI{74.69}{\percent} & 59 -- 101 & 78.8\\
    \gls{ELU}                 & \SI{95.42}{\percent} & $\sigma=3.57$ & \SI{75.09}{\percent} & $\sigma=2.39$         & \SI{98.54}{\percent} & \SI{78.66}{\percent} & 66 -- \hphantom{0}72 & 67.2\\
    \bottomrule
    \end{tabular}
    \caption[Activation function evaluation results on STL-10]{Test accuracy of
             adjusted baseline models trained with different activation
             functions on STL-10. For LReLU, $\alpha = 0.3$ was chosen.}
    \label{table:STL-10-accuracies-activation-functions}
\end{table}

\chapter{Hyperparameters}\label{ch:hyperparameters}
Hyperparameters are parameters of models which are not optimized automatically
(e.g., by gradient descent), but by methods like random
search~\cite{bergstra2012random}, grid search~\cite{EfficientBackprop} or
manual search.

\section{Preprocessing}\label{sec:preprocessing}
Preprocessing used to be of major importance in machine learning. However, with
the availability of data sets with hundreds of examples per class and the
possibility of \glspl{CNN} to learn features themselves, most models today rely
on raw pixel values. The only common preprocessing is size normalization. In
order to get a fixed input-size for a \gls{CNN}, the following procedure can be
used:

\begin{itemize}
    \item Take one or multiple crops of the image which have the desired
          aspect ratio.
    \item Scale the crop(s) to the desired size.
    \item In training, all crops can be used independently. In testing, all
          crops can be passed through the network and the output probability
          distributions can get fusioned, for example by averaging.
\end{itemize}

Other preprocessing methods are:
\begin{itemize}
    \item Color space transformations (RGB, \glsdisp{HSV}{HSV}, etc.)
    \item Mean subtraction
    \item Standardization of pixel-values to $[0, 1]$ by dividing through $255$ (used by~\cite{huang2016densely})
    \item Dimensionality reduction
    \begin{itemize}
        \item \Gls{PCA}: An unsupervised linear transformation which can be
              learned in the first hidden layer. It is hence doubtful if PCA
              improves the network.
        \item \Gls{LDA}
    \end{itemize}
    \item \Gls{ZCA} whitening (used by~\cite{krizhevsky2009learning})
\end{itemize}

\section{Data augmentation}\label{sec:data-augmentation}
Data augmentation techniques aim at making artificially more data from real data
items by applying invariances. For computer vision, they include:

\glsunset{GAN}
\begin{table}[H]
    \centering
    \begin{tabular}{lll}
    \toprule
    Name            & Augmentation Factor                      & Used by \\\midrule
    Horizontal flip & $\hphantom{\sim 0}2$                     & \cite{AlexNet-2012,wu2015deep} \\
    Vertical flip   & $\hphantom{\sim 0}2$                     & \cite{dieleman2015rotation}\footnotemark\\
    Rotation        & $\sim 40$ ($\delta = 20$)                & \cite{NIPS2014_5548} \\
    Scaling         & $\sim 14$ ($\delta \in [0.7, 1.4]$)      & \cite{NIPS2014_5548} \\
    Crops           & $\hphantom{\sim \,\,}32^2 = 1024$        & \cite{AlexNet-2012,wu2015deep} \\
    Shearing        &                                          & \cite{graham2014fractional}\\
    \glspl{GAN}     &                                          & \cite{Bao2017}\\\midrule
    Brightness      & $\sim 20$ ($\delta \in [0.5, 1.5]$)      & \cite{howard2013some} \\
    Hue             & $\hphantom{\sim \,\,}51$ ($\delta = 0.1$)& \cite{7301739,NIPS2014_5548} \\
    Saturation      & $\sim 20$ ($\delta = 0.5$)               & \cite{NIPS2014_5548} \\
    Contrast        & $\sim 20$ ($\delta \in [0.5, 1.5]$)      & \cite{howard2013some} \\
    Channel shift   &                                          & \cite{AlexNet-2012} \\
    \bottomrule
    \end{tabular}
    \caption[Data augmentation techniques]{Overview of data augmentation
             techniques. The augmentation factor is calculated for typical
             situations. For example, the augmentation factor for random crops
             is calculated for $\SI{256}{\pixel} \times \SI{256}{\pixel}$
             images which are cropped to $\SI{224}{\pixel} \times
             \SI{224}{\pixel}$.}
    \label{table:data-augmentation-overview}
\end{table}
\footnotetext{Vertical flipping combined with $180^\circ$ rotation is equivalent to horizontal flipping}
\glsreset{GAN}

Taking several scales if the original is of higher resolution than desired is
another technique. Combinations of the techniques above can also be applied.
Please note that the order of operations does matter in many cases and hence
the order is another augmentation factor.

Less common, but also reasonable are:
\begin{itemize}
    \item Adding noise
    \item Elastic deformations
    \item Color casting (used by~\cite{wu2015deep})
    \item Vignetting (used by~\cite{wu2015deep})
    \item Lens distortion (used by~\cite{wu2015deep})
\end{itemize}

\section{Initialization}\label{sec:initialization}
Weight initializations are usually chosen to be small and centered around zero.
One way to characterize many initialization schemes is by
\[w \sim \alpha \cdot \mathcal{U}[-1, 1] + \beta \cdot \mathcal{N}(0, 1) + \gamma \text{ with } \alpha, \beta, \gamma \geq 0\]
\Cref{table:init-schemes} shows six commonly used weight initialization
schemes. Several schemes use the same idea, that unit-variance is desired for
each layer as the training converges faster~\cite{BatchNormalization-2015}.
\begin{table}[H]
    \centering
    \begin{tabular}{lllll}
    \toprule
    Name                  & $\alpha$    & $\beta$   & $\gamma$     & Reference \\\midrule
    Constant              & $\alpha=0$  & $\beta=0$ & $\gamma \geq 0$ & used by \cite{zeiler2014visualizing}         \\
    Xavier/Glorot uniform & $\alpha = \sqrt{\frac{6}{n_{in} + n_{out}}}$ & $\beta=0$ & $\gamma = 0$ & \cite{glorot2010understanding}\\
    Xavier/Glorot normal  & $\alpha=0$ & \(\beta = {\left (\frac{2}{(n_{in} + n_{out})} \right )}^2 \) & $\gamma = 0$ & \cite{glorot2010understanding}\\
    He                    & $\alpha=0$ & $\beta = \frac{2}{n_{in}}$ & $\gamma = 0$ & \cite{he2015delving} \\
    Orthogonal            & ---   & ---  & $\gamma = 0$ & \cite{saxe2013exact}\\
    LSUV                  & ---   & ---  & $\gamma = 0$ & \cite{mishkin2015all}\\
    \bottomrule
    \end{tabular}
    \caption[Weight initialization schemes]{Weight initialization schemes of
             the form $w \sim \alpha \cdot \mathcal{U}[-1, 1] + \beta \cdot
             \mathcal{N}(0, 1) + \gamma$.\\
             \(n_{in}, n_{out}\) are the number of units in the previous layer
             and the next layer. Typically, biases are initialized with
             constant 0 and weights by one of the other schemes to prevent
             unit-coadaptation. However, dropout makes it possible to use
             constant initialization for all parameters.\\
             LSUV and Orthogonal initialization cannot be described with this
             simple pattern.}
    \label{table:init-schemes}
\end{table}

\section{Objective function}\label{sec:objective-function}
For classification tasks, the cross-entropy
\[E_{CE}(W) = - \sum_{x \in X} \sum_{k=1}^K \left [t_k^x \log(o_k^x) + (1-t_k^x) \log(1-o_k^x) \right ]\]
is by far the most commonly used objective function (e.g., used
by~\cite{zeiler2014visualizing}). In this equation, $X$ is the set of training
examples, $K$ is the number of classes, $t_k^x \in \Set{0, 1}$ indicates if the
training example $x$ is of class~$k$, $o_k^x$ is the output of the classifier
for the training example~$x$ and class~$k$.

However, regularization terms weighted with a constant~$\lambda \in (0, +
\infty)$ are sometimes added:
\begin{itemize}
    \item LASSO: $\ell_1$ (e.g., used in~\cite{han2015learning})
    \item Weight decay: $\ell_2$ (e.g., $\lambda = 0.0005$ as in~\cite{mishkin2016systematic})
    \item Orthogonality regularization ($|(W^T \cdot W - I)|$, see~\cite{vorontsov2017orthogonality})
\end{itemize}

\section{Optimization Techniques}\label{sec:optimization-techniques}
Most relevant optimization techniques for \glspl{CNN} are based on \gls{SGD},
which updates the weights according to the rule
\[w_{ji} \gets w_{ji} + \Delta w_{ji} \text{ with } \Delta w_{ji} = - \eta \frac{\partial E_x}{\partial w_{ji}}\]
where $\eta \in (0, 1)$, typically $0.01$ (e.g., \cite{mishkin2016systematic}),
is called the \textit{learning rate}.

A slight variation of \gls{SGD} is mini-batch gradient descent with the
mini-batch~$B$ (typically mini-batch sizes are $|B| \in \Set{32, 64, 128, 256,
512}$, e.g. \cite{zeiler2014visualizing}). Larger mini-batch sizes lead to
sharp minima and thus poor generalization~\cite{keskar2016large}. Smaller
mini-batch sizes lead to longer training times due to computational overhead
and to more training steps due to gradient noise.

\[w_{ji} \gets w_{ji} + \Delta w_{ji} \text{ with } \Delta w_{ji} = - \eta \frac{\partial E_B}{\partial w_{ji}}\]

Nine variations which adjust the learning rate during training are:
\begin{itemize}
    \item Momentum:
          \[w_{ji}^{(t+1)} \gets w_{ji}^{(t)} + \Delta w_{ji}^{(t+1)} \text{ with } \Delta w_{ji}^{(t+1)} = - \eta \frac{\partial E_B}{\partial w_{ji}} + \alpha \Delta w_{ji}^{(t)}\]
          with $\alpha \in [0, 1]$, typically $0.9$ (e.g., \cite{zeiler2014visualizing,mishkin2016systematic})
    \item Adagrad~\cite{duchi2011adaptive}
    \item RProp and the mini-batch version RMSProp~\cite{tieleman2012lecture}
    \item Adadelta~\cite{zeiler2012adadelta}
    \item Power Scheduling~\cite{xu2011towards}:
          $\eta(t) = \eta(0) (1+a \cdot t)^{-c}$, where $t \in
          \mathbb{N}_0$ is the training step, $a, c$ are constants.
    \item Performance Scheduling~\cite{senior2013empirical}: Measure the error
          on the cross validation set and decrease the learning rate when the
          algorithms improvement is below a threshold.
    \item Exponential Decay Learning Rate~\cite{senior2013empirical}:
          $\eta(t) = \eta(0) \cdot 10^{-\frac{t}{k}}$ where $t \in \mathbb{N}_0$ is
          the training step, $\eta(0)$ is the initial learning rate, $k \in \mathbb{N}_{\geq 1}$
          is the number of training steps until the learning rate is decreased
          by $\frac{1}{10}$th.
    \item Newbob Scheduling~\cite{newbob}: Start with Performance Scheduling, then use Exponential Decay Scheduling.
    \item Adam and AdaMax~\cite{kingma2014adam}
    \item Nadam~\cite{dozat2015incorporating}
\end{itemize}
Some of those are explained in~\cite{ruder2016overview}.

Other first-order gradient optimization methods are:
\begin{itemize}
    \item Quickprop~\cite{quickprop}
    \item \Gls{NAG}~\cite{nesterov1983method}
    \item Conjugate Gradient method~\cite{charalambous1992conjugate}: Combines
          a line search for the step size with the gradients direction.
\end{itemize}

Higher-order gradient methods like Newtons method or quasi-Newton methods
like BFGS and L-BFGS need the inverse of the Hessian matrix which is intractable
for today's \glspl{CNN}.

However, there are alternatives which do not use gradient information:
\begin{itemize}
    \item Genetic algorithms such as \gls{NEAT}~\cite{stanley2002evolving}
    \item Simulated Annealing~\cite{vanLaarhoven1987}
    \item Twiddle: A local hill-climbing algorithm explained by Sebastian Thrun
          and described on~\cite{Thom2014}
\end{itemize}

There are also approaches which learn the optimization
algorithm~\cite{andrychowicz2016learning,li2016learning}.

\clearpage
\section{Network Design}\label{sec:network-design}
\Glspl{CNN} have the following hyperparameters:
\begin{itemize}[noitemsep]
    \item \textbf{Depth}: The number of layers
    \item \textbf{Width}: The number of filters per layer
    \item \textbf{Layer and block connectivity graph}
    \item \textbf{Layer and block hyperparameters}:
        \begin{itemize}
            \item Activation Functions as shown
                  in~\cref{table:activation-functions-overview}
            \item For more, see~\cref{sec:CNN-layer-types,sec:cnn-blocks}.
        \end{itemize}
\end{itemize}
\begin{table}[H]
    \centering
    \hspace*{-1cm}\begin{tabular}{lllll}
    \toprule
    Name                     & Function $\varphi(x)$ & Range of Values & $\varphi'(x)$ & Used by \\\midrule
    Sign function$^\dagger$  & $\begin{cases}+1 &\text{if } x \geq 0\\-1 &\text{if } x < 0\end{cases}$ & $\Set{-1,1}$                              & $0$                    & \cite{971754} \\
    \parbox[t]{2.6cm}{Heaviside\\step function$^\dagger$} & $\begin{cases}+1 &\text{if } x > 0\\0 &\text{if } x < 0\end{cases}$ & $\Set{0, 1}$  & $0$                       & \cite{mcculloch1943logical}\\
    Logistic function        & $\frac{1}{1+e^{-x}}$                           & $[0, 1]$                                                        & $\frac{e^x}{(e^x +1)^2}$  & \cite{duch1999survey} \\
    Tanh                     & $\frac{e^x - e^{-x}}{e^x + e^{-x}} = \tanh(x)$ & $[-1, 1]$                                                       & $\sech^2(x)$              & \cite{LeNet-5,Thoma:2014}\\
    \gls{ReLU}$^\dagger$           & $\max(0, x)$                                   & $[0, +\infty)$                                                  & $\begin{cases}1 &\text{if } x > 0\\0 &\text{if } x < 0\end{cases}$      & \cite{AlexNet-2012}\\
    \parbox[t]{2.6cm}{\gls{LReLU}$^\dagger$\footnotemark\\(\gls{PReLU})} & $\varphi(x) = \max(\alpha x, x)$                        & $(-\infty, +\infty)$                                             & $\begin{cases}1 &\text{if } x > 0\\\alpha &\text{if } x < 0\end{cases}$ & \cite{maas2013rectifier,he2015delving} \\
    Softplus                 & $\log(e^x + 1)$                                & $(0, +\infty)$                                       & $\frac{e^x}{e^x + 1}$ & \cite{dugas2001incorporating,glorot2011deep} \\
    \gls{ELU}                & $\begin{cases}x &\text{if } x > 0\\\alpha (e^x - 1) &\text{if } x \leq 0\end{cases}$ & $(-\infty, +\infty)$ & $\begin{cases}1 &\text{if } x > 0\\\alpha e^x &\text{otherwise}\end{cases}$ & \cite{clevert2015fast} \\
    Softmax$^\ddagger$       & $o(\mathbf{x})_j = \frac{e^{x_j}}{\sum_{k=1}^K e^{x_k}}$    & $[0, 1]^K$                                           & $o(\mathbf{x})_j \cdot \frac{\sum_{k=1}^K e^{x_k} - e^{x_j}}{\sum_{k=1}^K e^{x_k}}$         & \cite{AlexNet-2012,Thoma:2014}\\
    Maxout$^\ddagger$        & $o(\mathbf{x}) = \max_{x \in \mathbf{x}} x$                 & $(-\infty, +\infty)$                                 & $\begin{cases}1 &\text{if } x_i = \max \mathbf{x}\\0 &\text{otherwise}\end{cases}$         & \cite{goodfellow2013maxout}       \\
    \bottomrule
    \end{tabular}
    \caption[Activation functions]{Overview of activation functions. Functions
             marked with $\dagger$ are not differentiable at 0 and functions
             marked with $\ddagger$ operate on all elements of a layer
             simultaneously. The hyperparameters $\alpha \in (0, 1)$ of Leaky
             ReLU and ELU are typically $\alpha = 0.01$. Other activation
             function like randomized leaky ReLUs exist~\cite{xu2015empirical},
             but are far less commonly used.\\
             Some functions are smoothed versions of others, like the logistic
             function for the Heaviside step function, tanh for the sign
             function, softplus for ReLU.\\
             Softmax is the standard activation function for the last layer of
             a classification network as it produces a probability
             distribution. See \Cref{fig:activation-functions-plot} for a plot
             of some of them.}
    \label{table:activation-functions-overview}
\end{table}
\footnotetext{$\alpha$ is a hyperparameter in leaky ReLU, but a learnable parameter in the parametric ReLU function.}

\begin{figure}[ht]
    \centering
    \begin{tikzpicture}
        \definecolor{color1}{HTML}{E66101}
        \definecolor{color2}{HTML}{FDB863}
        \definecolor{color3}{HTML}{B2ABD2}
        \definecolor{color4}{HTML}{5E3C99}
        \begin{axis}[
            legend pos=north west,
            legend cell align={left},
            axis x line=middle,
            axis y line=middle,
            x tick label style={/pgf/number format/fixed,
                                /pgf/number format/fixed zerofill,
                                /pgf/number format/precision=1},
            y tick label style={/pgf/number format/fixed,
                                /pgf/number format/fixed zerofill,
                                /pgf/number format/precision=1},
            grid = major,
            width=16cm,
            height=8cm,
            grid style={dashed, gray!30},
            xmin=-2,     
            xmax= 2,     
            ymin=-1,     
            ymax= 2,     
            xlabel=x,
            ylabel=y,
            tick align=outside,
            enlargelimits=false]
          \addplot[domain=-2:2, color1, ultra thick,samples=500] {1/(1+exp(-x))};
          \addplot[domain=-2:2, color2, ultra thick,samples=500] {tanh(x)};
          \addplot[domain=-2:2, color4, ultra thick,samples=500] {max(0, x)};
          \addplot[domain=-2:2, color4, ultra thick,samples=500, dashed] {ln(exp(x) + 1)};
          \addplot[domain=-2:2, color3, ultra thick,samples=500, dotted] {max(x, exp(x) - 1)};
          \addlegendentry{$\varphi_1(x)=\frac{1}{1+e^{-x}}$}
          \addlegendentry{$\varphi_2(x)=\tanh(x)$}
          \addlegendentry{$\varphi_3(x)=\max(0, x)$}
          \addlegendentry{$\varphi_4(x)=\log(e^x + 1)$}
          \addlegendentry{$\varphi_5(x)=\max(x, e^x - 1)$}
        \end{axis}
    \end{tikzpicture}
    \caption[Activation functions]{Activation functions plotted in $[-2, +2]$.
             $\tanh$ and ELU are able to produce negative numbers. The image of
             ELU, ReLU and Softplus is not bound on the positive side, whereas
             $\tanh$ and the logistic function are always below~1.}
    \label{fig:activation-functions-plot}
\end{figure}
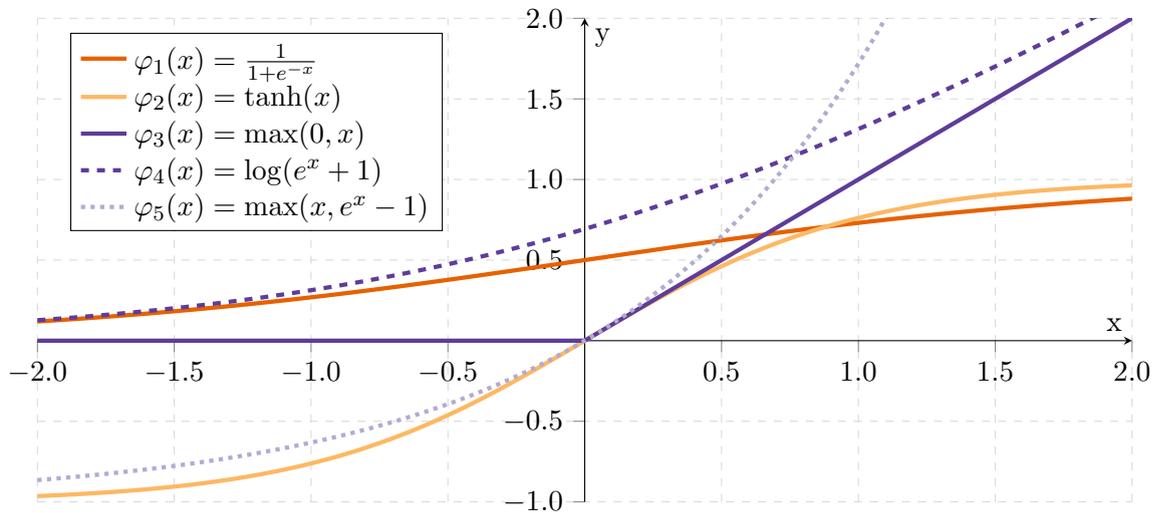

\section{Regularization}\label{sec:regularization}
Regularization techniques aim to make the fitted function smoother and reduce
overfitting. Regularization techniques are:
\begin{itemize}
    \item $\ell_1$, $\ell_2$, and Orthogonality regularization: See~\cref{sec:objective-function}
    \item Max-norm regularization (e.g. used ins~\cite{srivastava2014dropout})
    \item Dropout (introduced in~\cite{srivastava2014dropout}), DropConnect (see~\cite{wan2013regularization}), Stochastic Depth (see~\cite{huang2016deep})
    \item Feature scale clipping (see~\cite{zeiler2014visualizing})
    \item Data augmentation (according to~\cite{zhang2016understanding})
    \item Global average pooling (according to~\cite{zhou2015learning})
    \item Dense-Sparse-Dense training (see~\cite{han2016dsd})
    \item Soft targets (see~\cite{hinton2015distilling})
\end{itemize}

\chapter{Calculating Network Characteristics}
\section{Parameter Numbers}
\begin{itemize}
    \item A fully connected layer with $n$ nodes, $k$ inputs has $n \cdot (k+1)$
          parameters. The $+1$ is due to the bias.
    \item A convolutional layer $i$ with $k_i$ filters of size $n \times m$
          being applied to $k_{i-1}$ feature maps has $k_i \cdot k_{i-1} (n
          \cdot m + 1)$ parameters. The $+1$ is due to the bias.
    \item A fully connected layer with $n$ nodes after $k$ feature maps of size
          $m_1 \times m_2$ has $n \cdot (k \cdot m_1 \cdot m_2 + 1)$ parameters.
    \item A dense block with a depth of $L$, a growth rate of $n$ and $3\times 3$ filters has
          $L + n \cdot 3^2 + 3^2 \cdot n^2 \sum_{i=0}^L (L - i) = L + 9n + 9n^2 \frac{L^2 - L}{2}$ parameters.
\end{itemize}

According to~\cite{han2015learning}, AlexNet has 60~million parameters which is
roughly the number calculated in~\cref{table:AlexNet-architecture}.

\section{FLOPs}
The \glspl{FLOP} of a layer depend on the implementation, the compiler and the
hardware. Hence the following number are only giving rough estimates.

In the following, $n_\varphi$ denotes the number of \glspl{FLOP} to compute
the non-linearity~$\varphi$. For simplicity, $n_\varphi = 5$ was chosen.

\begin{itemize}[noitemsep]
    \item A fully connected layer with $n$ nodes and $k$ inputs has to
          calculate $\varphi(W \cdot x + b)$ with $W \in \mathbb{R}^{n \times
          k}$, $x \in \mathbb{R}^{k\times 1}$, $b \in \mathbb{R}^{n \times 1}$.
          It hence needs about $n \cdot (k + (k-1) + 1) = 2nk$ additions /
          multiplications before the non-linearity $\varphi$ is calculated.
          The total number of \glspl{FLOP} is~$2 \cdot n \cdot k + n \cdot n_\varphi$.
    \item In the following, biases are ignored. A convolutional layer with
          $k_i$ filters of size $n \times m$ being applied to $k_{i-1}$ filter
          maps of size $w \times h$ results in $k_i$ filter maps of size $w
          \times h$ if padding is applied.
          For each element of each filter map, $n \cdot m \cdot k_{i-1}$ multiplications and
          $(n \cdot m \cdot k_{i-1} - 1)$ additions have to be made. This results in
          $(2nm k_{i-1} - 1) \cdot (k_i \cdot w \cdot h)$ operations.
          The total number of \glspl{FLOP} is~$(2 \cdot n \cdot m \cdot k_{i-1} - 1) \cdot (k_i \cdot w \cdot h) + k_i \cdot w \cdot h \cdot n_\varphi$.\\
          This is, of course, a naive way of calculating a convolution. There
          are other ways of calculating convolutions~\cite{lavin2016fast}.
    \item A fully connected layer with $n$ nodes after $k$ feature maps of size
          $w \times h$ needs $2n(k \cdot w \cdot h)$ \glspl{FLOP}.
          The total number of \glspl{FLOP} is~$2n \cdot (k \cdot w \cdot h) + n \cdot n_\varphi$.
    \item As Dropout is only calculated during training, the number of
          \glspl{FLOP} was set to~0.
    \item The number of \glspl{FLOP} for max pooling is dominated by the number
          of positions to which the pooling kernel is applied. For a feature
          map of size $w \times h$ a max pooling filter with stride $s$ gets
          applied $\frac{w \cdot h}{s^2}$. The number of \glspl{FLOP} per
          application depends on the kernel size. A $2 \times 2$ kernel is
          assumed to need 5~\glspl{FLOP}.
    \item The number of \glspl{FLOP} for Batch Normalization is the same as the
          number of its parameters.
\end{itemize}

Here are some references which give information for the \glspl{FLOP}:
\begin{itemize}[noitemsep]
    \itemsep0em
    \item AlexNet
    \begin{itemize}[noitemsep]
        \item 1.5B in total~\cite{han2015learning}.
        \item  725M in total~\cite{kim2015compression}.
        \item 3300M in total in~\cref{table:AlexNet-architecture}
    \end{itemize}
    \item VGG-16:
        \begin{itemize}[noitemsep]
            \item 15484M in total~\cite{han2015learning}.
            \item 31000M in total in~\cref{table:VGG-16-D-architecture}.
        \end{itemize}
    \item GoogleNet: 1566M in total~\cite{han2015learning}.
\end{itemize}

One can see that the numbers are by a factor of 2 up to a factor of 4
different for the same network.

\section{Memory Footprint}\label{sec:memory-footprint}
The memory footprint of \glspl{CNN} determines when networks can be used at all
and if they can be trained efficiently. In order to be able to train
\glspl{CNN} efficiently, one weight update step has to fit in the memory of the
\gls{GPU}. This includes the following:
\begin{itemize}
    \item \textbf{Activations}: All activations of one mini-batch in order to
          calculate the gradients in the backward pass. This is the number of
          floats in the feature maps of all weight layers combined.
    \item \textbf{Weights}
    \item \textbf{Optimization algorithm}: The optimization algorithm introduces
          some overhead. For example, Adam stores two parameters per weights.
\end{itemize}

At inference time, every two consecutive layers have to fit into memory. When
the forward pass of layer~A to layer~B is calculated, the memory can be freed
if no skip connections are used.

\chapter{Common Architectures}
In the following, some of the most important \gls{CNN} architectures are
explained. Understanding the development of these architectures helps
understanding critical insights the machine learning community got in the past
years for convolutional networks for image recognition.

It starts with LeNet-5 from 1998, continues with AlexNet from 2012, VGG-16 D
from 2014, the Inception modules v1 to v3 as well as ResNets in 2015. The
recently developed Inception-v4 is also covered.

The summation row gives the sum of all floats for the output size column. This
allows conclusions about the maximum mini-batch size which can be in memory for
training.


\clearpage
\section{LeNet-5}
One of the first \glspl{CNN} used was LeNet-5~\cite{LeNet-5}. LeNet-5 uses two
times the common pattern of a single convolutional layer with $\tanh$ as a
non-linear activation function followed by a pooling layer and three fully
connected layers. One fully connected layer is used to get the right output
dimension, another one is necessary to allow the network to learn a non-linear
combination of the features of the feature maps.

\begin{figure}[ht]
    \centering
    \includegraphics*[width=\linewidth, keepaspectratio]{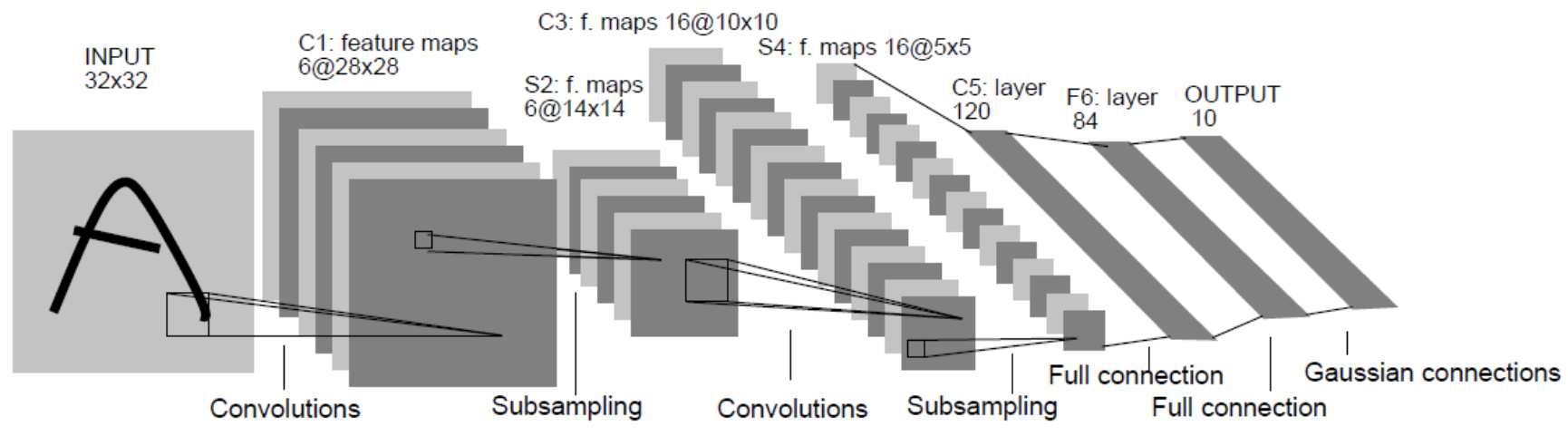}
    \caption[LeNet-5 architecture]{Architecture of LeNet-5 as shown in~\cite{LeNet-5}.}
    \label{fig:LeNet-architecture}
\end{figure}

Its exact architecture is shown in~\cref{fig:LeNet-architecture} and described
in \cref{table:LeNet-5-architecture}. It reaches a test error rate of
$\SI{0.8}{\percent}$ on MNIST.

\begin{table}[ht]
    \renewrobustcmd{\bfseries}{\fontseries{b}\selectfont}
    \sisetup{detect-weight,mode=text,group-minimum-digits = 4}
    \centering
\small
\addtolength{\tabcolsep}{-1.5pt}
\begin{tabular}{
  @{}cl
  S[table-format=3.0]@{\,}c@{\,}l@{\,}c@{\,}l
  S[table-format=5.0]
  S[table-format=8.0]
  S[table-format=4.0]@{\,}c@{\,}S[table-format=2.0]@{}>{${}}c<{{}$}@{}S[table-format=2.0]
  @{}
    }
\toprule
\# & Type
   & \multicolumn{5}{c}{\begin{tabular}[t]{l}Filters @\\ Patch size / stride\end{tabular}}
   & {Parameters} & {FLOPs} & \multicolumn{5}{c}{Output size}\\\midrule
   &    Input
       &&&&&   &           0       &       0   & 1 & @ & 32 & \times & 32    \\
1 &    Convolution
       & 6 & @ & $5 \times 5 \times 1$ & / & 1
       & 156 & 307 800 & \bfseries 6 & @ & \bfseries 28 & \times & \bfseries 28 \\
2 &    Scaled average pooling
       &&& $2 \times 2$ & / & 2
       & 2 & 336 & 6 & @ & 14 & \times & 14 \\
3 &    Convolution
       & 16 & @ & $5 \times 5 \times 6$ & / & 1
       & 2 416 & \bfseries 942 400 & 16 & @ & 10 & \times & 10 \\
4 &    Scaled average pooling
       &&& $2 \times 2$ & / & 2
       & 2 & 1 600 & 16 & @ & 5 & \times & 5 \\
5 &    Fully Connected
        & 120 & \multicolumn{4}{@{}l}{\ neurons}
        & \bfseries 48 120 & 240 000   & 120 \\
6 &    Fully Connected
        & 84 & \multicolumn{4}{@{}l}{\ neurons}
        & 10 164 & 20 580  & 84 \\
7 & Fully Connected (output)
        & 10 & \multicolumn{4}{@{}l}{\ neurons}
        & 850 & 1 730 & 10 \\ \midrule
$\sum$  &&&&&&& 61 710 & 15 144 446 & \multicolumn{5}{r}{9118}\\
\bottomrule
\end{tabular}
    \caption[LeNet-5 architecture]{LeNet-5 architecture: After layers~1, 3, 5 and~6 the $\tanh$ activation
             function is applied. After layer~7, the softmax function is applied.
             One can see that convolutional layer need much fewer parameters, but
             an order of magnitude more \glspl{FLOP} per parameter than
             fully connected layers.}
    \label{table:LeNet-5-architecture}
\end{table}

\clearpage
\section{AlexNet}\label{subsec:AlexNet}
\begin{figure}[H]
    \centering
    \includegraphics*[width=\linewidth, keepaspectratio]{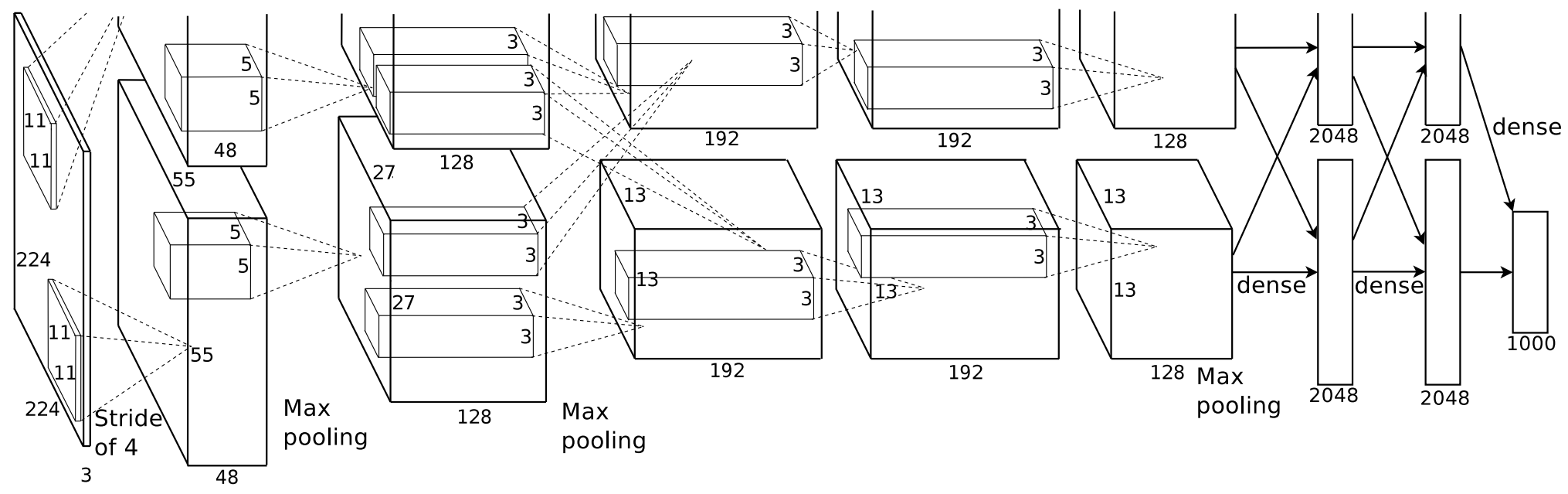}
    \caption[AlexNet architecture]{Architecture of AlexNet as shown in~\cite{AlexNet-2012}: Convolutional Layers are followed by pooling layers multiple times. At the end, a fully connected network is applied. Conceptually, it is identical to the architecture of LeNet-5 (see~\cref{fig:LeNet-architecture}).}
    \label{fig:AlexNet-architecture}
\end{figure}

The first \gls{CNN} which achieved major improvements on the ImageNet dataset
was AlexNet~\cite{AlexNet-2012}. Its architecture is shown
in~\cref{fig:AlexNet-architecture} and described in~\cref{table:AlexNet-architecture}. It has about $60 \cdot 10^6$ parameters. A trained AlexNet can be downloaded at \href{http://www.cs.toronto.edu/~guerzhoy/tf_alexnet/}{www.cs.toronto.edu/\~guerzhoy/tf\_alexnet}.
Note that the uncompressed size is at least
$\SI{60965224}{\floats} \cdot \SI{32}{\bit\per\float} \approx \SI{244}{\mega\byte}$.

\glsunset{LCN}\glsunset{FC}
\begin{table}[ht]
    \centering
    \begin{tabular}{cll
                    S[detect-weight,group-minimum-digits=4,table-format=8.0]
                    rl}
    \toprule
    \# & Type                 & \parbox[t]{3cm}{Filters @\par Patch size / stride}      & {Parameters}   &                       {FLOPs} & Output size   \\\midrule
    ~  & Input                & ~                                                       & ~              &                             ~ & \hphantom{00}3 @ $224 \times 224$ \\
    1  & Convolution          & \hphantom{0}96 @ $11 \times 11 \times 3$ / 4            &          34944 &       \SI{211}{\mega\nothing} & \hphantom{0}\textbf{96\,@}\,$\hphantom{0}\mathbf{55 \times \hphantom{0}55}$  \\
    ~  & \gls{LCN}            & ~                                                       & ~              &        \SI{12}{\mega\nothing} & \hphantom{0}\textbf{96\,@}\,$\hphantom{0}\mathbf{55 \times \hphantom{0}55}$  \\
    2  & Max pooling          & \hphantom{000 @} $3 \times 3\hphantom{\times 0000}$ / 2 &              0 &       \SI{301}{\kilo\nothing} &  \hphantom{0}96 @ $\hphantom{0}27 \times \hphantom{0}27$  \\
    3  & Convolution          & 256 @ $5 \times 5 \times \hphantom{0}48$ / 1            &         307456 &\ubold \SI{448}{\mega\nothing} & 256 @ $\hphantom{0}13 \times \hphantom{0}13$ \\
    ~  & \gls{LCN}            & ~                                                       & ~              &         \SI{3}{\mega\nothing} & 256 @ $\hphantom{0}13 \times \hphantom{0}13$ \\
    4  & Max pooling          & \hphantom{000 @} $3 \times 3\hphantom{\times 0000}$ / 2 &              0 &        \SI{50}{\kilo\nothing} & 256 @ $\hphantom{0}13 \times \hphantom{0}13$ \\
    5  & Convolution          & 384 @ $3 \times 3 \times 256$ / 1                       &         885120 &       \SI{299}{\mega\nothing} & 384 @ $\hphantom{0}13 \times \hphantom{0}13$ \\
    7  & Convolution          & 384 @ $3 \times 3 \times 192$ / 1                       &         663936 &       \SI{224}{\mega\nothing} & 384 @ $\hphantom{0}13 \times \hphantom{0}13$  \\
    9  & Convolution          & 256 @ $3 \times 3 \times 192$ / 1                       &         442624 &       \SI{150}{\mega\nothing} & 256 @ $\hphantom{0}13 \times \hphantom{0}13$ \\
    10 & Max pooling          & \hphantom{000 @} $3 \times 3\hphantom{\times 0000}$ / 2 &              0 &        \SI{50}{\kilo\nothing} & 256 @ $\hphantom{00}6 \times \hphantom{00}6$ \\
    11 & \gls{FC}             & 4096 neurons                                            &\ubold 37752832 &        \SI{75}{\mega\nothing} & 4096          \\
    12 & \gls{FC}             & 4096 neurons                                            &       16781312 &        \SI{34}{\mega\nothing} & 4096          \\
    13 & \gls{FC}             & 1000 neurons                                            &        4097000 &         \SI{8}{\mega\nothing} & 1000 \\\midrule
    $\sum$ &                  &                                                         &       60965224 &      \SI{3300}{\mega\nothing} & \multicolumn{1}{r}{\num{1122568}}\\
    \bottomrule
    \end{tabular}
    \caption[AlexNet architecture]{AlexNet architecture: One special case of
             AlexNet is grouping of convolutions due to computational
             restrictions at the time of its development. This also reduces the
             number of parameters and allows parallel computation on separate
             \glspl{GPU}. However, to make the architecture easier to compare,
             this grouping was ignored for the parameter count. The
             \glspl{FLOP} are taken from~\cite{han2015learning} and combined
             with rough estimates for Local Contrast Normalization and max
             pooling.\\
             The calculated number of parameters was checked against the
             downloaded version. It also has \num{60965224} parameters.}
    \label{table:AlexNet-architecture}
\end{table}

\clearpage
\section{VGG-16 D}\label{subsec:vgg-16-d}
\begin{figure}[H]
    \hspace*{-0.8cm}
    \resizebox {1.05\columnwidth} {!} {
    \newcommand{\width}{0.2}
\tikzstyle{input}=[draw,fill=red!50]
\tikzstyle{conv}=[draw,fill=black!20]
\tikzstyle{max}=[draw,dashed,fill=black!10]
\tikzstyle{dropout}=[draw,dashed,fill=blue!10]
\tikzstyle{fc}=[draw,fill=green!10]
\tikzstyle{output}=[draw,fill=red!50]
\def \coldist {2.3}
\def \widthb {1.9}

\begin{tikzpicture}[scale=2]
    \draw[->, -Latex, line width=5pt]
        (1.0+0*\coldist, 0.5) --(1.0+0*\coldist, -2.2) --(2.1+0*\coldist, -2.2) --(2.2+0*\coldist, 0.5)
     -- (1.0+1*\coldist, 0.5) --(1.0+1*\coldist, -2.2) --(2.1+1*\coldist, -2.2) --(2.2+1*\coldist, 0.5)
     -- (1.0+2*\coldist, 0.5) --(1.0+2*\coldist, -2.9) --(2.1+2*\coldist, -2.9) --(2.2+2*\coldist, 0.5)
     -- (1.0+3*\coldist, 0.5) --(1.0+3*\coldist, -2.9) --(2.1+3*\coldist, -2.9) --(2.2+3*\coldist, 0.5)
     -- (1.0+4*\coldist, 0.5) --(1.0+4*\coldist, -2.9) --(2.1+4*\coldist, -2.9) --(2.2+4*\coldist, 0.5)
     -- (1.0+5*\coldist, 0.5) --(1.0+5*\coldist, -4.5);

    \draw[draw=none] (0*\coldist,-0.5) rectangle (1.0,-0.7) node[pos=.5] {$224 \times 224$};
    \draw[input]     (0*\coldist, 0.0) rectangle (2.0,-0.5) node[pos=.5] {Input};
    \draw[conv]      (0*\coldist,-0.7) rectangle (2.0,-1.2) node[pos=.5] {C $64@3 \times 3 / 1$};
    \draw[conv]      (0*\coldist,-1.4) rectangle (2.0,-1.9) node[pos=.5] {C $64@3 \times 3 / 1$};

    \draw[draw=none] (1*\coldist,-0.5) rectangle (1*\coldist+\widthb/2,-0.7) node[pos=.5] {$112 \times 112$};
    \draw[max]       (1*\coldist, 0.0) rectangle (1*\coldist+\widthb,-0.5) node[pos=.5] {max pooling $2\times 2 / 1$};
    \draw[conv]      (1*\coldist,-0.7) rectangle (1*\coldist+\widthb,-1.2) node[pos=.5] {C $128@3 \times 3 / 1$};
    \draw[conv]      (1*\coldist,-1.4) rectangle (1*\coldist+\widthb,-1.9) node[pos=.5] {C $128@3 \times 3 / 1$};

    \draw[draw=none] (2*\coldist,-0.5) rectangle (2*\coldist+\widthb/2,-0.7) node[pos=.5] {$56 \times 56$};
    \draw[max]       (2*\coldist, 0.0) rectangle (2*\coldist+\widthb,-0.5) node[pos=.5] {max pooling $2\times 2 / 1$};
    \draw[conv]      (2*\coldist,-0.7) rectangle (2*\coldist+\widthb,-1.2) node[pos=.5] {C $256@3 \times 3 / 1$};
    \draw[conv]      (2*\coldist,-1.4) rectangle (2*\coldist+\widthb,-1.9) node[pos=.5] {C $256@3 \times 3 / 1$};
    \draw[conv]      (2*\coldist,-2.1) rectangle (2*\coldist+\widthb,-2.6) node[pos=.5] {C $256@3 \times 3 / 1$};

    \draw[draw=none] (3*\coldist,-0.5) rectangle (3*\coldist+\widthb/2,-0.7) node[pos=.5] {$28 \times 28$};
    \draw[max]       (3*\coldist,-0.0) rectangle (3*\coldist+\widthb,-0.5) node[pos=.5] {max pooling $2\times 2 / 1$};
    \draw[conv]      (3*\coldist,-0.7) rectangle (3*\coldist+\widthb,-1.2) node[pos=.5] {C $512@3 \times 3 / 1$};
    \draw[conv]      (3*\coldist,-1.4) rectangle (3*\coldist+\widthb,-1.9) node[pos=.5] {C $512@3 \times 3 / 1$};
    \draw[conv]      (3*\coldist,-2.1) rectangle (3*\coldist+\widthb,-2.6) node[pos=.5] {C $512@3 \times 3 / 1$};

    \draw[draw=none] (4*\coldist,-0.5) rectangle (4*\coldist+\widthb/2,-0.7) node[pos=.5] {$14 \times 14$};
    \draw[max]       (4*\coldist,-0.0) rectangle (4*\coldist+\widthb,-0.5) node[pos=.5] {max pooling $2\times 2 / 1$};
    \draw[conv]      (4*\coldist,-0.7) rectangle (4*\coldist+\widthb,-1.2) node[pos=.5] {C $512@3 \times 3 / 1$};
    \draw[conv]      (4*\coldist,-1.4) rectangle (4*\coldist+\widthb,-1.9) node[pos=.5] {C $512@3 \times 3 / 1$};
    \draw[conv]      (4*\coldist,-2.1) rectangle (4*\coldist+\widthb,-2.6) node[pos=.5] {C $512@3 \times 3 / 1$};

    \draw[draw=none] (5*\coldist,-0.5) rectangle (5*\coldist+\widthb/2,-0.7) node[pos=.5] {$7 \times 7$};
    \draw[max]       (5*\coldist,-0.0) rectangle (5*\coldist+\widthb,-0.5) node[pos=.5] {max pooling $2\times 2 / 1$};
    \draw[fc]        (5*\coldist,-0.7) rectangle (5*\coldist+\widthb,-1.2) node[pos=.5] {Fully Connected 4096};
    \draw[dropout]   (5*\coldist,-1.4) rectangle (5*\coldist+\widthb,-1.9) node[pos=.5] {Dropout, $p=0.5$};
    \draw[fc]        (5*\coldist,-2.1) rectangle (5*\coldist+\widthb,-2.6) node[pos=.5] {Fully Connected 4096};
    \draw[dropout]   (5*\coldist,-2.8) rectangle (5*\coldist+\widthb,-3.3) node[pos=.5] {Dropout, $p=0.5$};
    \draw[output]    (5*\coldist,-3.5) rectangle (5*\coldist+\widthb,-4.0) node[pos=.5] {Fully Connected 1000};
\end{tikzpicture}
    }
    \caption[VGG-16~D architecture]{Architecture of VGG-16~D. \texttt{C $512@3
    \times 3 / 1$} is a convolutional layer with 512~filters of kernel size $3
    \times 3$ with stride~1. All convolutional layers use \texttt{SAME}
    padding.}
    \label{fig:VGG-16-architecture}
\end{figure}
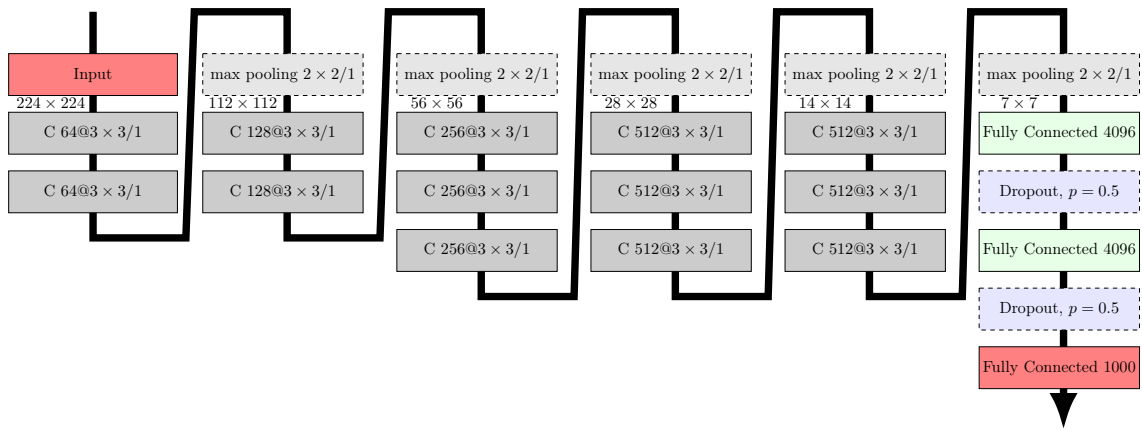

Another widespread architecture is the VGG-16 (D)~\cite{VGG-16}. VGG comes from
the \textbf{V}isual \textbf{G}eometry \textbf{G}roup in Oxford which
developed this architecture. It has \textbf{16} layers which can learn
parameters. A major difference compared to AlexNet is that VGG-16 uses only
$3\times 3$ filters and is much deeper. A visualization of the architecture is
shown in~\cref{fig:VGG-16-architecture} and a detailed textual description is
given in~\cref{table:VGG-16-D-architecture}.

A trained VGG-16 D for Tensorflow can be downloaded at \url{https://github.com/machrisaa/tensorflow-vgg}. Note that the uncompressed size is at least
$\SI{138357544}{\floats} \cdot \SI{32}{\bit\per\float} \approx \SI{520}{\mega\byte}$.
The downloaded Numpy binary file \verb+npz+ needs $\SI{553}{\mega\byte}$ without
compression and $\SI{514}{\mega\byte}$ with compression.

\clearpage

\begin{table}[ht]
    \centering
    \begin{tabular}{cll
                    S[detect-weight,group-minimum-digits=4,table-format=9.0]
                    rl}
    \toprule
    \# & Type             & \parbox[t]{3cm}{Filters @\par Patch size / stride}      & {Parameters}  &{FLOPs}                  & Output size   \\\midrule
    ~  & Input            & ~                                                       &  ~            &                         & \hphantom{00}3 @ $224 \times 224$ \\
    1  & Convolution      & \hphantom{0}64 @ $3 \times 3 \times \hphantom{00}3$ / 1 &         1792  & \SI{186}{\mega\nothing} & \hphantom{0}\textbf{64\,@} $\mathbf{224 \times 224}$ \\
    2  & Convolution      & \hphantom{0}64 @ $3 \times 3 \times \hphantom{0}64$ / 1 &        36928  &\ubold \SI{3712}{\mega\nothing} & \hphantom{0}\textbf{64\,@} $\mathbf{224 \times 224}$ \\
    ~  & Max pooling      & \hphantom{000 @} $2\times 2 \hphantom{\times 0000}$ / 2 &            0  &   \SI{2}{\mega\nothing} & \hphantom{0}64 @ $112 \times 112$ \\
    3  & Convolution      & 128 @ $3 \times 3 \times \hphantom{0}64$ / 1            &        73856  &\SI{1856}{\mega\nothing} & 128 @ $112 \times 112$ \\
    4  & Convolution      & 128 @ $3 \times 3 \times 128$ / 1                       &       147584  &\SI{3705}{\mega\nothing} & 128 @ $112 \times 112$ \\
    ~  & Max pooling      & \hphantom{000 @} $2\times 2 \hphantom{\times 0000}$ / 2 &            0  &   \SI{1}{\mega\nothing} & 128 @ $\hphantom{0}56 \times \hphantom{0}56$ \\
    5  & Convolution      & 256 @ $3 \times 3 \times 128$ / 1                       &       295168  &\SI{1853}{\mega\nothing} & 256 @ $\hphantom{0}56 \times \hphantom{0}56$ \\
    6  & Convolution      & 256 @ $3 \times 3 \times 256$ / 1                       &       590080  &\SI{3703}{\mega\nothing} & 256 @ $\hphantom{0}56 \times \hphantom{0}56$ \\
    7  & Convolution      & 256 @ $3 \times 3 \times 256$ / 1                       &       590080  &\SI{3703}{\mega\nothing} & 256 @ $\hphantom{0}56 \times \hphantom{0}56$ \\
    ~  & Max pooling      & \hphantom{000 @} $2\times 2 \hphantom{\times 0000}$ / 2 &            0  &  <\SI{1}{\mega\nothing} & 256 @ $\hphantom{0}28 \times \hphantom{0}28$ \\
    8  & Convolution      & 512 @ $3 \times 3 \times 256$ / 1                       &      1180160  &\SI{1851}{\mega\nothing} & 512 @ $\hphantom{0}28 \times \hphantom{0}28$ \\
    9  & Convolution      & 512 @ $3 \times 3 \times 512$ / 1                       &      2359808  &\SI{3701}{\mega\nothing} & 512 @ $\hphantom{0}28 \times \hphantom{0}28$ \\
    10 & Convolution      & 512 @ $3 \times 3 \times 512$ / 1                       &      2359808  &\SI{3701}{\mega\nothing} & 512 @ $\hphantom{0}28 \times \hphantom{0}28$ \\
    ~  & Max pooling      & \hphantom{000 @} $2\times 2 \hphantom{\times 0000}$ / 2 &            0  &  <\SI{1}{\mega\nothing} & 512 @ $\hphantom{0}14 \times \hphantom{0}14$ \\
    11 & Convolution      & 512 @ $3 \times 3 \times 512$ / 1                       &      2359808  & \SI{925}{\mega\nothing} & 512 @ $\hphantom{0}14 \times \hphantom{0}14$ \\
    12 & Convolution      & 512 @ $3 \times 3 \times 512$ / 1                       &      2359808  & \SI{925}{\mega\nothing} & 512 @ $\hphantom{0}14 \times \hphantom{0}14$ \\
    13 & Convolution      & 512 @ $3 \times 3 \times 512$ / 1                       &      2359808  & \SI{925}{\mega\nothing} & 512 @ $\hphantom{0}14 \times \hphantom{0}14$ \\
    ~  & Max pooling      & \hphantom{000 @} $2\times 2 \hphantom{\times 0000}$ / 2 &            0  &  <\SI{1}{\mega\nothing} & 512 @ $\hphantom{00}7 \times \hphantom{00}7$ \\
    14 & \gls{FC}         & 4096 neurons                                            & \ubold 102764544 &\SI{206}{\mega\nothing} & 4096          \\
    ~  & Dropout          &                                                         &             0 &                         0 & 4096          \\
    15 & \gls{FC}         & 4096 neurons                                            &      16781312 &    \SI{34}{\mega\nothing} & 4096          \\
    ~  & Dropout          &                                                         &             0 &                         0 & 4096          \\
    16 & \gls{FC}         & 1000 neurons                                            &       4097000 &     \SI{8}{\mega\nothing} & 1000 \\\midrule
    $\sum$ &              &                                                         &     138357544 & \SI{31000}{\mega\nothing} & \multicolumn{1}{r}{\num{15245800}}\\
    \bottomrule
    \end{tabular}
    \caption[VGG-16~D architecture]{VGG-16~D architecture: The authors chose to
             give only layers a number which have learnable parameters. All
             convolutions are zero padded to prevent size changes and use ReLU
             activation functions. The channels mean is subtracted from each
             pixel as a preprocessing step ($-103.939, - 116.779, -123.68$). As
             Dropout is only calculated during training time, the number of
             \glspl{FLOP} is 0. The dropout probability is \num{0.5}.\\
             The calculated number of parameters was checked against the
             downloaded version. It also has \num{138357544} parameters.}
    \label{table:VGG-16-D-architecture}
\end{table}

\clearpage
\section{GoogleNet, Inception v2 and v3}
\begin{figure}[ht]
    \centering
    \includegraphics*[width=0.6\linewidth, keepaspectratio]{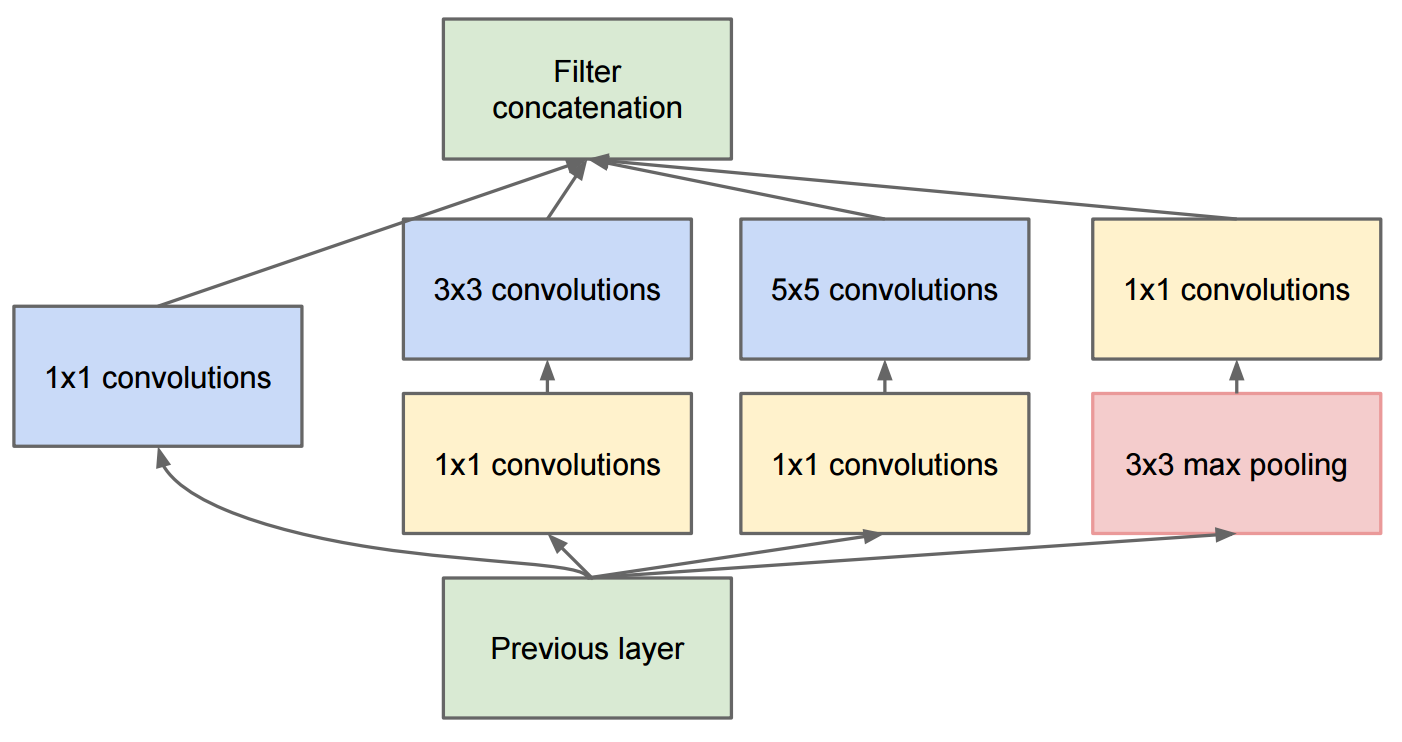}
    \caption[Inception module]{Inception module\\Image source:~\cite{GoogleNet-Inception}}
    \label{fig:inception-v1-module}
\end{figure}

The large number of parameters and operations is a problem when such models
should get applied in practice to thousands of images. In order to reduce the
computational cost while maintaining the classification quality,
GoogleNet~\cite{GoogleNet-Inception} and the Inception module were developed.
The Inception module essentially only computes $1 \times 1$ filters, $3 \times
3$ filters and $5 \times 5$ filters in parallel, but applied bottleneck $1
\times 1$ filters before to reduce the number of parameters. It is shown
in~\cref{fig:inception-v1-module}.

\begin{figure}[ht]
    \centering
    \includegraphics*[width=0.6\linewidth, keepaspectratio]{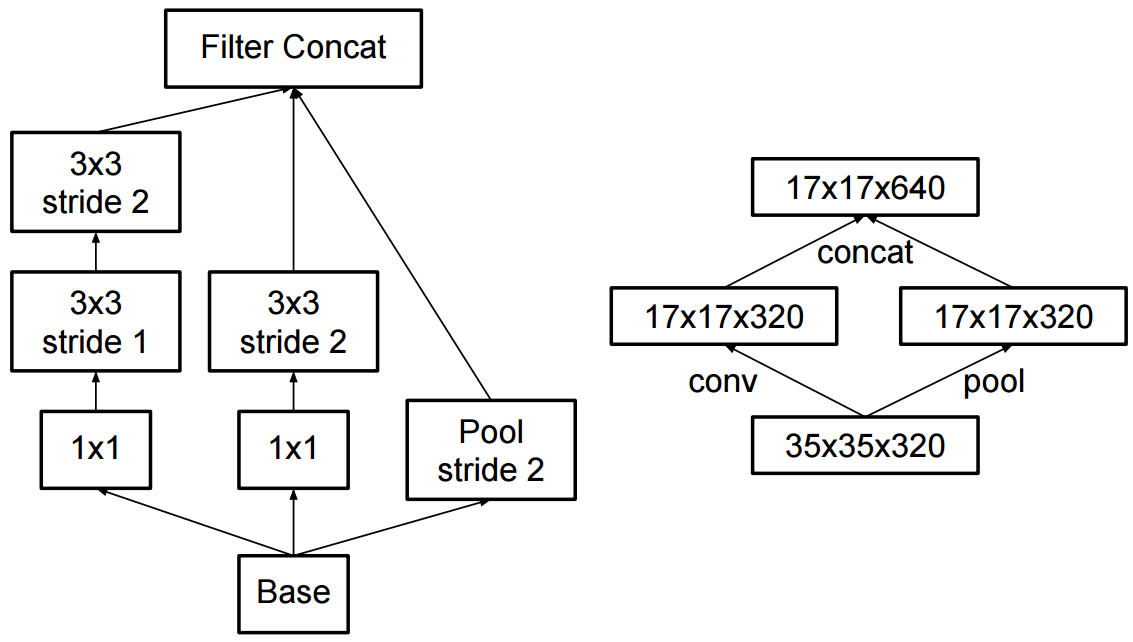}
    \caption[Inception v2 module]{Inception v2 module\\Image source:~\cite{Inception-v3}}
    \label{fig:inception-v2-module}
\end{figure}

Compared to GoogleNet, Inception~v2~\cite{Inception-v3} removed the $5 \times
5$ filters and replaced them by two successive layers of $3 \times 3$ filters.
A visualization of an Inception~v2 module is given
in~\cref{fig:inception-v2-module}. Additionally, Inception~v2 applies
successive asymmetric filters to approximate symmetric filters with fewer
parameters. The authors call this approach \textit{filter factorization}.

Inception~v3 introduced Batch Normalization to the network~\cite{Inception-v3}.

\clearpage
\section{Inception-v4}
Inception-v4 as described in~\cite{inception-v4} consists of four main building
blocks: The stem, Inception~A, Inception~B and Inception~C. To quote the
authors: Inception-v4 is a deeper, wider and more uniform simplified
architecture than Inception-v3. The stem, Reduction~A and Reduction~B use
max-pooling, whereas Inception~A, Inception~B and Inception~C use average
pooling. The stem, module~B and module~C use separable convolutions.

\begin{table}[ht]
    \centering
    \begin{tabular}{crlrr}
    \toprule
    \#  & $\times$  & Type                             & Parameters    & Output size    \\\midrule
    ~   & ~         & Input                            & ~             & \hphantom{000}3 @ $299 \times 299$  \\
    1   & ~         & Stem                             &  \num{605728} & \hphantom{0}384 @ \hphantom{0}$35 \times \hphantom{0}35$  \\
    2   & $4\times$ & \cellcolor{blue!5}Inception A    &  \cellcolor{blue!5}\num{317632} & \cellcolor{blue!5}\hphantom{0}384 @ \hphantom{0}$35 \times \hphantom{0}35$\\
    3   & ~         & Reduction A                      & \num{2306112} & 1024 @ \hphantom{0}$17 \times \hphantom{0}17$ \\
    4   & $7\times$ & \cellcolor{blue!5}Inception B    & \cellcolor{blue!5}\num{2936256} & \cellcolor{blue!5}1024 @ \hphantom{0}$17 \times \hphantom{0}17$\\
    5   & ~         & Reduction B                      & \num{2747392} & 1536 @ \hphantom{00}$8 \times \hphantom{00}8$   \\
    6   & $3\times$ & \cellcolor{blue!5}Inception C    & \cellcolor{blue!5}\num{4553088} & \cellcolor{blue!5}1536 @ \hphantom{00}$8 \times \hphantom{00}8$\\
    ~   & ~         & Global Average Pooling           &            0  & 1536 @ \hphantom{00}$1 \times \hphantom{00}1$   \\
    ~   & ~         & Dropout (p=0.8)                  &            0  & 1536 @ \hphantom{00}$1 \times \hphantom{00}1$   \\
    7   & ~         & Softmax                          & \num{1537000} & 1000           \\\midrule
 $\sum$ & ~         & ~                                &\num{42679816} & \\
    \bottomrule
    \end{tabular}
    \caption[Inception-v4 network]{Inception-v4 network.}
    \label{table:inception-v4-module}
\end{table}
\clearpage

\chapter{Datasets}\label{ch:datasets}
Well-known benchmark datasets for classification problems in computer vision
are listed in~\cref{table:classification-databases}. The best results known to
me are given in~\cref{table:classification-databases-results}. However, every
semantic segmentation dataset (e.g., PASCAL VOC) can also be used to benchmark
image classifiers using~\cref{alg:create-classification-dataset-from-segmentation-dataset}.

\begin{table}[H]
    \centering
    \begin{tabular}{@{}lcrrcl@{}}
    \toprule
    Database      & \parbox{3.5cm}{Image Resolution\\(width $\times$ height)}          & \parbox{1cm}{\centering Number of\\Images}  & \parbox{1cm}{\centering Number of\\Classes}  & Channels & Data source\\\midrule
    MNIST         & $\hphantom{00}\SI{28}{\pixel} \times \SI{28}{\pixel}$            &  \num{70000}                                &  10                                          & 1        & \cite{YannLeCun1998,LeNet-5} \\
    HASYv2        & $\hphantom{00}\SI{32}{\pixel} \times \SI{32}{\pixel}$            & \num{168233}                                & 369                                          & 1        & \cite{thoma2017hasyv2}\\
    SVHN          & $\hphantom{00}\SI{32}{\pixel} \times \SI{32}{\pixel}$            & \num{630420}                                &  10                                          & 3        & \parbox{1cm}{\cite{YuvalNetzer2011},\\\cite{netzer2011reading}} \\
    CIFAR-10      & $\hphantom{00}\SI{32}{\pixel} \times \SI{32}{\pixel}$            &  \num{60000}                                &  10                                          & 3        & \cite{CIFAR-10,krizhevsky2009learning} \\
    CIFAR-100     & $\hphantom{00}\SI{32}{\pixel} \times \SI{32}{\pixel}$            &  \num{60000}                                & 100                                          & 3        & \cite{CIFAR-10,krizhevsky2009learning} \\
    STL-10        & $\hphantom{00}\SI{96}{\pixel} \times \SI{96}{\pixel}$            &  \num{13000}                                &  10                                          & 3        & \cite{STL-10,coates2010analysis} \\[0.4cm]
    Caltech-101   & \parbox{3.5cm}{$\hphantom{\times}(\SIrange{80}{3481}{\pixel})$\\[-0.1cm]$\times (\SIrange{92}{3999}{\pixel})$} & \num{9144}                        & 102      & 3        & \cite{Caltech-101,fei2006one}\\[0.4cm]
    Caltech-256   & \parbox{3.5cm}{$\hphantom{\times}(\SIrange{75}{7913}{\pixel})$\\[-0.1cm]$\times (\SIrange{75}{7913}{\pixel})$} & \num{30607}                       & 257      & 3        & \cite{Griffin2006,GregGriffin2007}\\[0.4cm]
    ILSVRC 2012\footnotemark & \parbox{3.5cm}{$\hphantom{\times}(\SIrange{8}{9331}{\pixel})$\\[-0.1cm]$\times (\SIrange{10}{6530}{\pixel})$} & $1.2 \cdot 10^6$        & \num{1000} & 3      & \cite{ImageNet-download,russakovsky2014imagenet}\\[0.4cm]
    Places365\footnotemark & \parbox{3.5cm}{$\hphantom{\times}(290\text{px}-3158\text{px})$\\[-0.1cm]$\times (225\text{px}-2630\text{px})$} & $1.8 \cdot 10^6$         &  365     & 3        & \cite{Zhou2016,zhou2016places}\\[0.4cm]
    GTSRB         &  \parbox{3.5cm}{$\hphantom{\times}(\SIrange{25}{266}{\pixel})$\\[-0.1cm]$\times (\SIrange{25}{232}{\pixel})$}  & \num{51839}                       &   43     & 3        & \cite{JohannesStallkamp,Stallkamp2012GTSRB}\\[0.4cm]
    Asirra\footnotemark & \parbox{3.5cm}{$\hphantom{\times}(\SIrange{4}{500}{\pixel})$\\[-0.1cm]$\times (\SIrange{4}{500}{\pixel})$} & \num{25000}                     &  2       & 3        & \cite{Asirra2017,asirra-a-captcha-that-exploits-interest-aligned-manual-image-categorization}\\
    Graz-02       & \hspace{-0.7cm}\parbox{3.5cm}{$\hphantom{\text{and }}\SI{480}{\pixel} \times \SI{640}{\pixel}$\\[-0.2cm]and $\SI{640}{\pixel} \times \SI{480}{\pixel}$} & \num{1096}&3&3 & \cite{IG02-dataset,marszalek2007accurate}\\[0.4cm]
    \bottomrule
    \end{tabular}
    \caption[Image Benchmark datasets]{An overview over publicly available
    image databases for classification. The number of images row gives the sum
    of the training and the test images. Some datasets, like SVHN, have
    additional unlabeled data which is not given in this table.}
    \label{table:classification-databases}
\end{table}
\addtocounter{footnote}{-2}
\footnotetext{ImageNet Large Scale Visual Recognition Competition}
\addtocounter{footnote}{1}
\footnotetext{The dimensions are only calculated for the validation set.}
\addtocounter{footnote}{1}
\footnotetext{Asirra is a CAPTCHA created by Microsoft and was used in the \enquote{Cats vs Dogs} competition on Kaggle}

\clearpage

\begin{table}[H]
    \centering
    \begin{tabular}{llrll}
    \toprule
    Dataset       & Model type / name         & Result                                                   & Score       & \parbox{2cm}{Achieved / Claimed by} \\\midrule
    MNIST         & ---                       &  $\SI{0.21}{\percent}\hphantom{ \pm \SI{0.8}{\percent}}$ & error       & \cite{wan2013regularization} \\
    HASYv2        & TF-CNN                    & $\SI{81.00}{\percent}\hphantom{ \pm \SI{0.8}{\percent}}$ & accuracy    & \cite{thoma2017hasyv2}\\
    SVHN          & DenseNet ($k = 24$)       &  $\SI{1.59}{\percent}\hphantom{ \pm \SI{0.8}{\percent}}$ & error       & \cite{huang2016densely}\\
    CIFAR-10      & DenseNet-BC ($k = 40$)    &  $\SI{3.46}{\percent}\hphantom{ \pm \SI{0.8}{\percent}}$ & error       & \cite{huang2016densely}\\
    CIFAR-100     & WRN-28-10                 & $\SI{16.21}{\percent}\hphantom{ \pm \SI{0.8}{\percent}}$ & error       & \cite{loshchilov10sgdr}\\
    STL-10        & SWWAE-4layer              & $\SI{74.80}{\percent}\hphantom{ \pm \SI{0.8}{\percent}}$ & accuracy    & \cite{zhao2015stacked}\\
    Caltech-101   & SPP-net (pretrained)      & $\SI{93.42}{\percent}         { \pm \SI{0.5}{\percent}}$ & accuracy    & \cite{he2014spatial}\\
    Caltech-256   & ZF-Net (pretrained)       &  $\SI{74.2}{\percent}         { \pm \SI{0.3}{\percent}}$ & accuracy    & \cite{zeiler2014visualizing}\\
    ImageNet 2012 & ResNet ensemble           &  $\SI{3.57}{\percent}\hphantom{ \pm \SI{0.8}{\percent}}$ & Top-5 error & \cite{deep-residual-networks-2015}\\
    GTSRB         & MCDNN                     & $\SI{99.46}{\percent}\hphantom{ \pm \SI{0.8}{\percent}}$ & accuracy    & \cite{6033589}\\
    Asirra        & SVM                       &  $\SI{82.7}{\percent}\hphantom{ \pm \SI{0.8}{\percent}}$ & accuracy    & \cite{golle2008machine}\\
    Graz-02       & Optimal NBNN              & $\SI{78.98}{\percent}\hphantom{ \pm \SI{0.8}{\percent}}$ & accuracy    & \cite{behmo2010towards}\\
    \bottomrule
    \end{tabular}
    \caption[State of the Art results]{An overview over state of the art results achieved in computer vision datasets.}
    \label{table:classification-databases-results}
\end{table}

\begin{algorithm}[H]
    \begin{algorithmic}
    \Require Semantic segmentation dataset ($D_S$)
    \Procedure{CreateDataset}{Annotated dataset $D_S$}
        \State $D_C \gets \Call{List}{}$
        \State $w \gets $ desired image width
        \State $h \gets $ desired image height
        \For{Image and associated label $(x, y)$ in $D_S$}
            \State $i \gets \Call{randint}{0, L.\text{width} - w}$
            \State $j \gets \Call{randint}{0, L.\text{height} - h}$
            \State $c_L \gets \Call{crop}{y, (i, j), (i+w, j+h)}$
            \If{at least 50\% of $s$ are of one class}
                \State $c_I \gets \Call{crop}{x, (i, j), (i+w, j+h)}$
                \State $D.\Call{append}{(c_I, c_L)}$
            \EndIf
        \EndFor
        \State \Return ($D_C$)
    \EndProcedure
    \end{algorithmic}
\caption{Create a classification dataset from a semantic segmentation dataset}
\label{alg:create-classification-dataset-from-segmentation-dataset}
\end{algorithm}

\listoftables
\listoffigures
\bibliography{content/bibliography}
\printglossaries%
\cleardoublepage%
\printindex

\end{document}